\pdfoutput=1
\documentclass{article} 
\usepackage{times}
\usepackage{geometry}
\usepackage{mathtools}

\DeclarePairedDelimiter\floor{\lfloor}{\rfloor}

\usepackage{hyperref}
\usepackage{url}

\usepackage{authblk}

\usepackage{graphicx}
\usepackage{subfigure}
\usepackage{multirow}
\usepackage{floatrow}
\newfloatcommand{capbtabbox}{table}[][\FBwidth]

\PassOptionsToPackage{numbers}{natbib}
\usepackage{natbib}

\usepackage{xspace}

\usepackage{enumerate}
\usepackage{amsthm,amsmath,amssymb}
\usepackage{dsfont}
\usepackage{mathtools}

\usepackage[utf8]{inputenc} 
\usepackage[T1]{fontenc}    
\usepackage{hyperref}       
\usepackage{booktabs}       
\usepackage{amsfonts}       
\usepackage{nicefrac}       
\usepackage{microtype}      
\usepackage{xcolor}         
\usepackage{standard_definitions}

\usepackage{algorithm,algorithmicx}
\usepackage[noend]{algpseudocode}
\usepackage{setspace}
\usepackage{xspace}
\usepackage{enumitem}
\usepackage{multirow}
\usepackage[normalem]{ulem} 

\newcommand\blfootnote[1]{%
  \begingroup
  \renewcommand\thefootnote{}\footnote{#1}%
  \addtocounter{footnote}{-1}%
  \endgroup
}

\usepackage{caption}

\newcommand{\sd}{s}
\newcommand{\sx}{\xi}

\newcommand{\Sf}{\mathcal{Z}}
\newcommand{\Sfcard}{Z}

\newcommand{\sff}{z}

\newcommand{\Sd}{\mathcal{S}}
\newcommand{\Sdcard}{S}
\newcommand{\Sdh}[1][h]{\Sd_{#1}}

\newcommand{\Sx}{\Xi}
\newcommand{\Sxh}[1][h]{\Sx_{#1}}
\newcommand{\Sxcard}{|\Xi|}

\newcommand{\PiEnd}{\Pi_d}

\newcommand{\pt}{\upsilon}
\newcommand{\pts}{\Upsilon}

\newcommand{\policies}{\Pi}
\newcommand{\nspolicies}{\Pi_{\textrm{NS}}}

\newcommand{\TV}[1]{\left\|#1\right\|_{\textrm{TV}}}

\newcommand{\ps}{\phi^\star_z}
\newcommand{\edo}{\phi^\star}
\newcommand{\exo}{\phi^\star_{\xi}}

\newcommand{\hatedoh}{\hat{\phi}_{h}}

\newcommand{\unf}{{\tt Unf}}

\newcommand{\PSDP}{{\tt PSDP}\xspace}
\newcommand{\VI}{{\tt VI}\xspace}
\newtheorem{assumption}{Assumption}

\newcommand{\algppe}{{\tt PPE}\xspace}
\newcommand{\algppelong}{Predictive Path Elimination \xspace}
\newcommand{\mycomment}[1]{\hfill\textcolor{blue}{//#1}}

\usepackage{tikz}
\usetikzlibrary{trees}

\usepackage{thmtools,thm-restate}

\usepackage{comment}
\usepackage{color-edits}
\addauthor{ak}{orange}

\usepackage{xcolor}

\newcommand{\alekh}[1]{\textcolor{green}{\{AA: #1\}}}

\newcommand{\YE}[1]{\textcolor{magenta}{\{YE: #1\}}}
\newcommand{\remove}[1]{}

\title{Provable RL with Exogenous Distractors via Multistep Inverse Dynamics}

\begin{document}

\title{Provable RL with Exogenous Distractors via Multistep Inverse Dynamics}

\author[1]{
Yonathan Efroni}
\author[1]{
Dipendra Misra}
\author[1]{
Akshay Krishnamurthy}
\author[2]{
Alekh Agarwal$^\dagger$}
\author[1]{
John Langford}
\affil[1]{Microsoft Research, New York, NY}
\affil[2]{Google}
\maketitle
\blfootnote{$^\dagger$Work was done while the author was at Microsoft Research.}
\blfootnote{\{yefroni, dimisra, akshaykr, jcl\}@microsoft.com, alekhagarwal@google.com}
\begin{abstract}
Many real-world applications of reinforcement learning (RL) require the agent to deal with high-dimensional observations such as those generated from a megapixel camera. Prior work has addressed such problems with representation learning, through which the agent can provably extract endogenous, latent state information from raw observations and subsequently plan efficiently. However, such approaches can fail in the presence of temporally correlated noise in the observations, a phenomenon that is common in practice. We initiate the formal study of latent state discovery in the presence of such \emph{exogenous} noise sources by proposing a new model, the Exogenous Block MDP (EX-BMDP), for rich observation RL. We start by establishing several negative results, by highlighting failure cases of prior representation learning based approaches. Then, we introduce the \algppelong ($\algppe$) algorithm, that  learns a generalization of inverse dynamics and is provably sample and computationally efficient in EX-BMDPs when the endogenous state dynamics are near deterministic. The sample complexity of $\algppe$ depends polynomially on the size of the latent endogenous state space while not directly depending on the size of the observation space, nor the exogenous state space. We provide experiments on challenging exploration problems which show that our approach works empirically. 
\end{abstract}

\section{Introduction}
In many real-world applications such as robotics there can be large disparities in the size of agent's observation space (for example, the image generated by agent's camera), and a much smaller latent state space (for example, the agent's location and orientation) governing the rewards and dynamics. This size disparity offers an opportunity: how can we construct reinforcement learning (RL) algorithms which can learn an optimal policy using samples that scale with the size of the latent state space rather than the size of the observation space? Several families of approaches have been proposed based on solving various ancillary prediction problems including autoencoding~\citep{tang2017exploration,hafner2019learning}, inverse modeling~\citep{pathak2017curiosity,burda2018large}, and contrastive learning~\citep{laskin2020curl} based approaches.  These works have generated some significant empirical successes, but are there provable (and hence more reliable) foundations for their success?  More generally, what are the right principles for learning with latent state spaces?  

\begin{figure}[t]
    \centering
    \includegraphics[scale=0.39]{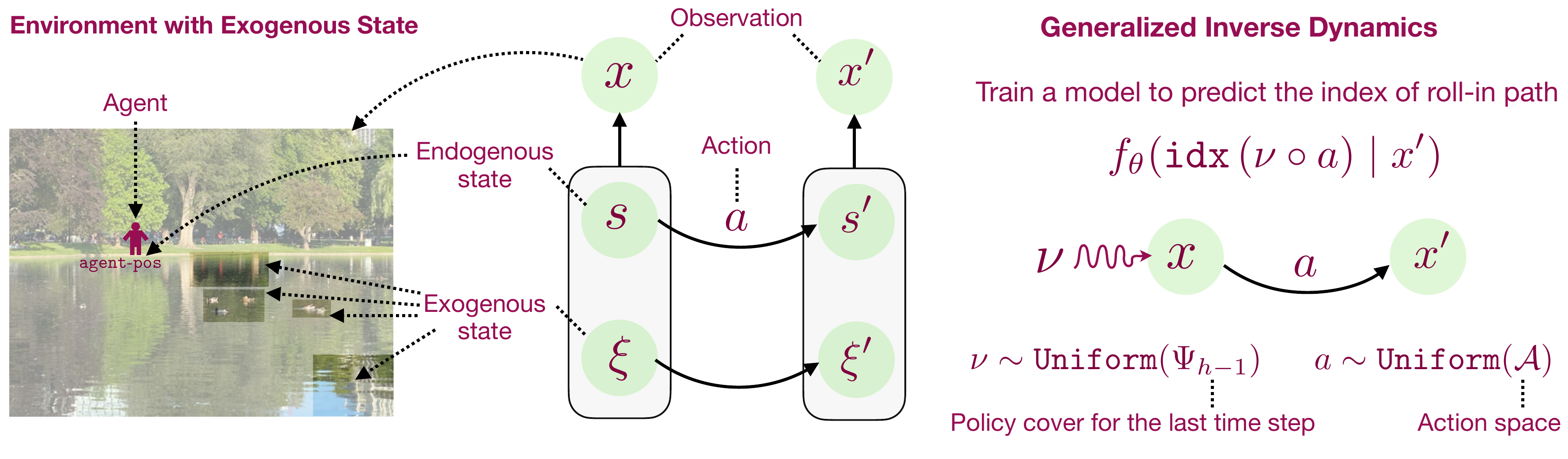}
    \caption{ \footnotesize \textbf{ \footnotesize Left:}  Shows an example where an agent is walking next to a pond in a park and observes the world as an image ($x$). The world consists of a latent endogenous state, containing variable such as agent's position, and a much larger latent exogenous state containing variables such as motion of ducks, changes in sunlight, ripples in the water, etc. \textbf{\footnotesize Center:} The novel \emph{exogenous block MDP} setting introduced here. The exogenous state is unaffected by agent's action. \textbf{\footnotesize Right:} We show that $\algppe$  learns a generalized form of inverse dynamics that is able to recover the endogenous state.}%
    \label{fig:main-intro}%
\end{figure}

In real-world applications, a key issue is robustness to noise in the observation space.  When noise comes from the observation process itself, such as due to measurement error, several  approaches have been recently developed to either explicitly identify~\citep{du2019provably,misra2020kinematic,agarwal2020flambe} or implicitly leverage~\citep{jiang2017contextual} the presence  of latent state structure for provably sample-efficient RL. However, in many real-world scenarios, the observations consist of many elements (e.g. weather, lighting conditions, etc.) with  temporally correlated dynamics (see e.g.~\pref{fig:main-intro} and the example below) that are entirely independent of the agent's actions and rewards. The temporal dynamics of these elements precludes us from treating them as uncorrelated noise, and as such, most previous approaches resort to modeling their dynamics. 
However, this 
is clearly wasteful as these elements have no bearing on the RL problem being solved.

As an example, consider the setting in~\pref{fig:main-intro}. An agent is walking in a park on a lonely sidewalk next to a pond. The agent's observation space is the image generated by its camera, the latent endogenous state is its position on the sidewalk, and the exogenous noise is provided by motion of ducks or people in the background, swaying of trees and changes in lighting conditions, typically unaffected by the agent's actions. \remove{As an example, consider walking down a sidewalk from one house to another.  The agent's observation space is what it sees, the latent space is the position on the sidewalk, and structured uncontrolled noise is provided by the cars driving by on the road.} \remove{ We demonstrate that prior approaches that can effectively leverage the latent state space in the absence of correlated exogenous noise are either prone to create too many latent states in these scenarios or create a too-coarse abstraction that disregards relevant information.  \remove{In this example, they might declare a different latent state based on every combination of cars driving down the road, resulting in a combinatorially large latent state space.} In this example, they might declare a different latent state based on every combination of position of ducks and people, resulting in a combinatorially large latent state space.} While there is a line of recent empirical work that aims to remove causally irrelevant aspects of the observation~\citep{gelada2019deepmdp,zhang2020learning}, theoretical treatment is quite limited~\citep{dietterich2018discovering} and no prior works address sample-efficient learning with provable guarantees.  Given this, the key question here is:

\emph{How can we learn using an amount of data scaling with just the size of the \emph{endogenous} latent state, while ignoring the temporally correlated \emph{exogenous} observation noise?}


We initiate a formal \remove{theoretical} treatment of RL settings where the learner's observations are jointly generated by a latent endogenous state and an uncontrolled exogenous state, which is unaffected by the agent's actions and does not affect the agent's task. \remove{We provide a number of structural results for such problems relating to the properties of optimal policies and value functions, visitation distributions and the complexity of exploration. Based on this understanding, we further.} We study a subset of such problems called Exogenous Block MDPs (EX-BMDPs), where the endogenous state is discrete and decodable from the observations. We first highlight the challenges in solving EX-BMDPs by illustrating the failures of many prior representation learning approaches~\citep{pathak2017curiosity,misra2020kinematic,jiang2017contextual,agarwal2020flambe,zhang2020learning}. These failure happen either due to creating too many latent states, such as one for each combination of ducks and passers-by in the example above leading to sample inefficiency in exploration, or due to lack of exhaustive exploration. 

We identify one recent approach developed by~\citet{du2019provably} with favorable properties for a class of EX-BMDPs with near-deterministic latent state dynamics. In~\pref{sec:learning} and~\pref{sec: theoretical analysis}, we develop a variation of their algorithm and analyze its performance for near-deterministic EX-BMDPs. The algorithm, called Path Prediction and Elimination ($\algppe$), learns a form of \emph{multi-step inverse dynamics} by predicting the identity of the path that generates an observation. For near-deterministic EX-BMDPs, we prove that $\algppe$ successfully explores the environment using $O((SA)^2H\log (|\Fcal|/\delta))$ samples where $S$ is the size of the latent \emph{endogenous} state space, $A$ is the number of actions, $H$ is the horizon and $\Fcal$ is a function class employed to solve a maximum likelihood problem of inferring the path which generated an observation.  Several prior works~\citep{gregor2016variational,paster2020planning} have also considered a multi-step inverse dynamics approach to learn a near optimal policy. However, these works do not consider the EX-BMDP model.
Further, it is unknown whether these algorithms have guarantees for EX-BMDP similar to what we present for $\algppe$. Theoretical analysis of the performance of these algorithms in the presence of exogenous noise is an interesting future work direction.



Empirically, in~\pref{sec: experiments}, we demonstrate the performance of $\algppe$ and various prior baselines in a challenging exploration problem with exogenous noise. We show that baselines fail to decode the endogenous state and either end up over-abstracting (creating an abstract state which aliases essentially-different states) or under-abstracting (creating essentially the same clones of states). We further, show that $\algppe$ is able to recover the latent endogenous model in a visually complex navigation problem, in accordance with the theory. 

\section{Exogenous Block MDP Setting}
\label{sec: setting ex block mdp}

We introduce a novel \emph{Exogenous Block Markov Decision Process} (EX-BMDP) setting to model environments with exogenous noise. We briefly describe notations before presenting a formal treatment of EX-BMDP.

\paragraph{Notations.} For a given set $\Ucal$, we use $\Delta(\Ucal)$ to denote the set of all probability distributions over $\Ucal$. For a given natural number $N \in \NN$, we use the notation $[N]$ to denote the set $\{1, 2, \cdots, N\}$. Lastly, for a probability distribution $p \in \Delta(\Ucal)$, we define its support as $supp~p = \{u \mid p(u) > 0, u \in \Ucal\}$. 

We start with describing the Block Markov Decision Process (BMDP)~\cite{du2019provably}. This process consists of a finite set of observations $\Xcal$, a set of \emph{latent} states $\Sf$ with cardinality $\Sfcard$, a finite set of actions $\Acal$ with cardinality $A$, a transition function $T: \Sf \times \Acal \rightarrow \Delta(\Sf)$, an emission function $q: \Sf \times \Acal \rightarrow \Delta(\Xcal)$, a reward function $R: \Xcal \times \Acal \rightarrow [0, 1]$, a horizon $H \in \NN$, and a start state distribution $\mu \in \Delta(\Sf)$. The agent interacts with the environment by repeatedly generating $H$-step trajectories $(\sff_1, x_1, a_1, r_1, \cdots, \sff_H, x_H, a_H, r_H)$ where $\sff_1 \sim \mu(\cdot)$ and for every $h \in [H]$ we have $x_h \sim q(\cdot \mid \sff_h)$, $r_h = R(x_h, a_h)$, and if $h < H$, then $\sff_{h+1} \sim T(\cdot \mid \sff_h, a_h)$. The agent does not observe the states $\rbr{\sff_1, \cdots, \sff_H}$, instead receiving only the observations $\rbr{x_1,\cdots,x_H}$ and rewards $\rbr{r_1, \cdots, r_H}$. We assume that the emission distributions of any two latent states are disjoint, usually referred as \emph{the block assumption}: 
$supp(q(\cdot\mid z_1))\cap supp(q(\cdot|z_2))=\emptyset\text{ when $z_1\neq z_2.$}$
The agent chooses actions using a policy $\pi: \Xcal \rightarrow \Delta(\Acal)$. 
We also define the set of non-stationary policies $\nspolicies = \policies^H$ as a $H$-length tuple, with $(\pi_1, \cdots, \pi_H) \in \nspolicies$ denoting that the action at time step $h$ is taken as $a_h \sim \pi_h(\cdot \mid x_h)$. The value $V(\pi)$ of a policy $\pi$ is the expected episodic sum of rewards $V(\pi) \defeq \EE_\pi[\sum_{h=1}^H R(x_h, a_h)]$. The optimal policy is given by $\pi^\star = \arg\max_{\pi \in \nspolicies} V(\pi)$. We denote by $\PP_h(x|\pi)$ the probability distribution over observations $x$ at time step $h$ when following a policy $\pi$. Lastly, we refer to an \emph{open loop} 
policy as an element in all $\Acal^H$ sequences of actions. An open loop policy follows a pre-determined sequence of actions $\cbr{a_1,..,a_H}$ for $H$ time steps, unaffected by state information. 

Given the aforementioned definitions, we define an EX-BMDP as follows:

\begin{definition}[Exogenous Block Markov Decision Processes]\label{def: exo endo model}
An EX-BMDP is a BMDP such that the latent state can be decoupled into two parts $\sff =(\sd,\sx)$ where $\sd \in\Sd$ is the endogenous state and $\sx\in \Sx$ is the exogenous state. For $\sff\in \Sf$ the initial distribution and transition functions are decoupled, that is: $\mu(\sff) = \mu(\sd) \mu_{\sx}(\sx)$, and
    $
        T(\sff' \mid \sff, a) = T(\sd' \mid \sd, a) T_{\sx}(\sx' \mid \sx).
    $
\end{definition}

The observation space $\Xcal$ can be arbitrarily large to model which could be a high-dimensional real vector denoting an image, sound, or haptic data in an EX-BMDP. The endogenous state $\sd$ captures the information that can be manipulated by the agent.~\pref{fig:main-intro}, center, visualizes the transition dynamics factorization. We assume that the set of all endogenous states~$\Sd$ is finite with cardinality~$\Sdcard$. The exogenous state $\sx$ captures all the other information that the agent cannot control and does not affect the information it can manipulate. Again, we make no assumptions on the exogenous dynamics nor on its cardinality $\Sxcard$ which may be arbitrarily large. We note that the block assumption of the EX-BMDP implies the existence of two inverse mappings: $\edo: \Xcal \rightarrow \Sd$ to map an observation to its endogenous state, and $\exo: \Xcal\rightarrow \Sx$ to map it to its exogenous state.

\paragraph{Justification of assumptions.} The block assumption has been made by prior work (e.g., \cite{du2019provably}, \cite{zhang2020learning}) to model many real-world settings where the observation is \emph{rich}, i.e., it contains enough information to decode the latent state. The decoupled dynamics assumption made in the EX-BMDP setting is a natural way to characterize exogenous noise; the type of noise that is not affected by our actions nor affects the endogenous state but may have non-trivial dynamic. This decoupling captures the movement of ducks, captured in the visual field of the agent in~\pref{fig:main-intro}, and many additional exogenous processes (e.g., movement of clouds in a navigation task). 

\paragraph{Goal.} Our formal objective is reward-free learning. We wish to find a set of policies, we call a \emph{policy cover}, that can be used to explore the entire state space. Given a policy cover, and for any reward function, we can find a near optimal policy by applying dynamic programming (e.g.,~\cite{bagnell2004policy}), policy optimization (e.g., \cite{kakade2002approximately,agarwal2020optimality,shani2020adaptive}) or value (e.g., \cite{antos2008learning}) based methods.

\begin{definition}[$\alpha$-policy cover]\label{def: approximate policy cover}
Let $\Psi_h$ be a finite set of non-stationary policies. We say $\Psi_h$ is an $\alpha$-policy cover for the  $h^{th}$ time step if for all $\sff\in \Sf$ it holds that ${\max_{\pi\in \Psi_h}\PP_{h}(\sff | \pi)\geq \max_{\pi\in \Pi_{NS}}\PP_{h}(\sff | \pi) -\alpha.}$
If $\alpha=0$ we call $\Psi_h$ a policy cover.
\end{definition}

For standard BMDPs the policy cover is simply the set of policies that reaches each latent state of the BMDP~\citep{du2019provably,misra2020kinematic,agarwal2020flambe}. Thus, for a BMDP, the cardinality of the policy cover scales with $\abr{\Zcal}$. The structure of EX-BMDPs allows to reduce the size of the policy cover significantly to $\abr{\Scal} \ll \abr{\Zcal} = \abr{\Scal}\abr{\Xi}$ when the size of the exogenous state space is large. Specifically, we show that the set of policies that reach each \emph{endogenous} state, and \emph{do not depend on the exogenous} part of the state is also a policy cover (see~\pref{app: structural results of EX-BMDP},~\pref{prop: policy cover and endognous policies}). Further, the proposed algorithm $\algppe$, learns such a policy cover and requires number of samples that depends polynomially on $|\Scal|$ instead of $|\Zcal|$ or $|\Xcal|$.

\section{Failures of Prior Approaches}\label{sec: limitations of current approaches}

We now describe the limitation of prior RL approaches in the presence of exogenous noise. We provide an intuitive analysis over here, and defer a formal statement and proof to~\pref{app:failure of existing approaches}.

\paragraph{Limitation of Noise-Contrastive learning.} Noise-contrastive learning has been used in RL to learn a state abstraction by exploiting temporal information. Specifically, the \textsc{Homer} algorithm~\citep{misra2020kinematic} trains a model to distinguish between \emph{real} and \emph{imposter} transitions. This is done by collecting a dataset of quads $(x, a, x', y)$ where $y=1$ means the transition was $(x, a, x')$ was observed and $y=0$ means that $(x, a, x')$ was not observed. \textsc{Homer} then trains a model $p_\theta(y \mid x, a, \phi_\theta(x'))$ with parameters $\theta$, on the dataset, by predicting whether a given pair of transition was observed or not. This provides a state abstraction $\phi_\theta: \Xcal \rightarrow \NN$ for exploring the environment. \textsc{Homer} can provably solve Block MDPs. Unfortunately, in the presence of exogenous noise, \textsc{Homer} distinguishes between two transitions that represent transition between the same latent endogenous states but different exogenous states. In our walk in the park example, even if the agent moves between same points in two transitions, the model maybe able to tell these transitions apart by looking at the position of ducks which may have different behaviour in the two transitions. This results in the \textsc{Homer} creating $\Ocal(|\Zcal|)$ many abstract states. We call this the \emph{under-abstraction} problem.

\paragraph{Limitation of Inverse Dynamics.}
Another common approach in empirical works is based on modeling the inverse dynamics of the system, such as the \textsc{ICM} module of~\citet{pathak2017curiosity}. In such approaches, we learn a
representation by using consecutive observations to predict the action that was taken between them. Such a representation can ignore all information that is not relevant for action prediction, which includes all exogenous/uncontrollable information. However, it can also ignore controllable information. This may result in a failure to
sufficiently explore the environment. In this sense, inverse dynamics approaches result in an \emph{over-abstraction} problem where observations from different endogenous states can be mapped to the same abstract state. The over-abstraction problem was described at~\cite{misra2020kinematic}, when the starting state is random. In~\pref{app: failure of IK with determinstic starting state} we show inverse dynamics may over-abstract when the initial starting state is deterministic.

\paragraph{Limitation of Bisimulation.}~\cite{zhang2020learning} proposed learning a bisimulation metric to learn a representation which is invariant to exogenous noise. Unfortunately, it is known that bisimulation metric cannot be learned in a sample-efficient manner~(\citet{modi2020sample}, Proposition B.1). Intuitively, when the reward is same everywhere, then bisimulation merges all states into a single abstract state. This creates exploration issue in sparse reward settings, since the agent can merge all states into a single abstract state until it receives a non-trivial reward. Hence, the learned abstract states provide no help in exploring the environment.



\paragraph{Bellman rank might depend on $\Sxcard$.} The Bellman rank was introduced in~\cite{jiang2017contextual} as a complexity measure for the learnability of an RL problem with function approximations. To date, most of the learnable RL problems have a small Bellman rank. However, we show in~\pref{app:failure of existing approaches} that Bellman rank for EX-BMDP can scale as $\Ocal(|\Sx|)$. This shows that EX-BMDP is a highly non-trivial setting as we don't even have sample-efficient algorithms that can solve it without being computationally-efficient.

In~\pref{app:failure of existing approaches} we also describe the failures of ${\tt FLAMBE}$~\citep{agarwal2020flambe}) and autoencoding based approaches~\citep{tang2017exploration}.



\section{Reinforcement Learning for EX-BMDPs}\label{sec:learning}
\begin{algorithm}[t]
\caption{$\algppe(\delta,\eta)$: \algppelong}
\label{alg:genik_path_elim}
\setstretch{1.25}
\begin{algorithmic}[1] 
\State Set $\Psi_1 = \{\bot\}$, stochasticity level $\eta\leq \frac{1}{4SH}$ \mycomment{$\bot$ denotes an empty path}
\For{$h=2,\ldots,H$}
\State Set $N = 16\rbr{|\Psi_{h-1}\circ A|}^2\log\rbr{\frac{|\Fcal||\Psi_{h-1}|AH}{\delta}}$ 
\State Collect a dataset $\Dcal$ of $N$ \emph{i.i.d.} tuples $(x, \pt)$ where $\pt \sim \unf(\Psi_{h-1}\circ \Acal)$ and $x \sim \PP(x_h \mid \pt)$. \label{line:explore}
\State Solve multi-class classification problem:
            $
            \hat{f}_{h} = \arg\max_{f \in \Fcal} \sum_{(x, \pt) \in \Dcal} \ln f(\textrm{idx}(\pt) \mid x)
            $.\label{line:predict}
\For{$1 \le i < j \le |\Psi_{h-1}\circ \Acal|$}\label{line:for lopp ppe}
        \State Calculate the path prediction gap:
            $
          \widehat{\Delta}(i, j) = \frac{1}{N}\sum_{(x, \pt) \in \Dcal} \abr{\hat{f}_h(i | x) - \hat{f}_h(j | x)}.
            $\label{line:gap}
        \State If $\widehat{\Delta}(i, j) \leq \frac{5/8}{|\Psi_{h-1}\circ \Acal|}$, then eliminate path $\pt$ with $\mathrm{idx}(\pt)=j$. \mycomment{$\pt_i$ and $\pt_j$ visit the same state} \label{line:elimination}
    \EndFor
\State $\Psi_h$ is defined as the set of all paths in $\Psi_{h-1} \circ \Acal$ that have not been eliminated in~\pref{line:elimination}.
\EndFor
\end{algorithmic}
\end{algorithm}

In this section, we present an algorithm \emph{Predictive Path Elimination} ($\algppe$) that we later show can provably solve any EX-BMDP with nearly deterministic dynamics and start state distribution of the endogenous state, while making no assumptions on the dynamics or start state distribution of the exogenous state (\pref{alg:genik_path_elim}). Before describing $\algppe$, we highlight that $\algppe$ can be thought of as a computationally-efficient and simpler alternative to Algorithm 4 of~\cite{du2019provably} who studied rich-observation setting without exogenous noise.\footnote{Alg. 4 has time complexity of $\Ocal(S^4A^4H)$ compared to $\Ocal(S^3A^3H)$  for $\algppe$. Furthermore, Alg. 4 requires an upper bound on $S$, whereas $\algppe$ is adaptive to it. Lastly, ~\cite{du2019provably} assumed deterministic setting while we provide a generalization to near-determinism.} Due to the apparent difference of the algorithms we keep the presentation of our algorithm fully self-contained. 

$\algppe$ performs iterations over the time steps $h \in \{2, \cdots, H\}$. In the $h^{th}$ iteration, it learns a policy cover $\Psi_h$ for time step $h$ containing open-loop policies. This is done by first augmenting the policy cover for previous time step by one step. Formally, we define $\pts_h = \Psi_{h-1} \circ \Acal = \{\pi \circ a \mid \pi \in \Psi_{h-1}, a \in \Acal\}$ where $\pi \circ a$ is an open-loop policy that follows $\pi$ till time step $h-1$ and then takes action $a$. Since we assume the transition dynamics to be near-deterministic, therefore, we know that there exists a policy cover for time step $h$ that is a subset of $\pts_h$ and whose size is equal to the number of reachable states at time step $h$. Further, as the transitions are near-deterministic, we refer to an open-loop policy as a path, as we can view the policy as tracing a path in the latent transition model. $\algppe$ works by eliminating paths in $\pts_h$ so that we are left with just a single path for each reachable state. This is done by collecting a dataset $\Dcal$ of tuples $(x, \pt)$ where $\pt$ is a uniformly sampled from $\pts_h$ and $x \sim \PP_h(x \mid \pt)$ (\pref{line:explore}). We train a classifier $\hat{f}_h$ using $\Dcal$ by predicting the index $\textrm{idx}(\pt)$ of the path $\pt$ from the observation $x$ (\pref{line:predict}). Index of paths in $\pts_h$ are computed with respect to $\pts_h$ and remain fixed throughout training. Intuitively, if $\hat{f}_h(i \mid x)$ is sufficiently large, then we can hope that the path $\pt_i$ visits the state $\edo(x)$. Further, we can view this prediction problem as learning a multistep inverse dynamics model since the open-loop policy contains information about all previous actions and not just the last action.
For every pair of paths in $\pts_h$, we first compute a path prediction gap $\widehat{\Delta}$(\pref{line:gap}). If the gap is too small, we show it implies that these paths reach the same endogenous state, hence we can eliminate a single redundant path from this pair (\pref{line:elimination}). 
Finally, $\Psi_h$ is defined as the set of all paths in $\pts_h$ which were not eliminated. $\algppe$ reduces RL to performing $H$ standard classification problems. Further, the algorithm is very simple and in practice requires just a single hyperparameter ($N$). We believe these properties will make it well-suited for many problems.

\paragraph{Recovering an endogenous state decoder.} We can recover a endogenous state decoder $\hatedoh$ for each time step $h \in \{2, 3, \cdots, H\}$ directly from $\hat{f}_h$ as shown below:
$$\hatedoh(x) = \min \left\{ i \mid \hat{f}_h(i \mid x) \ge \max_j \hat{f}_h(j \mid x) - \Ocal(\nicefrac{1}{|\pts_h|}), i \in \left[|\pts_h|\right] \right\}.$$ Intuitively, this assigns the observation to the path with smallest index that has the highest chance of visiting $x$, and therefore, $\edo(x)$. We are implicitly using the decoder for exploring, since we rely on using $\hat{f}_h$ for making planning decisions. We will evaluate the accuracy of this decoder in~\pref{sec: experiments}.

\paragraph{Recovering the latent transition dynamics.}  $\algppe$ can also be used to recover a latent endogenous transition dynamics. The direct way is to use the learned decoder $\hatedoh$ along with episodes collected by $\algppe$ during the course of training and do count-based estimation. However, for most problems, recovering an approximate deterministic transition dynamics suffices, which can be directly read from the path elimination data. We accomplish this by recovering a partition of paths in $\Psi_{h-1} \times \Acal$ where two paths in the same partition set are said to be \emph{merged} with each other. In the beginning, each path is only merged with itself. When we eliminate a path $\pt_j$ on comparison with $\pt_i$ in~\pref{line:elimination}, then all paths currently merged with $\pt_j$ get merged with $\pt_i$. We then define an abstract state space $\widehat{\Scal}_h$ for time step $h$ that contains an abstract state $j$ for each path $\pt_j \in \Psi_h$. Further, we recover a latent deterministic transition dynamics for time step $h-1$ as $\hat{T}_{h-1}: \widehat{\Scal}_{h-1} \times \Acal \rightarrow \widehat{\Scal}_h$ where we set $\hat{T}_{h-1}(i, a) = j$ if the path $\pt_j \in \Psi_h$ gets merged with path $\pt'_i \circ a \in \Psi_h$ where $\pt'_i \in \Psi_{h-1}$.

\paragraph{Learning a near optimal policy given a policy cover.} $\algppe$ runs in a reward-free setting. However, the recovered policy cover and dynamics can be directly used to optimize any given reward function with existing methods. 
If the reward function depends on the exogenous state then we can use the $\PSDP$ algorithm~\citep{bagnell2004policy} to learn a near-optimal policy. $\PSDP$ is a model-free dynamic programming method that only requires policy cover as input (see~\pref{app: plannning in EX-BMDP analysis PSDP} for details). However, if the reward function only depends on the endogenous state, then we can use a computationally cheaper value-iteration algorithm $\VI$ that uses the policy cover and recovered transition dynamics. $\VI$ is a model-based algorithm that estimates the reward for each state and action, and performs dynamic programming on the model  (see~\pref{app: plannning in EX-BMDP analysis VI} for details). In each case, the sample complexity of learning a near-optimal policy, given the output of the $\algppe$ algorithm, scales with the size of endogenous and not the exogenous state space or the size of observation space.

\section{Theoretical Analysis and Discussion}\label{sec: theoretical analysis}

We provide the main sample complexity guarantee for $\algppe$ as well as additional intuition for why it works. We analyze the algorithm in near-deterministic MDPs defined as follows: Two transition functions $T_1$ and $T_2$ are \emph{$\eta$-close} if for all $h\in [H],a\in \Acal, s\in \Scal_h$ it holds that $\norm{T_{1}(\cdot\mid s,a) - T_{2}(\cdot\mid s,a)}_1\leq \eta$. Analogously, two starting distribution $\mu_1$ and $\mu_2$ are $\eta$-close if $\norm{\mu_1(\cdot) - \mu_2(\cdot)}_1 \leq \eta$. We emphasize that near-deterministic dynamics are common in real-world applications like robotics. 

\begin{assumption}[Near deterministic endogenous dynamics]\label{assumption: near determistic}
We assume the endogenous dynamics is $\eta$-close to a deterministic model $(\mu_{D,\eta}, T_{D,\eta})$ where $\eta\leq 1/(4 \Sdcard H)$. 
\end{assumption}

We make a realizability assumption for the regression problem solved by~$\algppe$ (\pref{line:predict}). We assume that $\Fcal$ is expressive enough to represent the Bayes optimal classifier of the regression problems created by $\algppe$. 
\begin{assumption}[Realizability] \label{assum: realizability Fcal}
For any $h \in [H]$, and any set of paths $\pts \subseteq \Acal^h$ with $|\pts| \le SA$ and where $\Acal^h$ denotes the set of all paths of length $h$, there exists $f^\star_{\pts, h} \in \Fcal$ such that:
$
    f^\star_{\pts, h}({\tt idx}(\pt) \mid x) = \frac{\PP_h(\edo(x)) \mid \pt)}{\sum_{\pt' \in \pts} \PP_h(\edo(x)) \mid \pt')},
$ for all $\pt \in \pts$ and $x \in \Xcal$ with $\sum_{\pt' \in \pts} \PP_h(\edo(x)) \mid \pt') > 0$.
\end{assumption}

In this assumption, the function $f^\star_{\pts, h}(\pt \mid x)$ denotes the Bayes optimal classifier for the regression problem in the $h^{th}$ iteration when $\pts_h = \pts$. Realizability assumptions are common in theoretical analysis (e.g., \cite{misra2020kinematic}, \cite{agarwal2020flambe}). In practice, we use expressive neural networks to solve the regression problem, so we expect the realizability assumption to hold. Note that there are at most $A^{S(H+1)}$ Bayes classifiers for different prediction problems. However, this is acceptable since our guarantees will scale as $\ln|\Fcal|$ and, therefore, the function class $\Fcal$ can be exponentially large to accommodate all of them. 

We now state the formal sample complexity guarantees for $\algppe$ below.
\begin{restatable}[Sample Complexity]{theorem}{theoremPPRSampleComplexity}\label{thm: PPE sample complexity} Fix $\delta \in (0, 1)$.
Then, with probability greater than $1-\delta$, $\algppe$ returns a policy cover $\cbr{\Psi_h}_{h=1}^H$ such that for any $h \in [H]$, $\Psi_h$ is a $\eta H$-policy cover for time step $h$ and $|\Psi_h| \leq S$, which gives the total number of episodes used by $\algppe$ as $\Ocal\left(S^2 A^2 H\ln\frac{|\Fcal| SAH}{\delta}\right)$.
\end{restatable}

We defer the proof to~\pref{app: sample complexity DPCID}. Our sample complexity guarantees do not depend directly on the size of observation space or the exogenous space. Further, since our analysis only uses standard uniform convergence arguments, it extends straightforwardly to infinitely large function classes by replacing $\ln |\Fcal|$ with other suitable complexity measures such as Rademacher complexity. 

\paragraph{Why does $\algppe$ work?} We provide an asymptotic analysis to explain why $\algppe$ works. Consider a deterministic setting and the $h^{th}$ iteration of $\algppe$. Assume by induction that $\Psi_{h-1}$ is an exact policy cover for time step $h-1$. Let $\PP_h(\xi)$ denote the distribution over exogenous states at time step $h$ which is independent of agent's policy. The Bayes optimal classifier ($f^\star_h \defeq f_{\pts_h, h}$) of the prediction problem can be derived as:
{\small \begin{equation*}
    f^\star_h({\tt idx}(\pt) \mid x) \defeq \PP_h(\pt \mid x) = \frac{\PP_h(x \mid \pt) \PP(\pt)}{\sum_{\pt' \in \pts_h} \PP_h(x \mid \pt') \PP(\pt')} \stackrel{(a)}{=} \frac{\PP_h(x \mid \pt) }{\sum_{\pt' \in \pts_h} \PP_h(x \mid \pt')} \stackrel{(b)}{=} \frac{\PP_h(\edo(x)) \mid \pt)}{\sum_{\pt' \in \pts_h} \PP_h(\edo(x)) \mid \pt')},
\end{equation*}}
where $(a)$ holds since all paths in $\pts_h$ are chosen uniformly, and $(b)$ critically uses the fact that for any open-loop policy $\pt$ we have: 
\begin{equation}\label{eqn:factorization}
    \PP_h(x \mid \pt) = q\left(x \mid \edo(x), \exo(x)\right) \PP_h(\edo(x) \mid \pt) \PP_h(\exo(x)) \qquad\qquad \mbox{(Factorization Lemma).}
\end{equation}
As $\Psi_{h-1}$ is a policy cover for time step $h-1$, therefore, $\pts_h$ is also a policy cover for time step $h$.
This implies that we can hope to converge to $f^\star_h(\cdot \mid x)$ for every observation reachable at time step $h$. It also implies that the denominator $\omega(\edo(x)) \defeq \sum_{\pt' \in \pts_h}\PP_h(\edo(x) \mid \pt')$ will be non-zero for every observation $x$ reachable at time step $h$, and, therefore, $f^\star(\cdot \mid x)$ is well-defined.

The Bayes optimal classifier depends only on the endogenous state $\edo(x)$ of the observation $x$.
Therefore, there is no signal available to separate observations that map to the same endogenous state, and we don't have an under-abstraction issue. Conversely, let $x$ and $x'$ be two observations reachable at time step $h$ that map to two different endogenous state $s$ and $s'$ respectively. Let $\pt \in \pts_h$ be a path that visits $s$ and has an index $i$. Then $\pt$ cannot visit $s'$ due to deterministic dynamics. Using the structure of $f^\star_h$ derived above, we get $f^\star_h(i \mid x) = \nicefrac{1}{\omega(s)} > \nicefrac{1}{|\pts_h|}$ and $f^\star_h(i \mid x') = 0$. This provides the signal to separate observations from different endogenous states. Therefore, there is no over-abstraction issue.

$\algppe$ uses a similar reasoning to filter out redundant paths that reach the same endogenous state. This is necessary to keep the size of policy cover small and get polynomial sample complexity. Let $\pt_1, \pt_2 \in \pts_h$ be two paths with indices $i$ and $j$ respectively. We define their exact path prediction gap as $\Delta(i,j) \defeq \EE_{x_h}\sbr{\abr{f^\star_h(i \mid x_h) - f^\star_h(j \mid x_h)}}$. Let $\pt_1$ visit an endogenous state $s$ at time step $h$. Then $f^\star_h(i \mid x_h) = \nicefrac{1}{\omega(s)}$ if $\edo(x_h)=s$, and $0$ otherwise. If $\pt_2$ also visits $s$ at time step $h$, then $f^\star_h(i \mid x_h) = f^\star_h(j \mid x_h)$ for all $x_h$. This implies $\Delta(i, j) = 0$ and $\algppe$ will filter out the path with higher index. Conversely, let $\pt_2$ visit a different state at time step $h$. If $x$ is an observation that maps to $s$, then $f^\star_h(i \mid x) = \nicefrac{1}{\omega(s)}$ and $f^\star_h(j \mid x) = 0$. This gives $|f^\star_h(i \mid x) - f^\star_h(j \mid x)| = \nicefrac{1}{\omega(s)} \ge \nicefrac{1}{|\pts_h|}$ and, consequently, $\Delta(i, j) > 0$. In fact, we can show $\Delta(i, j) \ge \Ocal(\nicefrac{1}{|\pts_h|)}$. Thus, $\algppe$ will not eliminate these paths upon comparison. Our complete analysis in the Appendix generalizes the above reasoning to finite sample setting where we can only approximate $f^\star_h$ and $\Delta$, as well as to EX-BMDPs with near-deterministic dynamics.


We stress that our analysis critically relies on~\pref{eqn:factorization} that holds for open-loop policies but not for an arbitrary policy class. This is the main reason why we build a policy cover with open-loop policies.

\section{Experiments}\label{sec: experiments}

\begin{figure}%
\addtolength{\belowcaptionskip}{-0.5em}
    \centering
    \subfigure[][\centering Combination lock $(H=2)$.]{{\includegraphics[width=0.315\textwidth, height=0.21\textwidth]{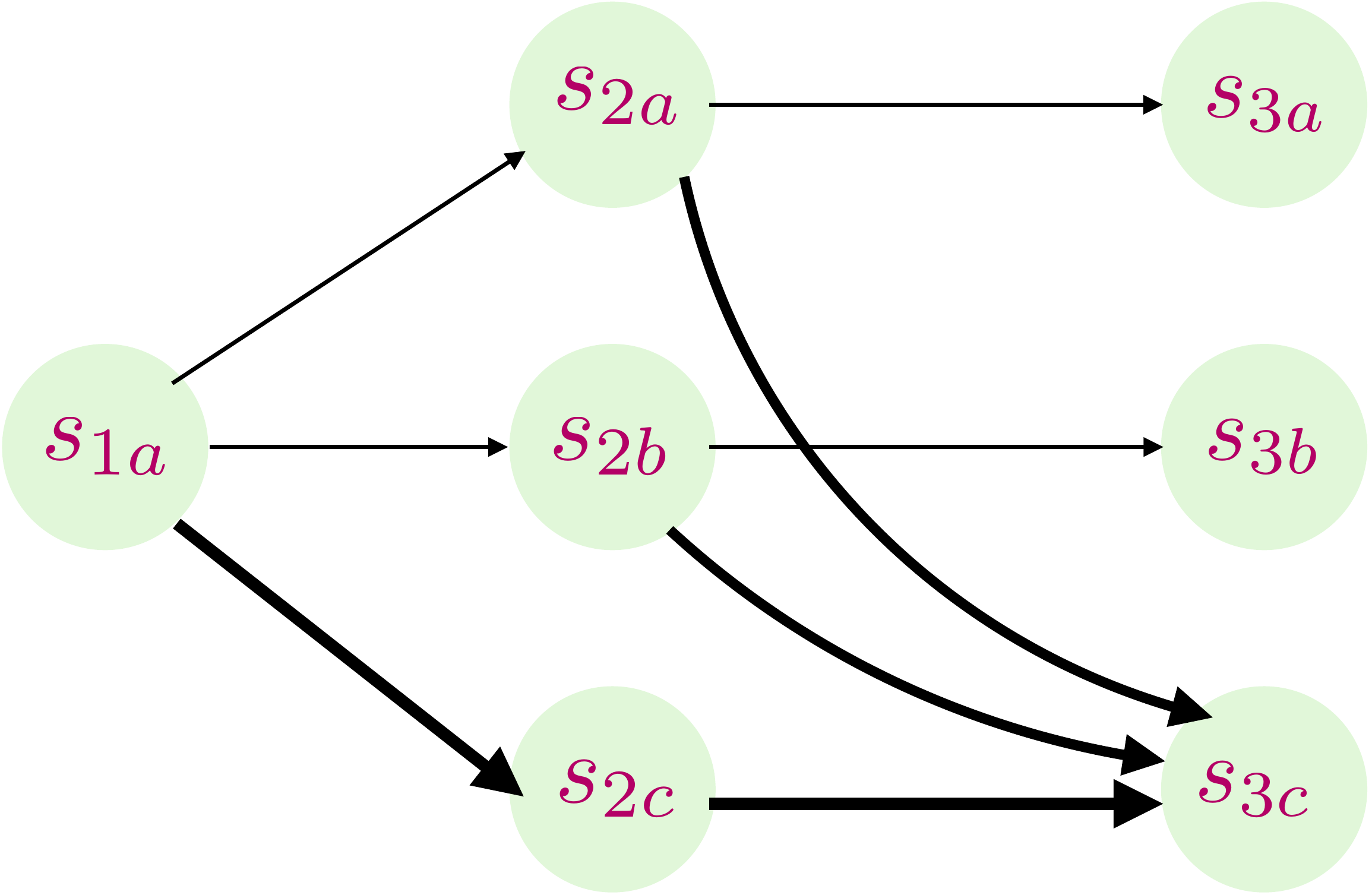}}}%
    \hspace{0.2cm}
    \subfigure[][\centering Regret plot]{{\includegraphics[width=0.29\textwidth]{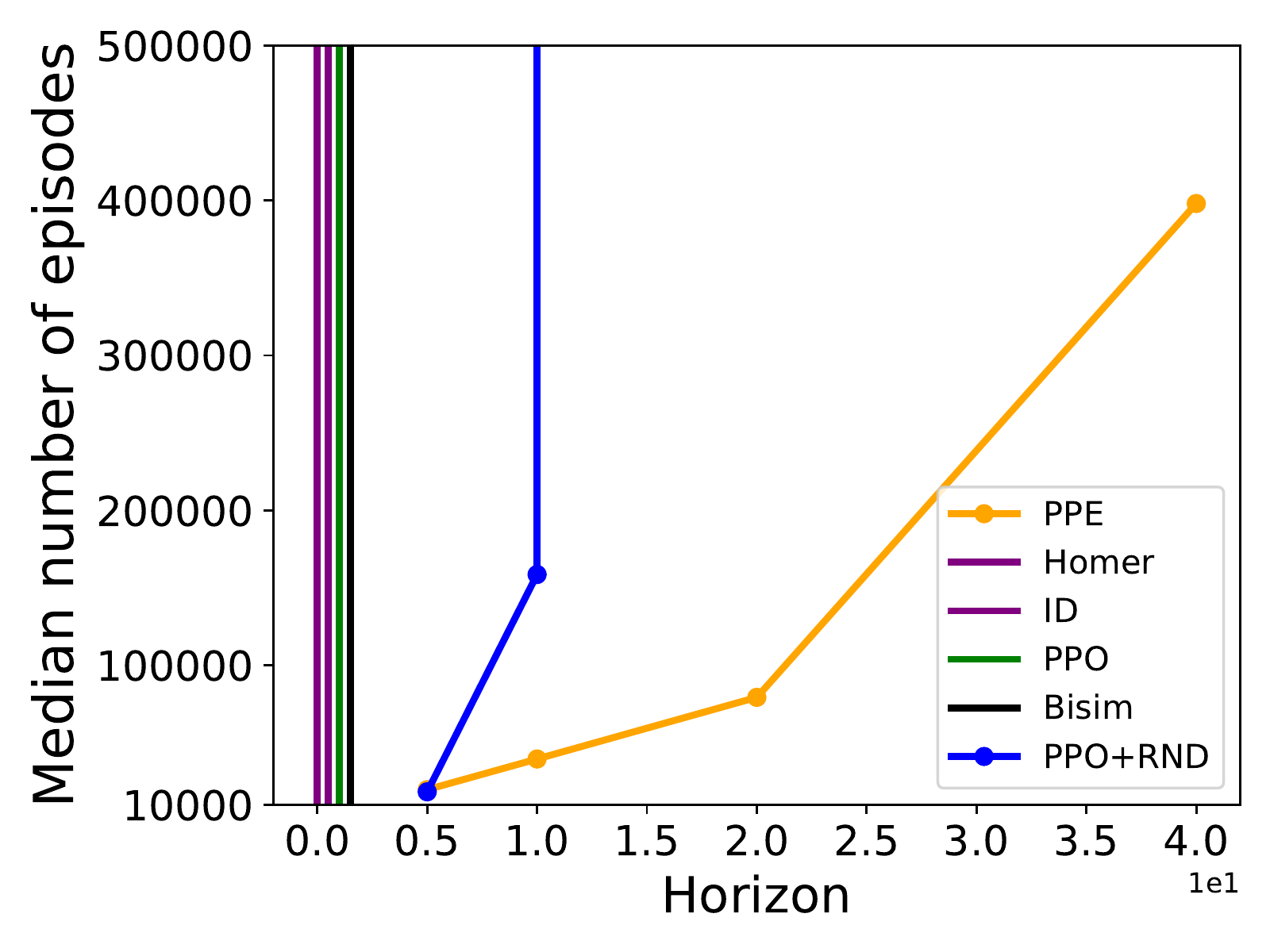} }}%
    \subfigure[][\centering Decoding accuracy]{{\includegraphics[width=0.29\textwidth]{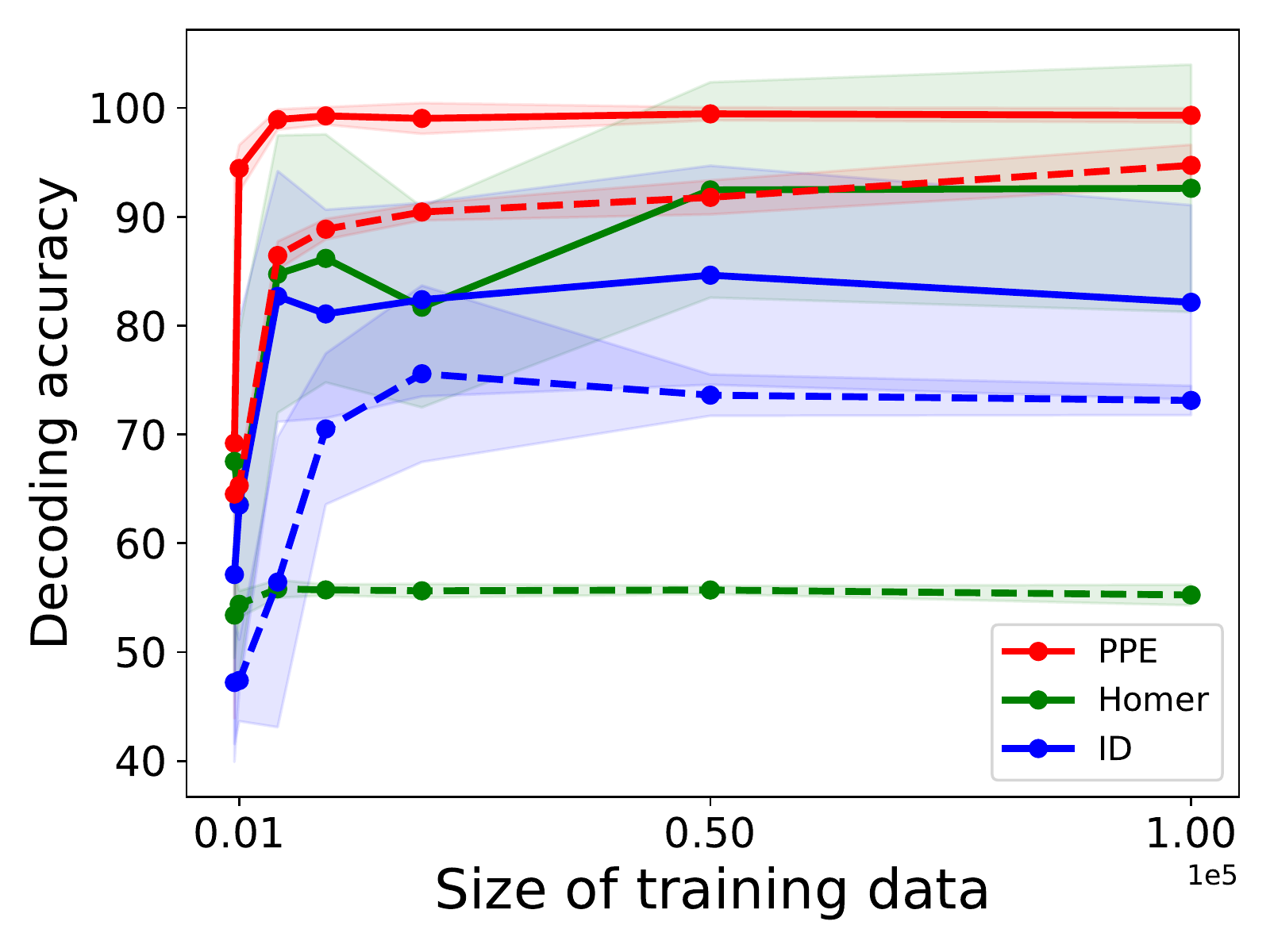}}}%
    \caption{ \footnotesize Results on combination lock. \textbf{ \footnotesize Left:} We show the latent transition dynamics of combination lock. Observations are not shown for brevity. \textbf{\footnotesize Center:} Shows minimal number of episodes needed to achieve a mean regret of at most $\nicefrac{V(\pi^\star)}{2}$. \textbf{\footnotesize Right:} State decoding accuracy (in percent) of decoders learned by different methods. Solid lines implies no exogenous dimension while dashed lines imply an exogenous dimension of 100.\vspace{-0.1cm}}%
    \label{fig:combolock-results}%
    \vspace{-0.1cm}
\end{figure}

We evaluate $\algppe$ on two domains: a challenging exploration problem called \emph{combination lock} to test whether $\algppe$ can learn an optimal policy and an accurate state decoder, and a visual-grid world with complex visual representations to test whether $\algppe$ is able to recover the latent dynamics. 

\paragraph{Combination Lock Experiments.} The combination lock problem is defined for a given horizon $H$
by an endogenous state space $\Sd = \{s_{1, a}\} \cup \{s_{h,a}, s_{h,b}, s_{h,c}\}_{h=2}^H$, an exogenous state space $\Sx = \{0, 1\}^H$, an action space $\Acal$ with 10 actions, and a deterministic endogenous start state of $s_{1,a}$. For any state $s_{h, g}$ we call $g$ as its \emph{type} which can be $a, b$ or $c$. States with type $a$ and $b$ are considered \emph{good} states and those with type $c$ are considered \emph{bad} states. Each instance of this problem is defined by two good action sequences $(a_h)_{h=2}^H, (a'_h)_{h=2}^H$ with $a_h \ne a'_h$, which are chosen uniformly randomly and kept fixed throughout training. At $h=1$, the agent is in $s_{1, a}$ and action $a_1$ leads to $s_{2, a}$, $a'_h$ leads to $s_{2, b}$, and all other actions lead to $s_{2, c}$. For $h>2$, taking action $a_h$ in $s_{h, a}$ leads to $s_{h+1, a}$ and taking action $a'_h$ in $s_{h, b}$ leads to $s_{h+1, b}$. In all other cases involving taking an action in a  state $s_{h, g}$, we transition to the next bad state $s_{h+1, c}$. We visualize the latent endogenous dynamics in~\pref{fig:combolock-results}a. The exogenous state evolves as follows: we sample $\sx_1$ as a vector in $\{0, 1\}^H$ by selecting the value of each dimension independently uniformly in $\{0, 1\}$. At time step $h$, $\sx_h$ is generated from $\sx_{h-1}$ by uniformly flipping each bit in $\sx_{h-1}$ independently with probability 0.1. There is a reward of 1.0 on taking the good action $a_{H,a}$ in $s_{H,a}$ and a reward of 0.1 on taking action $a_{H, b}$ in $s_{H, b}$, and in all other cases the agent gets a reward of 0. This gives a $V(\pi^\star) = 1$, and the probability that a random open loop policy gets this optimal return is $10^{-H}$. 

An observation $x$ is generated stochastically from a latent state $z = (\sd, \sx)$. We map $\sd$ to a vector $w$ encoding the identity of the state. We concatenate $(w, \sx)$, add Gaussian noise to each dimension, and multiply the result with a Hadamard matrix to generate $x$. See~\pref{app:exp} for full details. Our construction is inspired by similar combination lock problems in prior work~\citep{du2019provably,misra2020kinematic} (which did not consider exogenous distractors).

\paragraph{Baseline.} We compare $\algppe$ with five baselines on the combination lock problem. These include ${\tt PPO}$~\citep{schulman2017proximal} which is an actor-critic algorithm, ${\tt PPO+RND}$~\citep{burda2018exploration} which adds an exploration bonus to ${\tt PPO}$ using prediction errors, ${\tt Homer}$ that uses contrastive learning~\citep{misra2020kinematic}, and another algorithm ${\tt ID}$ which is similar to ${\tt Homer}$ but instead of contrastive learning it learns an inverse dynamics model to recover the state abstraction. Lastly, we also compare with ${\tt Bisim}$ that learns a bisimulation metric along with an actor-critic agent (\cite{zhang2020learning}). We use existing publicly available codebases for these baselines. Our implementation of $\algppe$ very closely follows the pseudo-code in~\pref{alg:genik_path_elim}. We model $\Fcal$ using a two-layer feed-forward network with ReLU non-linearity. We train $\Fcal$ with Adam optimization and use a validation set to do model selection. We refer readers to~\pref{app:exp} for additional experimental details.

\paragraph{Results.} \pref{fig:combolock-results}b shows results for values of $H$ in $\{5, 10, 20, 40\}$. For each value of $H$, we plot the minimal number of episodes $n$ needed to achieve a mean regret of at most $\nicefrac{V(\pi^\star)}{2} = 0.5$. We run each algorithm 5 times with different seeds and report the median performance. If an algorithm is unable to achieve the desired regret in $5\times 10^5$ episodes then we set $n=\infty$. We observe that ${\tt PPO}$ is unable to solve the problem at $H=5$. ${\tt PPO+RND}$ is able to solve the problem at $H=5$ and $H=10$, showing the exploration bonus induced by random network distillation helps. However, it is unable to solve the problem for larger values of $H$. We observe that ${\tt Homer}$ and ${\tt ID}$ are also unable to solve the problem for any value of $H$. ${\tt Bisim}$ also fails to solve the problem for any $H \ge 5$. This agrees with the theoretical prediction that ${\tt Bisim}$ provides no learning signal when running in sparse-reward settings. In such settings, there can be many episodes before any reward is received. In the absence of any reward, the bisimulation objective incentivizes mapping all observations to the same representation which is not helpful for further exploration. Lastly, $\algppe$ is able to solve the problem for all values of $H$ and is significantly more sample efficient than baselines. Since the reward function of the combination-lock problem depends only on the endogenous state, we run $\algppe$ and then a value-iteration like algorithm (see~\pref{app: plannning in EX-BMDP analysis VI}) to learn a near optimal policy. 

In order to understand the failure of ${\tt Homer}$ and ${\tt ID}$, we investigate the accuracy of the state abstraction learned by these methods and compare that with $\algppe$. We focus on the combination lock setting with $H=2$ and evaluate the learned decoder for the last time step. As the state abstraction models are invariant to label permutation we use the following evaluation metric: given a learned abstraction for the endogenous state $\hat{\phi}: \Xcal \rightarrow [N]$ we compute $\nicefrac{1}{m} \sum_{i=1}^m \one\{\hat{\phi}(x_{i,1}) = \hat{\phi}(x_{i,2}) \Leftrightarrow \edo(x_{i, 1}) = \edo(x_{i, 2})\}$, where $\{x_{i, 1}, x_{i, 2}\}_{i=1}^m$ are drawn independently from a fixed distribution $D$ with good support over all states. We report the percentage accuracy in~\pref{fig:combolock-results}c. When there is no exogenous noise, ${\tt Homer}$ is able to learn a good state decoder with enough samples while ${\tt ID}$ fails to learn, in accordance with the theory. On inspection, we found that ${\tt ID}$ suffers from the under-abstraction issue highlighted earlier as it has difficulty separating observations from $s_{3a}$ and $s_{3b}$. On adding exogenous noise, the accuracy of ${\tt Homer}$ plummets significantly. The accuracy of ${\tt ID}$ also drops but this drop is mild since unlike ${\tt Homer}$, the ${\tt ID}$ objective is able to filter exogenous noise. Lastly, we observe that $\algppe$ is always able to learn a good decoder and is more sample efficient than baselines.

\begin{figure}%
\addtolength{\belowcaptionskip}{-0.5em}
    \centering
    \subfigure[][\centering Grid World $h=1$.]{{\includegraphics[width=0.2\textwidth,height=0.2\textwidth]{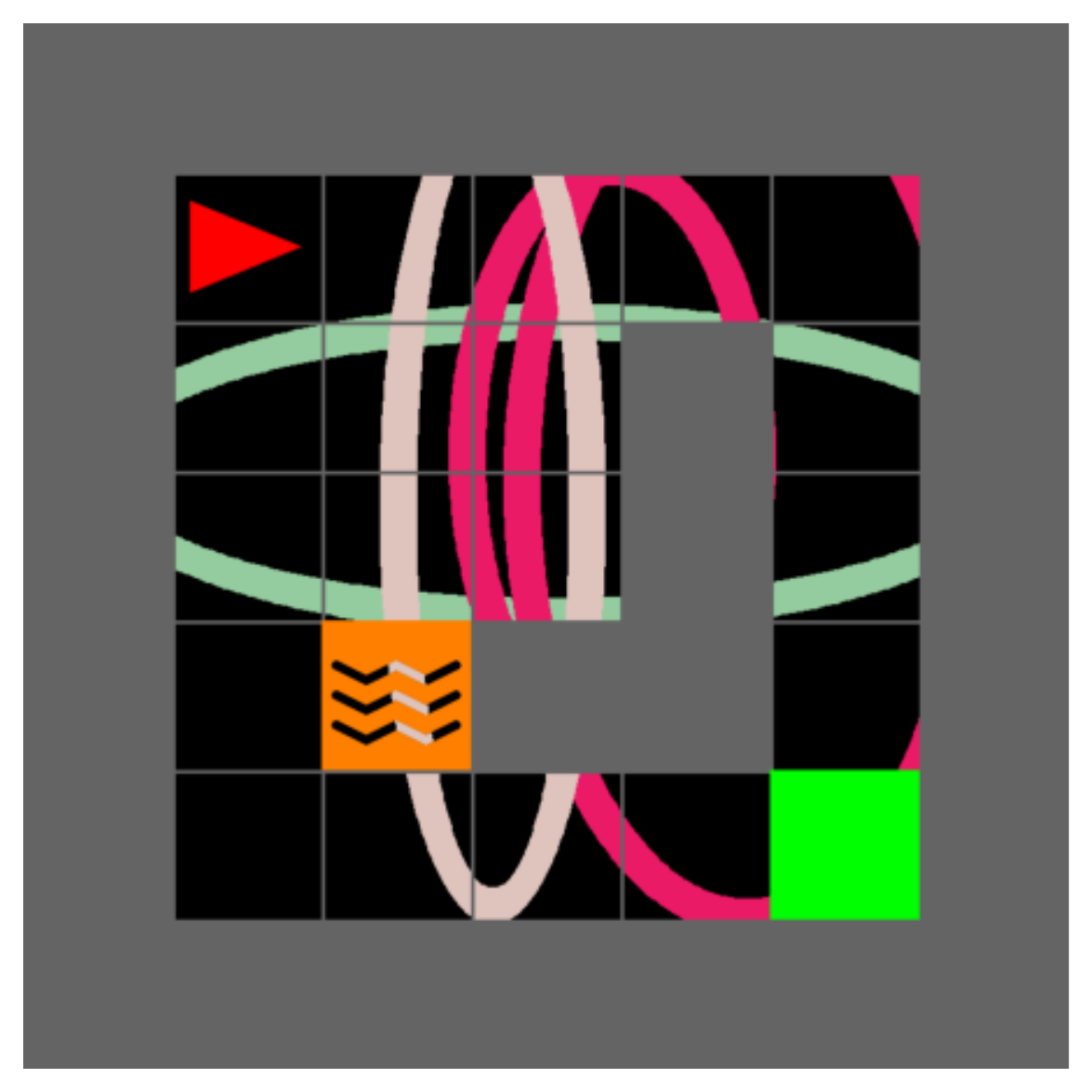}}}%
    \hspace{0.05cm}
    \subfigure[][\centering Grid World $h=2$.]{{\includegraphics[width=0.2\textwidth,height=0.2\textwidth]{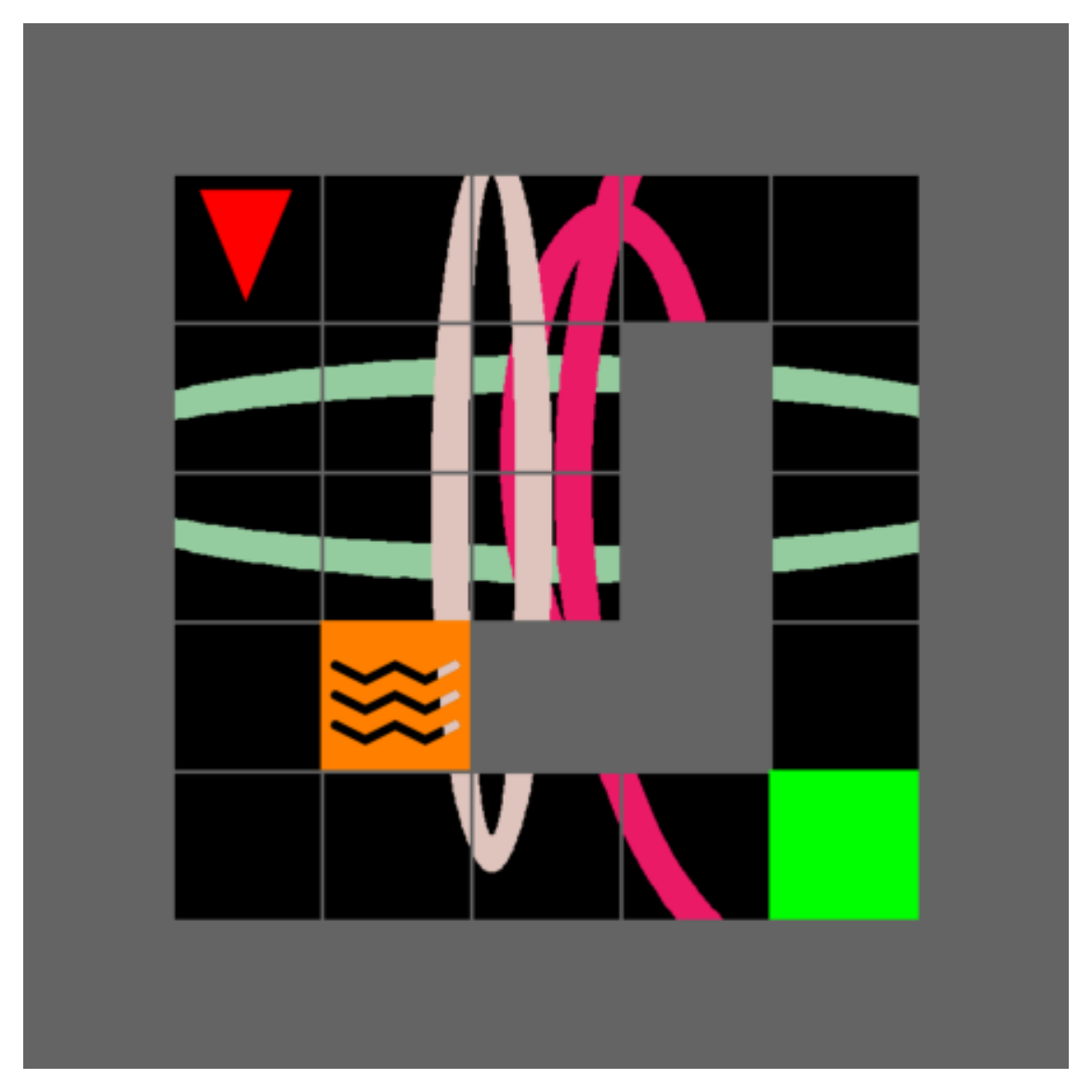}}}%
    \hspace{0.15cm}
    \subfigure[][\centering Model Errors]{{\includegraphics[width=0.27\textwidth]{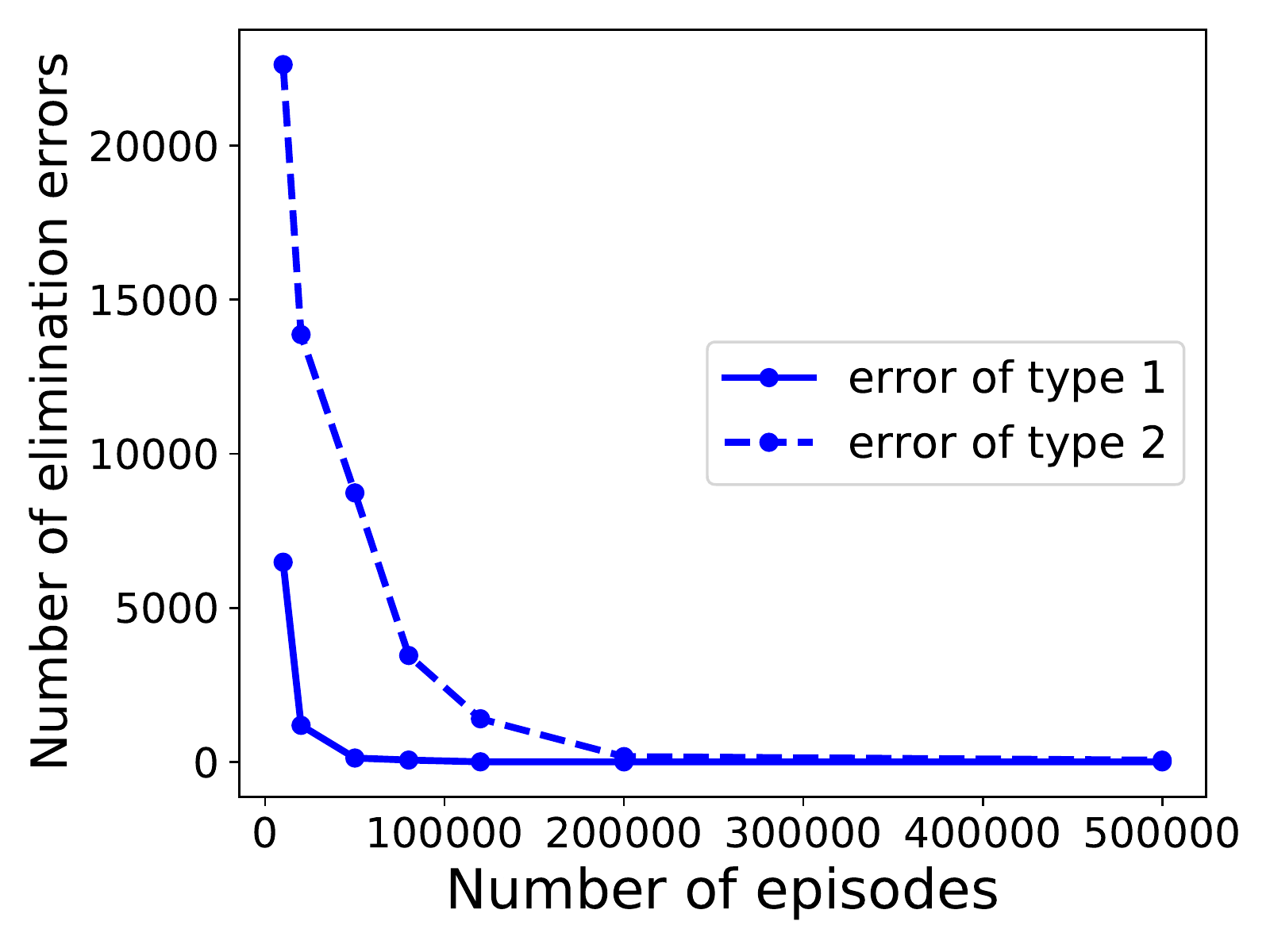} }}%
        \subfigure[][\centering State decoding accuracy]{{\includegraphics[width=0.27\textwidth]{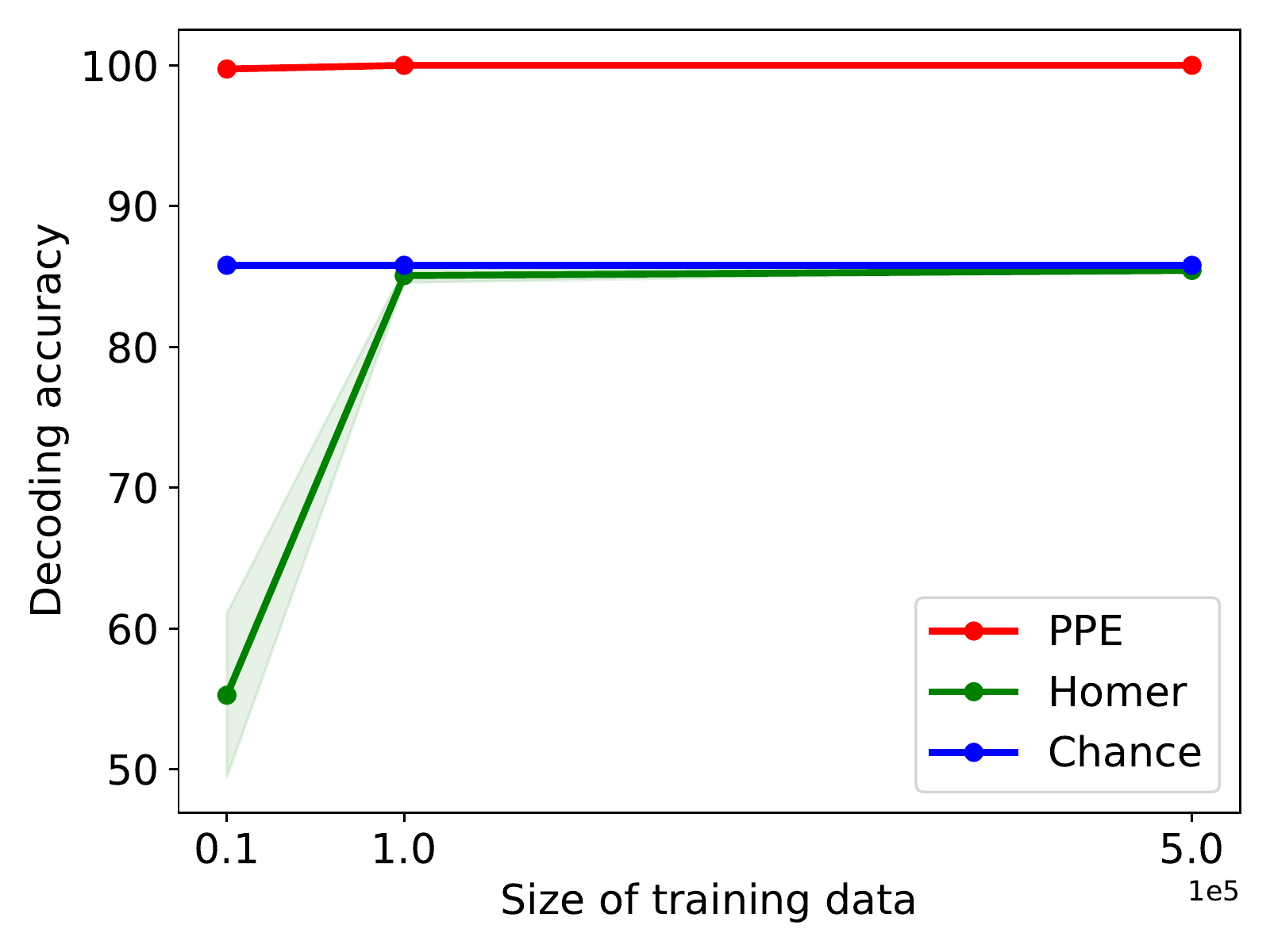} }}%
    \caption{ \footnotesize Results on visual grid world. \textbf{\footnotesize Left two:} Shows sampled observations for the first two steps from the visual gridworld domain. The agent is depicted as a red-triangle, lava in orange, walls in grey, and the goal in green. \textbf{\footnotesize Center Right:} Shows errors of type 1 and type 2 made by the $\algppe$ in recovering the latent endogenous dynamics. \textbf{\footnotesize Right:} State decoding accuracy of $\algppe$, ${\tt Homer}$ and a random uniform decoder. (see~\pref{sec: experiments})\vspace{-0.1cm}}%
    \label{fig:visual-gridworld-results}
\end{figure}

\paragraph{Visual Grid World Experiments.}
We test the ability of $\algppe$ to recover the latent endogenous transition dynamics in visual grid-world problem.\footnote{We use the following popular gridworld codebase: https://github.com/maximecb/gym-minigrid} 
The agent navigates in a $N \times N$ grid world where each grid can contain a stationary object, the goal, or the agent. The agent's endogenous state is given by its position in the grid and its direction amongst four possible canonical directions. The agent can take five different actions for navigation. The world is visible to the agent as a $8N \times 8N$ sized RGB image. We add exogenous noise as follows: at the beginning of each episode, we independently sample position, size and color of 5 ellipses. The position and size of these ellipses is perturbed after each time step independent of the action. We project these ellipses on top of the world's image.~\pref{fig:visual-gridworld-results} shows sampled observations from the $7\times 7$ gridworld that we experiment on. The exogenous state is given by the position, size and color of ellipses and is much larger than $|\Sd| \le 4N^2$. We model $\Fcal$ using a two-layer convolutional neural network and train it using Adam optimization. We defer the full details of setup to~\pref{app:exp}.

Since the problem has deterministic dynamics, we can evaluate the accuracy of the learned transition model by measuring it in terms of accuracy of the elimination step (\pref{alg:genik_path_elim}, \pref{line:elimination}), since this step induces our algorithm's mapping from observations to endogenous latent states. For a fixed $h \in \{2, \cdots, H\}$, let $\nu_i$ and $\nu_j$ be two paths in $\Psi_{h-1}\circ \Acal$. We compute two type of errors. Type 1 error computes whether $\algppe$ merged these paths, i.e., predicted them as mapping to the same abstract state, when they go to \emph{different endogenous} states. Type 2 error computes whether $\algppe$ predicted the paths as mapping to different abstract states, when they map to the \emph{same endogenous} state. We report the total number of errors of both types by summing over all values of $h$ and all pairs of different paths in $\Psi_{h-1} \circ \Acal$.  Type 1 errors are more harmful, since they can lead to exploration failure. Specifically, merging paths going to different states may result in the algorithm avoiding one of the two states when exploring at the next time step. Type 2 errors are less serious but lead to inefficiency due to using redundant paths for exploration. When both errors are 0, then the model recovers the exact latent model up to relabeling.

\paragraph{Results.} We report results on learning the model in in~\pref{fig:visual-gridworld-results}c. We see that $\algppe$ is able to reduce the number of type 1 errors down to 0 using $2\times 10^5$ episodes per time step. This is important since even a single type 1 error can cause exploration failures. Similarly, $\algppe$ is able to reduce type 2 errors and is able to get them down to 56 with $5 \times 10^5$ episodes. This is acceptable since type 2 errors do not cause exploration failures but only cause redundancy. Therefore, at $2 \times 10^5$ samples, the algorithm makes 0 type 1 errors and just a handful type 2 errors. This is remarkable considering that $\algppe$ compares roughly $2\times 10^5$ pairs of paths in the entire run. Hence, it makes only $\le 0.03\%$ type 2 errors. Further, the agent is able to plan using the learned transition model and receive the optimal return. We also evaluate the accuracy of state decoding on this problem. We compare the state decoding accuracy of $\algppe$ and ${\tt Homer}$ at $H=2$ using an identical evaluation setup to the one we used for combination lock.~\pref{fig:visual-gridworld-results}d shows the results. As expected, ${\tt PPE}$ rapidly learns a highly accurate decoder while ${\tt Homer}$ performs only as well as a random uniform decoder.

\section{Conclusion}\label{sec: summary}

In this work, we introduce the EX-BMDP setting, an RL setting that models exogenous noise, ubiquitous in many real-world systems. We show that many existing RL algorithms fail in the presence of exogenous noise. We present $\algppe$ that learns a multi-step inverse dynamics to filter exogenous noise and successfully explores. We derive theoretical guarantees for $\algppe$ in near-deterministic setting and provide  encouraging experimental evidence in support of our arguments. To our knowledge, this is the first such algorithm with guarantees for settings with exogenous noise. Our work also raises interesting future questions such as how to address the general setting with stochastic transitions, or handle more complex endogenous state representations. Another interesting line of future work direction is the analysis of other approaches that learn multi-step inverse dynamics ~\citep{gregor2016variational,paster2020planning} and understanding whether these approaches can also provably solve EX-BMDPs.

\section*{Acknowlegments}
We would like to thank the reviewers for their suggestions and comments.
We acknowledge the help of Microsoft's GCR team for helping with the compute. YE is partially supported by the Viterbi scholarship, Technion.

\bibliography{paper}

\begin{thebibliography}{30}
\providecommand{\natexlab}[1]{#1}
\providecommand{\url}[1]{\texttt{#1}}
\expandafter\ifx\csname urlstyle\endcsname\relax
  \providecommand{\doi}[1]{doi: #1}\else
  \providecommand{\doi}{doi: \begingroup \urlstyle{rm}\Url}\fi

\bibitem[Agarwal et~al.(2014)Agarwal, Hsu, Kale, Langford, Li, and
  Schapire]{agarwal2014taming}
Alekh Agarwal, Daniel Hsu, Satyen Kale, John Langford, Lihong Li, and Robert~E
  Schapire.
\newblock Taming the monster: A fast and simple algorithm for contextual
  bandits.
\newblock In \emph{International Conference on Machine Learning}, 2014.

\bibitem[Agarwal et~al.(2020{\natexlab{a}})Agarwal, Kakade, Krishnamurthy, and
  Sun]{agarwal2020flambe}
Alekh Agarwal, Sham Kakade, Akshay Krishnamurthy, and Wen Sun.
\newblock Flambe: Structural complexity and representation learning of low rank
  mdps.
\newblock \emph{Advances in Neural Information Processing Systems},
  2020{\natexlab{a}}.

\bibitem[Agarwal et~al.(2020{\natexlab{b}})Agarwal, Kakade, Lee, and
  Mahajan]{agarwal2020optimality}
Alekh Agarwal, Sham~M Kakade, Jason~D Lee, and Gaurav Mahajan.
\newblock Optimality and approximation with policy gradient methods in markov
  decision processes.
\newblock In \emph{Conference on Learning Theory}, 2020{\natexlab{b}}.

\bibitem[Antos et~al.(2008)Antos, Szepesv{\'a}ri, and Munos]{antos2008learning}
Andr{\'a}s Antos, Csaba Szepesv{\'a}ri, and R{\'e}mi Munos.
\newblock Learning near-optimal policies with bellman-residual minimization
  based fitted policy iteration and a single sample path.
\newblock \emph{Machine Learning}, 2008.

\bibitem[Bagnell et~al.(2004)Bagnell, Kakade, Schneider, and
  Ng]{bagnell2004policy}
J~Andrew Bagnell, Sham~M Kakade, Jeff~G Schneider, and Andrew~Y Ng.
\newblock Policy search by dynamic programming.
\newblock In \emph{Advances in Neural Information Processing Systems}, 2004.

\bibitem[Burda et~al.(2018)Burda, Edwards, Pathak, Storkey, Darrell, and
  Efros]{burda2018large}
Yuri Burda, Harri Edwards, Deepak Pathak, Amos Storkey, Trevor Darrell, and
  Alexei~A Efros.
\newblock Large-scale study of curiosity-driven learning.
\newblock In \emph{International Conference on Learning Representations}, 2018.

\bibitem[Burda et~al.(2019)Burda, Edwards, Storkey, and
  Klimov]{burda2018exploration}
Yuri Burda, Harrison Edwards, Amos Storkey, and Oleg Klimov.
\newblock Exploration by random network distillation.
\newblock In \emph{International Conference on Learning Representations}, 2019.

\bibitem[Dann et~al.(2017)Dann, Lattimore, and Brunskill]{dann2017unifying}
Christoph Dann, Tor Lattimore, and Emma Brunskill.
\newblock Unifying {PAC} and regret: Uniform {PAC} bounds for episodic
  reinforcement learning.
\newblock In \emph{Advances in Neural Information Processing Systems}, 2017.

\bibitem[Dann et~al.(2018)Dann, Jiang, Krishnamurthy, Agarwal, Langford, and
  Schapire]{dann2018oracle}
Christoph Dann, Nan Jiang, Akshay Krishnamurthy, Alekh Agarwal, John Langford,
  and Robert~E Schapire.
\newblock On oracle-efficient {PAC} {RL} with rich observations.
\newblock In \emph{Advances in Neural Information Processing Systems}, 2018.

\bibitem[Dietterich et~al.(2018)Dietterich, Trimponias, and
  Chen]{dietterich2018discovering}
Thomas~G Dietterich, George Trimponias, and Zhitang Chen.
\newblock Discovering and removing exogenous state variables and rewards for
  reinforcement learning.
\newblock \emph{arXiv preprint arXiv:1806.01584}, 2018.

\bibitem[Du et~al.(2019)Du, Krishnamurthy, Jiang, Agarwal, Dud{\'\i}k, and
  Langford]{du2019provably}
Simon~S Du, Akshay Krishnamurthy, Nan Jiang, Alekh Agarwal, Miroslav
  Dud{\'\i}k, and John Langford.
\newblock Provably efficient {RL} with rich observations via latent state
  decoding.
\newblock In \emph{International Conference on Machine Learning}, 2019.

\bibitem[Efroni et~al.(2021)Efroni, Merlis, and
  Mannor]{efroni2021reinforcement}
Yonathan Efroni, Nadav Merlis, and Shie Mannor.
\newblock Reinforcement learning with trajectory feedback.
\newblock In \emph{Proceedings of the AAAI Conference on Artificial
  Intelligence}, 2021.

\bibitem[Gelada et~al.(2019)Gelada, Kumar, Buckman, Nachum, and
  Bellemare]{gelada2019deepmdp}
Carles Gelada, Saurabh Kumar, Jacob Buckman, Ofir Nachum, and Marc~G Bellemare.
\newblock Deepmdp: Learning continuous latent space models for representation
  learning.
\newblock In \emph{International Conference on Machine Learning}, 2019.

\bibitem[Givan et~al.(2003)Givan, Dean, and Greig]{givan2003equivalence}
Robert Givan, Thomas Dean, and Matthew Greig.
\newblock Equivalence notions and model minimization in markov decision
  processes.
\newblock \emph{Artificial Intelligence}, 2003.

\bibitem[Gregor et~al.(2016)Gregor, Rezende, and
  Wierstra]{gregor2016variational}
Karol Gregor, Danilo~Jimenez Rezende, and Daan Wierstra.
\newblock Variational intrinsic control.
\newblock \emph{arXiv preprint arXiv:1611.07507}, 2016.

\bibitem[Hafner et~al.(2019)Hafner, Lillicrap, Fischer, Villegas, Ha, Lee, and
  Davidson]{hafner2019learning}
Danijar Hafner, Timothy Lillicrap, Ian Fischer, Ruben Villegas, David Ha,
  Honglak Lee, and James Davidson.
\newblock Learning latent dynamics for planning from pixels.
\newblock In \emph{International Conference on Machine Learning}, 2019.

\bibitem[Jiang et~al.(2017)Jiang, Krishnamurthy, Agarwal, Langford, and
  Schapire]{jiang2017contextual}
Nan Jiang, Akshay Krishnamurthy, Alekh Agarwal, John Langford, and Robert~E
  Schapire.
\newblock Contextual decision processes with low {B}ellman rank are
  {PAC}-learnable.
\newblock In \emph{International Conference on Machine Learning}, 2017.

\bibitem[Kakade and Langford(2002)]{kakade2002approximately}
Sham~M Kakade and John Langford.
\newblock Approximately optimal approximate reinforcement learning.
\newblock In \emph{International Conference on Machine Learning}, 2002.

\bibitem[Langford and Zhang(2008)]{langford2008epoch}
John Langford and Tong Zhang.
\newblock The epoch-greedy algorithm for multi-armed bandits with side
  information.
\newblock In \emph{Advances in Neural Information Processing Systems}, 2008.

\bibitem[Laskin et~al.(2020)Laskin, Srinivas, and Abbeel]{laskin2020curl}
Michael Laskin, Aravind Srinivas, and Pieter Abbeel.
\newblock Curl: Contrastive unsupervised representations for reinforcement
  learning.
\newblock In \emph{International Conference on Machine Learning}. PMLR, 2020.

\bibitem[Misra et~al.(2020)Misra, Henaff, Krishnamurthy, and
  Langford]{misra2020kinematic}
Dipendra Misra, Mikael Henaff, Akshay Krishnamurthy, and John Langford.
\newblock Kinematic state abstraction and provably efficient rich-observation
  reinforcement learning.
\newblock In \emph{International conference on machine learning}, pages
  6961--6971. PMLR, 2020.

\bibitem[Modi et~al.(2020)Modi, Jiang, Tewari, and Singh]{modi2020sample}
Aditya Modi, Nan Jiang, Ambuj Tewari, and Satinder Singh.
\newblock Sample complexity of reinforcement learning using linearly combined
  model ensembles.
\newblock In \emph{International Conference on Artificial Intelligence and
  Statistics}. PMLR, 2020.

\bibitem[Paster et~al.(2020)Paster, McIlraith, and Ba]{paster2020planning}
Keiran Paster, Sheila~A McIlraith, and Jimmy Ba.
\newblock Planning from pixels using inverse dynamics models.
\newblock In \emph{International Conference on Learning Representations}, 2020.

\bibitem[Pathak et~al.(2017)Pathak, Agrawal, Efros, and
  Darrell]{pathak2017curiosity}
Deepak Pathak, Pulkit Agrawal, Alexei~A Efros, and Trevor Darrell.
\newblock Curiosity-driven exploration by self-supervised prediction.
\newblock In \emph{International Conference on Machine Learning}, 2017.

\bibitem[Rosenberg and Mansour(2019)]{rosenberg2019online}
Aviv Rosenberg and Yishay Mansour.
\newblock Online convex optimization in adversarial markov decision processes.
\newblock In \emph{International Conference on Machine Learning}, 2019.

\bibitem[Schulman et~al.(2017)Schulman, Wolski, Dhariwal, Radford, and
  Klimov]{schulman2017proximal}
John Schulman, Filip Wolski, Prafulla Dhariwal, Alec Radford, and Oleg Klimov.
\newblock Proximal policy optimization algorithms.
\newblock \emph{arXiv:1707.06347}, 2017.

\bibitem[Shani et~al.(2020)Shani, Efroni, and Mannor]{shani2020adaptive}
Lior Shani, Yonathan Efroni, and Shie Mannor.
\newblock Adaptive trust region policy optimization: Global convergence and
  faster rates for regularized mdps.
\newblock In \emph{Proceedings of the AAAI Conference on Artificial
  Intelligence}, 2020.

\bibitem[Tang et~al.(2017)Tang, Houthooft, Foote, Stooke, Chen, Duan, Schulman,
  DeTurck, and Abbeel]{tang2017exploration}
Haoran Tang, Rein Houthooft, Davis Foote, Adam Stooke, OpenAI~Xi Chen, Yan
  Duan, John Schulman, Filip DeTurck, and Pieter Abbeel.
\newblock \#{E}xploration: A study of count-based exploration for deep
  reinforcement learning.
\newblock In \emph{Advances in Neural Information Processing Systems}, 2017.

\bibitem[Zhang et~al.(2020)Zhang, McAllister, Calandra, Gal, and
  Levine]{zhang2020learning}
Amy Zhang, Rowan McAllister, Roberto Calandra, Yarin Gal, and Sergey Levine.
\newblock Learning invariant representations for reinforcement learning without
  reconstruction.
\newblock \emph{arXiv preprint arXiv:2006.10742}, 2020.

\bibitem[Zhang et~al.(2021)Zhang, McAllister, Calandra, Gal, and
  Levine]{zhang2021learning}
Amy Zhang, Rowan~Thomas McAllister, Roberto Calandra, Yarin Gal, and Sergey
  Levine.
\newblock Learning invariant representations for reinforcement learning without
  reconstruction.
\newblock In \emph{International Conference on Learning Representations}, 2021.
\newblock URL \url{https://openreview.net/forum?id=-2FCwDKRREu}.

\end{thebibliography}
\bibliographystyle{plainnat}

\appendix

\newpage
\appendix 
\section*{Appendix}
We present the main notations in~\pref{tab:notations}. The rest of the Appendix is organized as follows:
\begin{enumerate}
    \item  In \pref{app:failure of existing approaches} we establish failure cases of several popular approaches for learning an optimal policy in the EX-BMDP model. 
    \item In \pref{app: structural results of EX-BMDP} we study structural properties of an EX-BMDP which highlight some of its key features.
    \item  In \pref{app: sample complexity DPCID} we prove \pref{thm: PPE sample complexity} our main performance guarantee on $\algppe$. There, we show that $\algppe$ returns an approximate policy cover with cardinality that is bounded by the size of the endogenous state space. Furthermore, the sample complexity of $\algppe$ does not depend on the cardinality of the exogenous state space at all.
    \item In \pref{app: plannning in EX-BMDP analysis} we analyze two planning approaches that utilize the output of $\algppe$ to find a near optimal policy when there exists an access to a reward function. We consider both the cases that the reward function is general, and a reward function that depends on the endogenous state.
    \item In \pref{app: existing results} we supply with several existing results that are used throughout the analysis.
    \item In \pref{app:exp} we describe in further detalis the experimental setting and supply with additional experiments.
\end{enumerate}
\begin{table*}[h]
    \centering
    \begin{tabular}{l|l}
    \hline
        \textbf{Notation} & \textbf{Meaning} \\
        \hline 
        $\Xcal$ & Countable observation space. Potentially infinite\\
        $\Sd$ & Finite endogenous state space. Assumed to be finite\\
        $\Sx$ & Countable exogenous state space. Potentially infinite\\
        $\Sf$ & State space given by $\Sf = \Sd \times \Sx$. Potentially infinite.\\
        $\Acal$ & Finite action space\\
        $q$ & Emission function $\Scal \rightarrow \Delta(\Xcal)$\\
        $T$ & Transition function $\Xcal \times \Acal \rightarrow \Delta(\Xcal)$ \\
        $H$ & Horizon \\
        $\ps$ & Maps observation to latent state $\Xcal \rightarrow \Sf$\\
        $\edo$ & Maps observation to endogenous state $\Xcal \rightarrow \Sd$\\
        $\exo$ & Maps observation to exogenous state $\Xcal \rightarrow \Sx$\\
        $\mu$ & start state distribution (overload $\mu(x) = q(x \mid \ps(x))\mu(\ps(x))$).\\
        $\Pi$ & Finite policy class \\
        $\Fcal$ & Finite function class $\Xcal \times [N] \rightarrow \mathbb{R}$\\
        $h$ & Indicates a time step $h \in [H]$ \\
        $\Mcal_D$ & An $\eta$ close deterministic MDP of the endogenous dynamics \\
        $T_D$ & Transition function of the $\eta$ close deterministic MDP of the endogenous dynamics \\
        $T_\xi$ & Transition function of the exogenous state space \\
        $\Psi_h$ & A policy cover of $h$ step policies\\
        $\pi(1: h)$ & The first $h$ step policies of a non-stationary policy of length greater than $h$ \\
        $\pi_h$ & The policy at the $h^{th}$ time step of a non-stationary policy\\
        $\Psi_h\circ\Acal$ & Extended policy cover of $\Psi_h$, $\Psi_h\circ\Acal = \cbr{\pi \textit{ an $h+1$ step policy }: \pi(1: h)\in\Psi_h, \pi_h\in \Acal}$ \\
        \hline
    \end{tabular}
    \caption{Notations used in the paper. We start indexing from 1. An episode is given by $(s_1, a_1, s_2, a_2, \cdots, s_H, a_H, s_{H+1})$}
    \label{tab:notations}
\end{table*}

\section{Failure of Existing Approaches in the Presence of Exogenous Noise}
\label{app:failure of existing approaches}

\subsection{Failure of Constrastive Learning}\label{app: failure of contrastive learning}

In this section we formally show that the objective defined in~\cite{misra2020kinematic} separates states according to the exogenous part. That is, states that share the same endogenous state, but have different exogenous state will be separated by the Backward Kinematic Inseparability criterion (BKI) on which the objective of HOMER relies upon (see \cite{misra2020kinematic}, Definition 3). Thus, the abstraction learned by HOMER will have the cardinality of $|\Scal|\times |\Xi|$ and will scale with the number of exogenous states.

Recall the definition of BKI given in~\cite{misra2020kinematic}.
\begin{definition}[Backward Kinematic Inseparability]\label{def: BKI}
Two states $(s'_1,\xi'_1),(s'_2,\xi'_2)\in \Zcal$ are backward kinematically inseparable if for all distributions $u\in \Delta(\Zcal\times \Acal)$ supported on $\Zcal\times \Acal$ and for all $z=(s,\xi)\in \Zcal,a\in \Acal$ we have
\begin{align*}
    \frac{T(s'_1,\xi'_1|s,\xi,a)u(s,\xi,a)}{\sum_{\tilde{s},\tilde{\xi},\tilde{a}} T(s'_1,\xi'_1| \bar{s},\bar{\xi},\bar{a})u(\bar{s},\bar{\xi},\bar{a})} = \frac{T(s'_2,\xi'_2|s,\xi,a)u(s,\xi,a)}{\sum_{\tilde{s},\tilde{\xi},\tilde{a}} T(s'_1,\xi'_1| \bar{s},\bar{\xi},\bar{a})u(\bar{s},\bar{\xi},\bar{a})}.
\end{align*}
\end{definition}
The BKI criterion unifies states if they cannot be differentiated w.r.t. any sampling distribution $u$ over the previous time step.

\begin{claim}\label{claim: BKI splits states with similar endo states}
BKI can splits states with similar endogenous states
\end{claim}
This claim is quite generic as we show below - for a very generic class of MDPs BKI splits states with similar endogenous state. Specifically, this occurs when the exogenous process is deterministic.
\begin{proof}
Consider any endogenous dynamics and a deterministic exogenous dynamics, i.e., $T(\xi'|\xi) = \ind\cbr{\xi'=\xi}$. Let $z'_1=(s',\xi_1')$ and $z_2'=(s',\xi'_2)$ be two states with similar endogenous dynamics and different exogenous dynamics. Then, BKI splits $z'_1$ and $z'_2$, i.e., it treats them as separate states.

Indeed, in this case, by the EX-BMDP model assumptions, it holds that
\begin{align*}
    &\frac{T(s',\xi'_1|s,\xi,a)u(s,\xi,a)}{\sum_{\tilde{s},\tilde{\xi},\tilde{a}} T(s',\xi'_1| \bar{s},\bar{\xi},\bar{a})u(\bar{s},\bar{\xi},\bar{a})} = \frac{T(s'_1|s,a)\ind\cbr{\xi'_1 = \xi}u(s,\xi,a)}{\sum_{\tilde{s},\tilde{\xi},\tilde{a}} T(s'| \bar{s},\bar{a})\ind\cbr{\xi'_1 = \xi}u(\bar{s},\bar{\xi},\bar{a})}\\
    &\neq \frac{T(s',\xi'_2|s,\xi,a)u(s,\xi,a)}{\sum_{\tilde{s},\tilde{\xi},\tilde{a}} T(s',\xi'_2| \bar{s},\bar{\xi},\bar{a})u(\bar{s},\bar{\xi},\bar{a})} = \frac{T(s'|s,a)\ind\cbr{\xi'_2 = \xi}u(s,\xi,a)}{\sum_{\tilde{s},\tilde{\xi},\tilde{a}} T(s'| \bar{s},\bar{a}) \ind\cbr{\xi_2'=\xi}u(\bar{s},\bar{\xi},\bar{a})} 
\end{align*}
where the inequality holds since $\xi'_1\neq \xi'_2$ (o.w. $z_1'=z_2'$).
\end{proof}

\subsection{Failure of Bisimulation Metric}\label{app: failure of bisumulation metric}

Recall the definition of the bisimulation relations~\cite{givan2003equivalence} (also given in~\cite{zhang2020learning}, Definition 1).
\begin{definition}[Bisimulation Relations]\label{def: bismulation relations}
Given an MDP $\mathcal{M},$ an equivalence relation $B$ between states is a bisimulation relation if for all states $s_i,s_j\in \Scal$ that are equivalent under $B$ it holds that
\begin{enumerate}
    \item $r(s_i,a) = r(s_j,a)\quad \forall a\in \Acal$,
    \item $\PP(G| s_i,a) = \PP(G|s_j,a)\quad \forall a\in \Acal,G\in \Scal_B$
\end{enumerate}
where $\Scal_B$ is the partition of $\Scal$ under the relation $B$ and $\PP(G|s,a) = \sum_{s'} T(s'|s,a)$.
\end{definition}

Let $d$ be some distribution over $\Scal$. We say that an equivalence relation $B$ is a $d$-restricted bisimulation relation if \pref{def: bismulation relations} holds for all $s\in \Scal$ such that $d(s)>0$. That is, if $B$ is a bisimulation for all states in the support of $d$. Indeed, we have no information on states that are not in the support of $d$. For this reason, we cannot obtain any information on these states.

\begin{claim}[With no reward function bismulation may unify all states]
Assume that $r(s,a)=0$ for all states $s\in \Scal$ for which $d(s)>0$, i.e., in the support of $d$. Then, the abstraction $\phi(s)=1$ (i.e., merge all states into a single state) for all $s\in \Scal$ is a valid $d$-restricted bisumlation relation.
\end{claim}
\begin{proof}
We show this abstraction is a valid $d$-restricted relation. 
\begin{enumerate}
    \item For all states restricted to $d$, meaning $s\in \cbr{s\in \Scal: d(s)>0}$ it holds that $r(s,a)=0$ for all $a\in \Acal$. Thus, the first requirement of \pref{def: bismulation relations} is satisfied.
    \item When all states are merges it holds that $G=\Omega$. Thus, $\PP(G|s_i,a) = \PP(G| s_j,a)=1$, and the second requirement of \pref{def: bismulation relations} is satisfied.
\end{enumerate}
\end{proof}

\subsection{Failure of Inverse Dynamics}\label{app: failure of IK with determinstic starting state}
We describe a totally deterministic setting where Inverse Dynamics (ID) fails. We comment that the counter-example for ID supplied by \cite{misra2020kinematic} has stochastic starting state. In this section, we show that even for a deterministic MDP with deterministic starting state ID fails. Specifically, we assume access to an exact solution $f_\star$ of a regression oracle that learns ID. Then, we show that a naive approach that uses $f_\star$ (see \pref{alg:id naive approach}) fails to return a policy cover.

\begin{algorithm}[t]
\caption{Exact Inverse Dynamics for Deterministic Initial State}
\label{alg:id naive approach}
\setstretch{1.25}
\begin{algorithmic}[1]
\State $\Psi_{1}=\emptyset$
\For{$h=1,\ldots,H-1$}
    \State Set the sampling distribution $\mu$ to be $\unf(\Psi_h)$\footnote{If $\Psi_h$ is an empty set then $x$ is a fixed deterministic state}
    \State Get $f_\star$ that exactly minimizes the loss
    \begin{align*}
        \EE_{x\sim\mu,a\sim \unf(\Acal),x'\sim T(\cdot|x,a)}\sbr{ \rbr{f(a|x,x') - \PP_{\mu}(a|x,x')}^2}
    \end{align*}
    \State Get a consistent ID abstraction (see~\pref{def: inverse dyamics consistency}) $\phi:\Xcal \rightarrow [N]$
    \State Set the policy cover for next time as $\Psi_{h+1} = \cbr{\pi_{i}^\star}_{i=1}^N$ where $\pi_{i}^\star$ is a deterministic policy that maximizes the reaching probability to the event $\phi(x)=i$.
\EndFor
\end{algorithmic}
\end{algorithm}

We start by formally defining the ID abstraction\footnote{There may be several abstractions which "agree" with the ID objective $\PP_\mu(a|x,x')$. Thus, we define a notion of consistent ID abstraction.}.
\begin{definition}[Inverse Dynamics Consistency]\label{def: inverse dyamics consistency}
Two observations $x'_1,x_2'\in \Xcal$ are consistent under the ID and an initial distribution $\mu$ if  $\forall x\in \Xcal, a\in \Acal$ either
\begin{enumerate}
    \item $\PP_\mu(a|x,x'_1)=\PP_\mu(a|x,x'_2)$,
    \item or either one of the following holds (i) $\PP_\mu(x,x'_1) =0$, (ii)  $\PP_{\mu}(x,x'_2| \unf(\Acal))=0$.
\end{enumerate}
where $x\sim \mu$ and $x_1',x_2'\sim T(\cdot |x,a)$.
\end{definition}
Relaying on this notion, we define an ID abstraction as follows.
\begin{definition}[Inverse Dynamics Abstraction]\label{def: inverse dyamics abstraction}
We say that an abstraction $\phi: \Xcal \rightarrow [N]$ is an \emph{ID abstraction} if for all $x_1',x_2'\in \Xcal$ for which $\phi(x_1')=\phi(x_2')$  it holds that $x_1',x_2'$ are consistent under the ID according to~\pref{def: inverse dyamics consistency}.
\end{definition}

Before addressing the problem that arises relying on the ID abstraction we elaborate on this definition, and specifically part two of its. We claim that this part is necessary to make the ID abstraction well defined. We motivate this definition by the two following arguments.

First, the ID object is a conditional probability function, $\PP_\mu(a|x,x')$. Thus, for $\PP_\mu(x,x')=0$ the conditional probability is not well defined. Indeed, part two of \pref{def: inverse dyamics consistency} is a possible solution to this issue -- without it, the definition of the ID abstraction is not mathematically defined. 

Second, a regression oracle that learns $\PP_\mu(a|x,x')$ is not affected -- i.e., has similar loss -- for all pairs $(x,x')$ for which $\PP_{\mu}(x,x'| \unf(\Acal))=0$. That is, the loss of any $f_\star(a|x,x')$ that approximates $P_\mu(a|x,x')$ is not affected by values of $f_\star(a|x,x')$ for which $\PP_{\mu}(x,x'| \unf(\Acal))=0$. Thus, the output of the regression oracle $f_\star$ can have arbitrary values on these pairs. This may result in  $f_\star(a|x,x_1') = f_\star(a|x,x_2') $  for observations for which $\PP_{\mu}(x,x'_1| \unf(\Acal))=0$ or $\PP_{\mu}(x,x'_2| \unf(\Acal))=0$. Put it differently, when learning $\PP_\mu(a|x,x')$ we cannot get any information outside the support of $\PP_{\mu}(x,x'| \unf(\Acal))=0$ and, thus, the values for these $(x,x')$ pairs can be arbitrary.

We say two observations $x'_1,x_2'$ have \emph{no shared common parent} if for any $x$ such that $\PP_{\mu}(x,x'_1| \unf(\Acal))>0$ it $\PP_{\mu}(x,x'_2| \unf(\Acal))=0$ and vice-versa. Our counter example relies on the following observation.
\begin{claim}[Inverse Dynamics may Merge Observations with no Shared Parent]\label{claim: key problem of ID}
If two observations $x_1',x_2'$ have no shared common parent then merging these states is always a consistent ID abstraction according to~\pref{def: inverse dyamics consistency}.
\end{claim}
This claim is a direct consequence of part two of~\pref{def: inverse dyamics consistency}: if for all $x$ it holds that either  $\PP_{\mu}(x,x'_1| \unf(\Acal))=0$ or  $\PP_{\mu}(x,x'_2| \unf(\Acal))=0$ then the two observations $x_1',x_2'$ may always be merged while resulting in a consistent abstraction according to the ID abstraction.

With this observation at hand, we construct a simple deterministic MDP for which an ID abstraction merges states that should not be merged; in the sense that no deterministic policy can reach both states.
\begin{proposition}[Failure of Inverse Dynamics]\label{prop: failure of inverse dynamics}
The exists a deterministic MDP such that~\pref{alg:id naive approach} does not return a policy cover on the states on the second time step.
\end{proposition}
\begin{proof}
Consider the MDP in~\pref{fig:combolock-results}, (a). At $h=1$, a consistent ID abstraction must separate the states $s_{2a},s_{2b},s_{2c}$ and $s_{2c}$. Thus,\pref{alg:id naive approach} separates all states at $h=1$. However, at $h=2$, since  $s_{3a}$ and $s_{3b}$ share no common parent, a consistent ID abstraction may merge these states (due to~\pref{claim: key problem of ID}). 

Then, since our policy class contains only deterministic policies, we get that necessarily at the end of the $h=2$ iteration, the policy cover that \pref{alg:id naive approach} returns does not contain a policy that reaches either $s_{3a}$ or $s_{3b}$. That holds since any deterministic policy that maximizes the reaching probability to $s_{3a}\cup s_{3b}$ will hit either $s_{3a}$ or $s_{3b}$. Thus, one of these states will not be reached by the policy cover.
\end{proof}

\newpage

\subsection{Bellman Rank Depends on the Exogenous State Cardinally}\label{app: bellman rank depends on exo cardinality}

\begin{restatable}[Bellman Rank Depends on the Exogenous State Cardinality]{proposition}{propositionBellmanRank}\label{prop: bellman rank}
There exists an Exogenous Block MDP $\Mcal$, policy class $\Pi$, and
value function class $\Fcal$ with the following properties: (1) the
endogenous state has size $2$, (2) the exogenous state has size $d$, (3) $|\Pi| = |\Fcal| = O(d)$, (4) the
optimal policy and value function are in $\Pi,\Fcal$ respectively, and
(5) the $(\Fcal,\Pi)$ bellman rank is $\Omega(d)$. 
Additionally, \textsc{Olive} has sample complexity $\Omega(\textrm{poly}(d)/\epsilon^2)$ to learn an $\epsilon$-optimal policy.
\end{restatable}
\begin{proof}
We construct the Exogenous Block MDP as follows. Let the horizon be
$H$ and set the starting endogenous state to be labeled $g_1$. From
$g_1$ there are two actions: action $a_1$ transits to $g_2$ while
action $a_2$ transits to $b_2$. We repeat this $H-1$ times randomizing
the good and bad action at each level, to arrive at either $g_H$ or
$b_H$. (Here $g$ denotes ``good'' and $b$ denotes ``bad''.) Let
$a_h^\star$ denote the good action at time $h$ and $\tilde{a}_h$
denote the bad action at time $h$.

There are
no actions at time $H$ and from $g_H$ the agent always receives reward
$1$, while from $b_H$ the agent always receives reward $0$. There are
no intermediate rewards. The exogenous state $\xi$ does not change
across time and, at the beginning of the episode, $\xi$ is drawn
uniformly from $[d]$.

The policy class $\Pi$ is defined as $\Pi =
\{\pi_\star,\pi_1,\ldots,\pi_d\}$ where $\pi_\star((g_h,\cdot)) =
a_h^\star$ and $\pi_i$ agrees with $\pi_\star$ everywhere, except for at state $(g_{H-1},i)$ where it takes $\tilde{a}_{H-1}$.
In other words, $\pi_i$ defects from the optimal policy only in the
good state at time $H-1$ when the exogenous variable
$\xi=i$. Meanwhile, the value function class is $\Fcal = \{f_\star,
f_1,\ldots,f_d\}$ where $f_\star$ is the optimal value function which satisfies
\begin{align*}
f_\star((g_h,\cdot)) = 1, \qquad f_\star((b_h,\cdot)) = 0
\end{align*}
and $f_i$ deviates from the $f_\star$ only on state $(b_H,i)$, where
$f_i((b_H,i)) = 1$.

First observe that $(f_i,\pi_i)$ is clearly bellman consistent at all
time steps except for the time step $H$, since $f_i$ predicts $1$ at
all states up to time $H-1$ and on all states visited by $\pi_i$ at
time $H$.

Now, let us examine the bellman error on roll-in $\pi_j$ for value
function/policy pair $(f_i,\pi_i)$ at the last time. First, the state
distribution visited by $\pi_j$ at time $H$ is
$\textrm{Unif}(\{g_H,k\}_{k \ne j} \cup \{b_H,j\})$. If $i \ne j$ then
$(f_i,\pi_i)$ has zero bellman error on all of these states since it
correctly predicts that the reward is $1$ in the good state and it
correctly predicts that the reward is $0$ on the bad state $(b_H,j)$.

On the other hand, if $i = j$ then the bellman error is $1/d$, since
$f_j$ \emph{incorrectly} predicts that the reward is $1$ on the bad
state. Thus we see that
\begin{align*}
\Ecal_H(\pi_i,(g_j,\pi_j)) := \EE_{(s_H,\xi_H,r_H) \sim \pi_i} \sbr{g_j(s_H,\xi_H) - r_H} = \frac{\one\{i = j\}}{d}.
\end{align*}
This verifies that the bellman rank is $d$.

Regarding \textsc{Olive}, note that the value functions are all
bellman consistent at the first $H-1$ time steps. In particular,
\begin{align*}
\forall h \leq H-1: \EE_{\pi_i} [f_i((g_h,\xi)) - f_i((s_{h+1},\xi))] = 0
\end{align*}
Thus \textsc{Olive} is unable to eliminate functions using data at the
first $H-1$ time steps. Additionally even with perfect evaluation of
expectation, due to adversarial tie breaking, \textsc{Olive} may take
$\Omega(d)$ iterations to find the optimal policy, since it may cycle
through $\pi_1,\ldots,\pi_d$ eliminating one at a time.

This argument can be extended to get $poly(d)/\epsilon^2$ sample complexity using standard technique. To prove a lower bound for the sample complexity of \texttt{OLIVE} we adjust the problem by making the rewards $1/2$ and $1/2+\epsilon$. Then, to eliminate a bad policy, we need to estimate the Bellman errors to accuracy of  $\epsilon/d$. This results in an $poly(d)/\epsilon^2$ lower bound for the sample complexity of \texttt{OLIVE}.

\end{proof}

\subsection{Failure of {\tt FLAMBE}~\citet{agarwal2020flambe}.}
In~\citet{agarwal2020flambe} the authors studied a representation learning problem for the linear MDP setting and suggested an algorithm,  {\tt FLAMBE}, that provably explores while learning the representation feature map of the linear MDP model. Their algorithm relies on a model-based approach to factorize the transition dynamics. However, focusing on the dynamics in observation space forces the modeling of the exogenous state as well, and the dimension of the factorization that they learn can scale with $\Sxcard$ (similar to the Bellman rank), leading to a $\text{poly}(\Sxcard)$ sample complexity for their approach.

\subsection{Failure of auto-encoding approaches}

 Much prior work uses auto-encoding or other unsupervised techniques
 for representation learning in RL. Examples include scalable
 count-based exploration methods~\citep{tang2017exploration} and \textsc{CURL}~\citep{laskin2020curl}. However, as these
 methods do not leverage the temporal nature of reinforcement learning,
 it is easy to see that the representations discovered may not be
 useful or relevant for exploration or policy learning, without relying
 heavily on inductive biases. More concretely, an autoencoding approach
 that aims to minimize reconstruction error on observations would
 prefer to memorize high-entropy irrelevant noise over lower-entropy
 relevant state information (see figure 4c in \citep{misra2020kinematic}). Unfortunately, the resulting learned
 representation may omit state information that is crucial for
 downstream planning.


\section{Structural Results for EX-BMDP} \label{app: structural results of EX-BMDP}
In this section we prove several useful structural results about the EX-BMDP model which will be essential for later analysis. A key definition which will be useful is the notion of \emph{endogenous policy.} We define the class of  \emph{endogenous policies} $\PiEnd$ as the policies that depend only on the \emph{endogenous} part of the state.  Formally, a policy $\pi\in \PiEnd$ has the property that for all  $x\in \Xcal:\,\,\,\pi(a|x) = \pi(a| \edo(x))$.  Restated, an endogenous policy chooses the same actions for a fixed endogenous state across varying exogenous states and observations. See that \emph{open loop} policies (see definition in~\pref{sec: setting ex block mdp}), which commit to a sequence of actions prior to the interaction, are always endogenous policies, since this sequence of actions is independent of the exogneous noise.

\begin{restatable}[Consequence of Endogenous Policy]{proposition}{propositionDecouplingOfEndognousState}\label{prop: decoupling of endognous policies}
Let $\pi\in \PiEnd$. Then, for any $h\in[H]$ it holds that $\PP_h(\sff| \pi) = \PP_h(\sd|\pi)\PP_h(\sx).$ Furthermore, there exists $\pi \in \Pi_{NS}\backslash\PiEnd$ such that $\PP_h(\sff| \pi) \neq \PP_h(\sd|\pi)\PP_h(\sx).$
\end{restatable}
\begin{proof}
{\bf First claim.}
We prove the result by induction.\\
{\bf Base case $h=1$.} The base case follows from the model assumption, that is,
\begin{align*}
    \PP_{h=1}(\sff|\pi) = \mu(\sff) = \mu(\sd) \mu_{\sx}(\sx).
\end{align*}
{\bf Induction step.} Assume the claim holds for $h$. We show it also holds for $h+1$ for $\pi\in \PiEnd$.  For any $\sff\in \Sf$, or, equivalently $\sff = (\sd,\sx)$ for $\sd\in \Sd,\sx\in \Sx$, it holds that
\begin{align*}
    &\PP_{h+1}(\sff|\pi) =  \sum_{\sff_{h}\in \Sf}\PP_{h+1}(\sff,\sff_{h-1}|\pi) \tag{Law of total probability}\\
    &= \sum_{\sff_{h}\in \Sf}\PP_{h}(\sff_{h}|\pi) T(\sff|\sff_{h},\pi) \tag{Bayes' theorem \& Markovian dynamics}\\
    &=\sum_{\sd\in \Sd,\sx\in \Sx}\PP_{h}(\sd_h|\pi)\PP_{h}(\sx_h) T(\sff|\sff_{h},\pi) \tag{Induction hypothesis}\\
    &=\sum_{\sd\in \Sd,\sx\in \Sx}\PP_{h}(\sd_h|\pi)\PP_{h}(\sx) T(\sd|\sd_{h},\pi)T(\sx|\sx_h) \tag{$\pi\in \PiEnd$ and EX-BMDP definition}\\
    &=\sum_{\sd\in \Sd}\PP_{h}(\sd_h|\pi)T(\sd|\sd_{h},\pi)\sum_{\sx\in \Sx}\PP_{h}(\sx) T(\sx|\sx_h)\\
    &=\PP_{h+1}(\sd|\pi)\PP_{h+1}(\sx),
\end{align*}
which concludes the proof.

{\bf Second claim.} Consider an MDP with $H=2$. The endogenous state is $b_d\in \cbr{0,1}$, the exogenous state is $b_\xi \in \cbr{0,1}$ and the full state is $\sff = \cbr{b_d, b_\xi}$. At the initial time step $b_d=1$ and $b_\xi \sim \unf(\cbr{0,1})$. Furthermore, the MDP is stationary, and its dynamics is given as follows. The exogenous state is fixed along trajectory, $$T(b_\xi'=1|b_\xi'=1)=T(b_\xi'=0|b_\xi'=0)=1.$$ The action set is of size two, $\Acal= \cbr{0,1}$, the endogenous dynamics evolves as 
\begin{align*}
&T(b_d' =1 | b_d=1,a=0) = T(b_d' =0 | b_d=1,a=1) = 1
\end{align*}
that is, applying $a=1$ the endogenous state switches to $0$ and applying $a=0$ leaves the endogenous state at $1$.

Let the policy be
\begin{align*}
    \pi(a=1 | b_\xi) = 
    \begin{cases}
    1 & b_\xi=1\\
    0 & o.w.
    \end{cases},
\end{align*}
and observe it depends on the exogenous state. 

We show that the next state distribution does not decouple the endogenous and exogenous part of the state space. 
\begin{align*}
    &\PP_{h=2}(\sff|\pi) = \PP_{h=2}(b_\xi)\PP_{h=2}(\sd|\pi,\sx) \\
    &= 0.5\rbr{\ind\cbr{b_\xi=1,\sd=0} + \ind\cbr{b_\xi=0,\sd=1}} \\
    &\neq \PP_{h=2}(\sd|\pi)\PP_{h=2}(b_\xi) =0.25.
\end{align*}
\end{proof}

\begin{restatable}[Existence of Endogenous Policy Cover]{proposition}{propositionPolicyCoverEndoPolicies}\label{prop: policy cover and endognous policies} Given an EX-BMDP, for any $h\in[H]$, let $\Psi_h$ be an endogenous policy cover of the endogenous dynamics, that is $\Psi_h\subseteq \PiEnd$ and for all $\sd\in\Sd$, $\max_{\pi\in \Psi_h} \PP_h(\sd|\pi) = \max_{\pi} \PP_h(\sd|\pi)$. Then, $\Psi_h$ is also a policy cover for the full EX-BMDP, that is: $\max_{\pi\in \Psi_h} \PP_h(\sff|\pi) = \max_{\pi} \PP_h(\sff|\pi)$.
\end{restatable}

\begin{proof}
Let $\Pi_{NS}$ be the set of tabular policies, $\pi\in \Pi_{NS}$ is a mapping $\pi: \Sf \rightarrow \Acal$\footnote{Observe that the tabular policies contain an optimal policy for the general Block-MDP model.}. Fix $\bar{\sff}\in \Sf_h$, where $\bar{\sff} = \rbr{ \bar{\sd},\bar{\sx} }$ for some $\bar{\sd}$ and $\bar{\sx}$ by the EX-BMDP model assumption and fix $h\in [H]$. We not show that for $\bar{\sff}$ there exists an optimal policy which is also endogenous policy that reaches $\bar{\sff}$. That is, we show that
\begin{align*}
     \max_{\pi\in \Pi_{NS}}\PP_h(\bar{\sff}|\pi) =  \max_{\pi\in \Pi_{d}}\PP_h(\bar{\sff}|\pi).
\end{align*}
See that
\begin{align*}
    \max_{\pi\in \Pi_{NS}}\PP_h(\bar{\sff}|\pi)  = \max_{\pi\in \Pi_{NS}}\EE\sbr{\sum_{h'=1}^h r_{h'}(\sff_{h'})\mid \pi}, 
\end{align*}
where $ r_{h}(\sff) = \ind\cbr{\sff = \bar{\sff}}$ and zero for all other time steps. We now show inductively that the optimal $Q$ function of the MDP $\Mcal_{\bar{\sff}} = (\Sf,\Acal, r,T,h)$ is given by
\begin{align*}
    Q^\star_{h'}(\sff,a) = \PP(\bar{\sx}_h|\sx_{h'}) Q^\star_{d,h'}(\sd,a),
\end{align*}
where $Q^\star_{d,h'}(\sd,a)$ does not depend on $\sx$ where $\PP(\bar{\sx}_h|\sx_{h'})$ is the probability the exogenous state at time step $h$ is $\sx_h$ given it is $\sx_{h'}$ at time step $h'$. Specifically, $Q^\star_{d,h'}(\sd,a)$ is the optimal $Q$ function on the endogenous MDP $\Mcal_{\bar{\sd}} = (\Sd,\Acal,r_{d}, T,h)$ with reward $r_{d,h}(\sd) = \ind\cbr{\sd = \bar{\sd}}$ and zero for all other time steps, defined on the endogenous state space. This implies that the there exists an optimal policy of $\Mcal_{\bar{\sff}}$ which is an endogenous policy. The policy
\begin{align*}
    \pi^\star_{\bar{\sff}} \in \arg\max_a Q^\star_{d,h'}(\sd,a)
\end{align*}
is an endogenous policy which is also optimal. 

{\bf Base case $h'=h$.} For the last time step, the claim trivially holds 
\begin{align*}
    &Q^\star_{h}(\sff,a) =  \ind\cbr{\sff = \bar{\sff}} = \ind\cbr{\sx = \bar{\sx}}\ind\cbr{\sd = \bar{\sd}} \\
    &= \PP(\bar{\sx}_h|\sx_{h})\ind\cbr{\sd = \bar{\sd}} = \PP(\bar{\sx}_h|\sx_{h})Q_{d,h}^\star(s,a).
\end{align*}

{\bf Induction step.} Assume the claim holds for $h'+1$ and prove it holds for $h'<h$. The optimal $Q$ function satisfies the following relations for any $h'\in [h]$ and $\sff_{h'}\in \Sf$, since $r_{h'}=0$ for any $h'<h$.
\begin{align*}
    &Q^\star_{h'}(\sff_{h'},a) = \EE\sbr{\max_{a'}  Q^\star_{h'+1}(\sff_{h'+1},a')|\sff_{h'},a}\\
    &=\EE\sbr{\max_{a'}  \PP(\bar{\sx}_h|\sx_{h'+1}) Q^\star_{d,h'+1}(\sd_{h'+1},a')|\sff,a} \tag{Induction step}\\
    & = \EE\sbr{ \PP(\bar{\sx}_h|\sx_{h'+1})\max_{a'} Q^\star_{d,h'+1}(\sd_{h'+1},a')|\sd_{h'},\sx_{h'},a}\\
    &=\sum_{\sx_{h'+1}} \PP(\bar{\sx}_h|\sx_{h'+1})T(\sx_{h'+1}|\sx_{h'})\sum_{\sd_{h'+1}}T(\sd_{h'+1}|\sd_{h'},a) \max_{a'}Q^\star_{d,h'+1}(\sd_{h'+1},a') \tag{Transition assumption}\\
    &=\PP(\bar{\sx_h}| \sx_{h'}) Q^\star_{d,h'}(\sd_{h'},a),
\end{align*}
where $$ Q^\star_{d,h'}(\sd_{h'},a)=  \sum_{\sd_{h'+1}} T(\sd_{h'+1}|\sd_{h'},a)\max_{a'}Q^\star_{d,h'+1}(\sd_{h'+1},a'),$$
does not depend on the exogenous part of the state space by the induction hypothesis since $Q^\star_{d,h'+1}(\sd_{h'+1},a')$ does not depend on the exogenous part of the state space. Furthermore, $Q^\star_{d,h'}(\sd_{h'},a)$ satisfy the Bellman equations of $\Mcal_{\bar{\sd}}$, and, thus, it is the optimal $Q$ function of $\Mcal_{\bar{\sd}}$.

This concludes the proof of the induction step.

\end{proof}

\begin{restatable}{proposition}{propositionEndoRewardOptPi}\label{prop: endo reward and optimal policy}
Consider an EX-BMDP and assume its reward function depends only on the endogenous part of the state, i.e., for all $x\in \Xcal,a\in \Acal$ and $h\in[H]$, $r_h(x,a) = r_h(\edo(x),a)$. Then, the class of endogenous policies $\PiEnd$ contains an optimal policy.
\end{restatable}

\begin{proof}
Let $r_h(x,a) = r_h(\edo(x),a)$ be a reward function that depends on the endogenous state. To establish the claim, it is sufficient to prove that for any $h\in [H],x\in \Xcal,a\in \Acal$ it holds that $Q^\star_h(x,a) = Q^\star_h(\edo(x),a)$, where $\cbr{Q^\star_h}_{h=1}^H$ is the optimal $Q$ function. This implies that
\begin{align*}
    \pi_{h}^\star\in \arg\max_a Q^\star_h(\edo(x),a),
\end{align*}
is an optimal policy. Furthermore, $\pi_{h}^\star$ is an endogenous policy, since it does not depend on the exogenous part of the state space.

We prove this claim by induction.\\
{\bf Base case, $h=H$.} Holds trivially since the reward function depends only on the endogenous state for all $x,a$. Thus, for all $x\in \Xcal,a\in \Acal$
\begin{align*}
    Q^\star_H(x,a) = r_H(x,a) =  r_H(\edo(x),a),
\end{align*}
by assumption.

{\bf Induction step, $h$.} Assume the claim holds for any $h+1$ for $h<H$. We now show it holds for the $h^{th}$ time step. The optimal $Q$ function satisfies the following relations  for all $x\in \Xcal,a\in \Acal$
\begin{align*}
    &Q^\star_h(x,a) = r_h(x,a) + \EE\sbr{ \max_a Q^\star_{h+1}(x',a)|x,a}\\
    & =  r_h(\edo(x),a) + \EE\sbr{ \max_a Q^\star_{h+1}(\edo(x'),a)|x,a} \tag{Induction hypothesis \& reward assumption}\\
    &= r_h(\edo(x),a) +  \sum_{\sd',\sx'} T(\sd'|\edo(x),a)T_\xi(\sx'| \exo(x)) \max_a Q^\star_{h+1}(\sd',a)\\
    &=r_h(\edo(x),a) +  \sum_{\sd'} T(\sd'|\edo(x),a)\max_a Q^\star_{h+1}(\sd',a).\tag{$\sum_{\sx'}T(\sx'|\exo(x))=1$}
\end{align*}
Thus, for all  $x\in \Xcal,a\in \Acal$ it holds that $Q^\star_h(x,a)  = Q^\star_h(\edo(x),a) $, i.e., is a function of the endogenous state.
\end{proof}

\newpage

\newpage

\section{Sample Complexity of $\algppe$} \label{app: sample complexity DPCID}

In this section, we present a proof showing that $\algppe$ learns a policy cover for nearly deterministic EX-BMDPs. More formally, to EX-BMDPS such that the endogenous dynamics is $\eta$ close to a deterministic MDP (see Assumption~\ref{assumption: near determistic} to quantification of $\eta$ close model).

\subsection{Inverse Dynamics Filters Exogenous State}

We start by establishing the following result which sheds further intuition on the performance of $\algppe$. In words, it says that the inverse dynamics objective filters exogenous noise and depends only on the endogenous part of the state.

\begin{restatable}[Inverse Dynamics Filters Exogenous State]{lemma}{LemmaIDFiltersExo}\label{lem: inverse model filters exo state}
For any endogenous policy $\pi\in \PiEnd$ and for all $a\in \Acal, h\in [H]$ it holds that
$
\PP_h(a,\edo(x_{h-1}) | x_{h},\pi) =  \PP_h(a,\edo(x_{h-1}) | \edo(x_{h}),\pi)$. If $\pi\in\Pi_{NS}\backslash\PiEnd$, $\PP_h(a,\edo(x_{h-1}) | x^{(2)}_{h},\pi) \neq \PP_h(a,\edo(x_{h-1}) | x^{(2)}_{h},\pi)$ for $\edo(x^{(1)}_{h}) = \edo(x^{(2)}_{h})$.
\end{restatable}

\begin{proof}
{\bf First statement.} Proven in \pref{lem: bayes optimal classifier} by explicitly applying Bayes' theorem and the decoupling property of the future state distribution of $\pi\in \Pi_d$ (see \pref{prop: decoupling of endognous policies}).

{\bf Second statement.} Consider an MDP with $H=2$. The endogenous state is $b_d\in \cbr{0,1}$, the exogenous state is $b_\xi \in \cbr{0,1}$ and the full state is $\sff = \cbr{b_d, b_\xi}$. At the initial time step $b_d=1$ and $b_\xi \sim \unf(\cbr{0,1})$. Furthermore, the MDP is stationary, and its dynamics is given as follows. The exogenous state is fixed along a trajectory,   $$T_\xi(b_\xi'=1|b_\xi'=1) = T_\xi(b_\xi'=0|b_\xi'=0)=1.$$ The action set is of size two, $\Acal= \cbr{0,1}$, the endogenous dynamics evolves as 
\begin{align*}
&T(b_d' =1 | b_d=1,a=1)= 1 - \alpha,\ T(b_d' =1 | b_d=1,a=0) =\alpha.
\end{align*}
Assume the policy is a function of the exogenous state given as follows
\begin{align*}
    \pi(a=1 | b_\xi) = 
    \begin{cases}
    1 & b_\xi=1\\
     0 & b_\xi=0
    \end{cases}.
\end{align*}
To conclude the proof, let $b'_d,b_\xi'$ denote the state at the second time step. We show that (suppressing the deterministic event $b_d=1$ at the first time step)
\begin{align*}
    \frac{\PP(a=1|b_d' =1,b_\xi'=1) }{\PP(a=1|b_d' =1,b_\xi'=0) } \neq1,
\end{align*}
thus, $\PP(a=1|b_d' =1,b_\xi'=1)\neq \PP(a=1|b_d' =1,b_\xi'=0)$. 

To prove this, by Bayes' theorem,
\begin{align}
    \frac{\PP(a=1|b_d' =1,b_\xi'=1) }{\PP(a=1|b_d' =1,b_\xi'=0) }  = \frac{\PP(b_d' =1,b_\xi'=1|a=1) }{\PP(b_d' =1,b_\xi'=0|a=1) } \frac{\PP(b_d' =1,b_\xi'=0) }{\PP(b_d' =1,b_\xi'=1) }. \label{eq: counter example 2}
\end{align}
Observe that $\PP(b_d' =1,b_\xi'=0|a=1)  = \PP(b_d' =1,b_\xi'=1|a=1) = 0.5\cdot \alpha$. Thus, 
\begin{align*}
\eqref{eq: counter example 2} =     \frac{\PP(b_d' =1,b_\xi'=0) }{\PP(b_d' =1,b_\xi'=1) }.
\end{align*}
See that
\begin{align*}
    &\PP(b_d' =1,b_\xi'=0) = \PP(b_d' =1|b_\xi'=0) \PP(b_\xi'=0) = 0.5\alpha \\
    &\PP(b_d' =1,b_\xi'=1) = \PP(b_d' =1|b_\xi'=1) \PP(b_\xi'=1) = 0.5(1-\alpha). 
\end{align*}
Thus, for $a\notin\cbr{0,1,0.5}$ we get that
\begin{align*}
    \frac{\PP(a=1|b_d' =1,b_\xi'=1) }{\PP(a=1|b_d' =1,b_\xi'=0) } = \frac{\alpha}{1-\alpha}\neq 1.
\end{align*}
\end{proof}
The intuition which underlies the construction of the second statement goes as follows. When a policy acts according to the exogenous state information on the exogenous state at time step $h=2$ may change the knowledge we have on the action taken at the previous time step $h=1$.

\begin{lemma}[Bayes Optimal Classifier]\label{lem: bayes optimal classifier} 
Assume that $\pi\in \PiEnd$ is an endogenous policy. Then, for every $x,x' \in \Xcal_h,a \in \Acal$ we have:
\begin{equation*}
    \PP(a, \edo(x) \mid  x',\pi) = \frac{T(\edo(x') \mid a, \edo(x))\PP(a, \edo(x) \mid \pi)}{\sum_{\sd\in \Sd,a\in \Acal}T(\edo(x') \mid a, \sd) \PP(a, \sd \mid \pi)}.
\end{equation*}
\end{lemma}
\begin{proof} 
We prove this result by applying Bayes' Theorem.
\begin{align}
    \PP(a, \edo(x) \mid  x',\pi) = \frac{\PP(x' \mid a, \edo(x),\pi) \PP(a, \edo(x) \mid \pi)}{\sum_{\sd\in \Sd,a\in \Acal}\PP(x' \mid a, \sd,\pi) \PP(a, \sd \mid \pi)} \label{eq: central result bayes optimal classifier}
\end{align}
Observe that by the model assumption,
\begin{align}
    &\PP(x' \mid a, \edo(x),\pi) =  q(x'|\edo(x'), \exo(x'))\PP(\edo(x'), \exo(x') \mid a, \edo(x),\pi)\label{eq: model assumption 1}. 
\end{align}
We focus on the second term. By the law of total probability and by applying Bayes' theorem,
\begin{align}
    &\PP(\edo(x'), \exo(x') \mid a, \edo(x),\pi) = \sum_{\sx}\PP(\edo(x'), \exo(x') \mid a, \edo(x),\sx, \pi)\PP(\sx\mid a, \edo(x), \pi) \nonumber \\
    & \underset{(a)}{=}   \sum_{\sx}\PP(\edo(x'), \exo(x') \mid a, \edo(x),\sx, \pi)\PP(\sx) \nonumber \\
    &\underset{(b)}{=} \sum_{\sx}T(\edo(x') \mid a, \edo(x))T_\xi( \exo(x') \mid \sx)\PP(\sx) \nonumber \\
    &=T(\edo(x') \mid a, \edo(x))\sum_{\sx}T_\xi( \exo(x') \mid \sx)\PP(\sx) \nonumber\\
    &=T(\edo(x') \mid a, \edo(x))\PP( \exo(x')). \label{eq: model assumption rel 2}
\end{align}
where $(a)$ holds since the policy $\pi\in \PiEnd$ followed by a fixed action $a$ is an endogenous policy and $\PP(\sx\mid a, \edo(x), \pi) = \PP(\sx)$ for an endogenous policy due to \pref{prop: decoupling of endognous policies}. The relation $(b)$ holds due to the Markov property of the latent state and the decoupling of the transition model of an EX-BMDP.

Plugging~\eqref{eq: model assumption 1} and~\eqref{eq: model assumption rel 2} into~\eqref{eq: central result bayes optimal classifier} we get
\begin{align*}
   &\PP(a, \edo(x) \mid  x',\pi) = \frac{T(\edo(x') \mid a, \edo(x))\PP(a, \edo(x) \mid \pi)}{\sum_{\sd\in \Sd,a\in \Acal}T(\edo(x') \mid a, \sd) \PP(a, \sd \mid \pi)}.
\end{align*}

\end{proof}

\subsection{Highlevel Analysis Overview of \pref{thm: PPE sample complexity}}

\paragraph{Analysis overview, deterministic dynamics.} Prior to addressing the near deterministic case, we consider the fully deterministic case. The analysis of this setting highlights the core ideas that are later utilized in the proof.

The idea which underlies the analysis of $\algppe$ is to prove, in an inductive manner, that at each time step $h\in 2,3.,..H$ the policy cover at $\Psi_{h-1}$ is a \emph{minimal policy cover} of the endogenous state space. That is, there is a one-to-one correspondence between endogenous states $\sd\in \Sd_{h-1}$ and open-loop policies $\nu\in \Psi_{h-1}$ such that for any $\sd\in \Sd_{h-1}$ there exists a unique open-loop policy $\pt\in \Psi_{h-1}$ that reaches $\sd$.

The base case holds since the starting state is deterministic. Assume the claim holds for $h-1$. Then, we need to show it holds for $h$. Since $\Psi_{h-1}$ is a minimal policy cover for the endogenous dynamics by a compositionality property of deterministic environments (see \pref{lem: optimal policy is a subset of initial Q}), the set of open-loop policies $\pts_h = \Psi_{h-1}\circ \Acal$ in which every policy in $\Psi_{h-1}$ is extended by all actions, is a policy cover for the $h^{th}$ time step. However, it may contain duplicates; two paths may reach to the same endogenous state.

Due to the above reasoning, by eliminating policies from $\pts_h$ we can recover a minimal policy cover for the $h^{th}$ time step and to establish to induction step. Let $\nu\in \pts_h$ and $f_h^\star(\nu,x') \equiv \PP_h(\nu|x', \unf(\pts_h))$ be the inverse dynamics which predicts the probability $\nu$ was taken while observing $x'$ and following the policy $\unf(\pts_h)$. Due to the one-to-one correspondence between $\cbr{(\sd,a)}_{a\in \Acal, \sd\in \Sd_{h-1}}$ and $\pts_h$ we expect this function to depend only on the endogenous state (see \pref{prop: policy cover and endognous policies}). Specifically, it is possible to show the following identity on which the elimination criterion of $\algppe$ is based upon.  Let $ \Delta^\star(i,j)= \EE_{x'\sim \PP_h(\cdot|\unf(\pts_h))}[\abr{ f_h^\star(\pt_i,x') - f_h^\star(\pt_j,x')}]$. For any $\pt_i,\pt_h\in \pts_h$ it holds that
\begin{align}
    \Delta^\star(i,j) = 
    \begin{cases}
    \frac{2}{|\pts_h|} & \textit{$\pt_i$ and $\pt_j$ lead to different endogenous state}\\
    0 & o.w.
    \end{cases} \label{eq: margin determinstic case supp}
\end{align}

Observe that we can learn $f_h^\star$ to sufficiently good accuracy with respect to the distribution $x'\sim \PP_h(\cdot|\unf(\pts_h))$ via standard regression guarantees of the MLE (see \pref{thm:mle}). Thus, for any $\pt_i,\pt_j\in \pts_h$ we can estimate $ \Delta^\star(i,j)$ and deduce whether they lead to the same endogenous state, when $\Delta^\star(i,j)$ is small, or not, when  $\Delta^\star(i,j)$ is large. This step is performed in the elimination step if $\algppe$, \pref{line:elimination}. Thus, we can safely eliminate open-loop policies form $\pts_h$ that reach that same endogenous state and be left with a minimal policy cover $\Psi_h$ for the next time step.

\paragraph{Analysis overview, near deterministic dynamics.} The analysis of the deterministic setting can be generalized to the near deterministic endogenous dynamics by a delicate modification of the above argument. 

For this setting, the simplest way we found to extend the argument for the deterministic case goes as follows. We prove via induction that for any $h\in [H]$ the set of open-loop policies $\Psi_h$ is a minimal policy cover of the \emph{$\eta$ close deterministic endogenous MDP}. Specifically, we use similar arguments as for the deterministic case, while replacing~\eqref{eq: margin determinstic case supp} with a proper generalization supplied in \pref{lem: margin for inverse path prediction near determinstic}. Observe that open loop policies are always endogenous policies ; such a policy does not depend on the state and is picked before  interacting with the environment. This fact, allows us to use an inverse dynamics objective which filters the exogenous information. 

Tho complete the proof, we show that a minimal policy cover of the $\eta$ close deterministic MDP is an $\eta H$ approximate policy cover (see \pref{def: approximate policy cover}). Furthermore, we also show that with this set of policies we can apply the \PSDP~\cite{bagnell2004policy} algorithm to get a near optimal policy when the reward function is an arbitrary function of observations, that might depend on the exogenous part of the state space.

\subsection{Proof of \pref{thm: PPE sample complexity}}
We start by formally defining a minimal policy cover for deterministic dynamics. This definition can be naturally generalized to general MDPs. Nevertheless, since we only study near deterministic dynamics we will only use the next, more specific definition.

\begin{definition}[Minimal Policy Cover for Deterministic Endogenous MDP]\label{definition: minimal policy cover}
Assume that the endogenous dynamics is deterministic. We say that a policy cover $\Psi_h$ is minimal for time step $h\in [H]$ if for every $\sd\in \Sdh$ there exists a unique path $\pt\in \Psi_h$ that reaches it, that is $\PP(\sd|\pt)=1$.
\end{definition}

In this section we supply the proof of \pref{thm: PPE sample complexity}. 

We now prove \pref{thm: PPE sample complexity} by establishing a more general result that will also be helpful when considering planning algorithms (see \pref{app: plannning in EX-BMDP analysis}). See that \pref{thm: PPE sample complexity} is a direct consequence of the second statement of the next result result.
\begin{theorem}[Sample Complexity: Policy Cover with $\algppe$]\label{thm: supp sample comeplxity PPE}
Assume that there exists an $\eta\leq \frac{1}{4\Sdcard H}$ close deterministic MDP for the endogenous dynamics. Let $\delta\in (0,1)$ and assume $\algppe$ has access to $O\rbr{S^2A^2\log\rbr{\frac{|\Fcal|SAH}{\delta}}}$ sample for each iteration $h\in [H]$. Then, with probability greater than $1-\delta$ the following holds.
\begin{enumerate}
    \item For any $h\in [H]$ the policy cover $\Psi_h$ is a minimal policy cover of the $\eta$ close near deterministic MDP of the endogenous dynamics.
    \item For any $h\in [H]$ the policy cover $\Psi_h$ is $\eta H$ approximate policy cover.
\end{enumerate}
\end{theorem}

\begin{proof}
{\bf First statement.} To prove this result we prove the following inductive argument. Denote by $\Mcal_D$ the $\eta$ close deterministic MDP, by $T_D$ its transition function, and let $\Sdh^D$ be the set of reachable states on $\Mcal_D$.  Let $G_h$ be the good event in which at the end of the $h^{th}$ time step for any $\sd \in \Sdh^D$ there exists a unique path  $\pt\in \Psi_h$ that reaches $\sd$ on the $\eta$ close deterministic MDP. More formally, let $\PP_h( \cdot |\nu,T_{D})$ denote the state distribution of time step $h$ on the $\eta$ close deterministic MDP when the open loop policy $\nu$ is applied. Then, the good event $G_h$ is defined as follows.
\begin{align}
    G_h = \cbr{\textit{at the end of the $h^{th}$ time step: } \forall \sd\in  \Sdh^D, \exists \textit{ a unique } \nu \in \Psi_h \textit{ s.t. } \PP_h(\sd| \nu,T_{D})=1 } \label{eq: definition the good event}
\end{align}

Conditioning on $G_{0:H}$ it holds that, for all $h\in [H]$, $\Psi_h$ is a minimal policy cover of the $\eta$ close deterministic MDP of the endognous dynamics. Hence, to conclude the proof of the first statement, we can prove that $\PP(G_{0: H}) \geq 1-\delta.$ To do so, and since $\PP(G_0)=1$, it is sufficient to prove that 
\begin{align}
    \PP(G_{h+1}| G_{0:h}) \geq 1-\frac{\delta}{H} \label{eq: what is needed to prove}
\end{align}
Then by an inductive argument it can be proved that $\PP(G_{0:h}) \geq 1-\frac{(h-1)\delta}{H}$. Indeed, the base case holds since $\PP(G_0)=1$, and the induction step holds since
\begin{align*}
    &\PP(G_1\cap \cdots\cap G_{h+1}) = \PP(G_1\cap \cdots\cap G_{h})\PP(G_{h+1}| G_1\cap \cdots\cap G_{h})\\
    &\geq \rbr{1-\frac{(h-1)\delta}{H}}\PP(G_{h+1}| G_1\cap \cdots\cap G_{h}) \tag{inductive assumption}\\
    &\geq \rbr{1-\frac{(h-1)\delta}{H}}\rbr{1-\frac{\delta}{H}} \tag{Assuming~\eqref{eq: what is needed to prove}}\\
    &\geq 1-\frac{h\delta}{H}.
\end{align*}
In \pref{lem: PPE conditional success} we prove that~\eqref{eq: what is needed to prove} holds and establish the theorem.

{\bf Second statement.} The endogenous minimal policy cover of the $\eta$ close deterministic MDP of the endogenous dynamics  implies an $\eta H$ approximate policy cover for the true EX-BMDP. We formally prove the claim in \pref{lem: translating policy covers}.
\end{proof}

\begin{lemma}[Translating Policy Covers]\label{lem: translating policy covers}
Let $\Mcal= (\Sf,\Acal, T, T_\xi,H,q)$ be an EX-BMDP with $\eta$ near deterministic endogenous dynamics  $T_D$. Assume that $\Psi_h$  is an endogenous minimal policy cover of the $\eta$ near deterministic endogenous for the $h^{th}$ time step. Assume that $\Mcal$ and $\Mcal_D$ are $\eta$ close. Then, $\Psi_h$ is an $h\eta$ approximate policy cover for $\Mcal$.
\end{lemma}
\begin{proof}
By \pref{lem: perturbations of state action frequency} it holds for any $s\in \Sdh$ that
\begin{align}
    \PP_h(s|\pi)\geq \PP_h(s|\pi, T_D) -\eta h,\textit{ and } \PP_h(s|\pi)\leq \PP_h(s|\pi, T_D) +\eta h . \label{eq: conseq of state action pertrub}
\end{align}
If $s\in \Sdh^D$ where $\Sdh^D$ is the set of reachable states at $h$ time step on $T_D$, there exists a policy $\pi\in \Psi_h$ such that $\PP_h(s|\pi, T_D)=1$. Thus, due to~\eqref{eq: conseq of state action pertrub}, for any $s\in \Sdh^D$ there exists a policy $\pi\in \Psi_h$ such 
\begin{align}
    \PP_h(s|\pi)\geq \PP_h(s|\pi, T_D) -\eta h = 1 -\eta h\geq \max_{\pi\in \PiEnd} \PP_h(s|\pi) -\eta h. \label{eq: translating relations 2}
\end{align}
Due to this result, since $\pi\in \Psi_h$ is an endogenous policy and by \pref{prop: decoupling of endognous policies} it holds that for any $\sx\in \Sx$
\begin{align*}
    &\PP_h(\sff|\pi) = \PP_h(\sx)\PP_h(\sd|\pi) \tag{\pref{prop: decoupling of endognous policies} \& $\pi\in \PiEnd$}\\
    &\geq \max_{\pi\in \PiEnd} \PP_h(\sx)\PP_h(s|\pi) -\eta h \tag{By~\eqref{eq: translating relations 2}}\\
    &= \max_{\pi\in \PiEnd} \PP_h(\sff|\pi) -\eta h \tag{\pref{prop: decoupling of endognous policies} \& $\pi\in \PiEnd$}\\
    &=  \max_{\pi\in \Pi_{NS}} \PP_h(\sff|\pi) -\eta h \tag{\pref{prop: policy cover and endognous policies}}.
\end{align*}
Hence, for any $\sff = (\sd,\sx)$ where $\sd\in \Sdh^D$ it holds that there exists $\pi\in \Psi_h$ such that
\begin{align*}
    \PP_h(\sff|\pi) \geq  \max_{\pi\in \Pi_{NS}} \PP_h(\sff|\pi) -\eta h.
\end{align*}

Assume that $\sd\in \Sdh/\Sdh^D$, that is, it is reachable in $h$ time steps on the true MDP $\Mcal$ but not on the deterministic MDP $\Mcal_D$. That is, for any $\pi$ it holds that $\PP_h(s|\pi, T_D) =0$. Thus, again by applying~\eqref{eq: conseq of state action pertrub}, it holds that for any $\pi\in \PiEnd$
\begin{align*}
    \PP_h(\sd|\pi)\leq \PP_h(\sd|\pi, T_D) +\eta h = \eta h.
\end{align*}
Fix $\sx\in \Sx$. By multiplying both sides by $\PP_h(\sx)$ we get that for any such $\sff= (\sd,\sx)$ it holds that
\begin{align*}
    &\PP_h(\sx)\max_{\pi\in \PiEnd}\PP_h(\sd|\pi) = \max_{\pi\in \PiEnd}\PP_h(\sx)\PP_h(\sd|\pi) \\
    & =  \max_{\pi\in \PiEnd}\PP_h(\sff|\pi) \tag{\pref{prop: decoupling of endognous policies} \& $\pi\in \PiEnd$}\\
    & \leq \eta h.
\end{align*}
On the other hand, by \pref{prop: policy cover and endognous policies} it holds that 
\begin{align*}
    \max_{\pi\in \PiEnd}\PP_h(\sff|\pi) = \max_{\pi\in \Pi_{NS}}\PP_h(\sff|\pi).
\end{align*}
Hence, for any $\sff = (\sd,\sx)$ where $\sd\in \Sdh/\Sdh^D$ it holds that $\max_{\pi\in \Pi_{NS}}\PP_h(\sff|\pi)\leq \eta H$. Hence, excluding policies that try to reach these state does not violate the definition of an $\eta h$ approximate policy cover (see \pref{def: approximate policy cover}).

\end{proof}

\begin{lemma}\label{lem: PPE conditional success}
The $\algppe$ algorithm at the $h^{th}$ time step satisfies that $$\PP(G_{h}| G_{0:h-1}) \geq 1-\frac{\delta}{H}$$ where the good event is defined in the proof of \pref{thm: supp sample comeplxity PPE}. That is, conditioning on the success of $\algppe$ at previous time-steps, with probability greater than $1-\frac{\delta}{H}$, $\algppe$ recovers the transition model of $\Mcal_D$ between all reachable states at the $h^{th}$ to the $h+1^{th}$ level.
\end{lemma}
\begin{proof}
We will prove the following claims conditioning on $G_{0:h-1}$. Combining the two concludes the proof of the lemma.
\begin{enumerate}
    \item \emph{Compositionality of policy cover.} The extended policy cover $\pts_h$ is a super set of a policy cover of the $h^{th}$ time step of the $\eta$ close deterministic transition model.
    \item \emph{Elimination succeeds.} Consider the $i,j$ iteration of the for loop in \pref{line:for lopp ppe} of \pref{alg:genik_path_elim}. If $\pt_i,\pt_j\in \pts_h = \Psi_{h-1}\times \Acal$ reaches the same endogenous state $\sd\in \Sdh$ on the $\eta$-close deterministic MDP, then $\algppe$ eliminates $\nu_j$ from $\pts_h$ with probability greater than ${1-\frac{\delta}{(SA)^2H}}$.
\end{enumerate}
Taking a union bound on all pairs we get that the elimination procedure succeeds. Thus, $\algppe$ outputs a perfect policy cover over $ \Sdh^D$ with probability greater than $1-\frac{\delta}{H}$. Thus concluding the proof.

We now prove the two claims.

{\bf Claim 1.} Conditioning on $G_{0:h-1}$ it holds that $\Psi_{h-1}$ is a perfect policy cover for the $\eta$ close deterministic transition model. By \pref{lem: optimal policy is a subset of initial Q}, utilizing the fact that $\Psi_{h-1}$ is a perfect policy cover, it holds that $\Upsilon_{h} = \Psi_{h-1}\times \Acal$ is a policy cover for the $h^{th}$ time step for the $\eta$ close determinstic dynamics.  That is, for any $\sd\in  \Sdh^D$ there exists \emph{at least} a single path $\pt\in \pts_h$ that reaches $\sd$ on the $\eta$ close deterministic MDP.

{\bf Claim 2.} Let $\pt_i,\pt_j\in \pts_h$ be a fixed and different paths in the extended policy cover at the $h^{th}$ time step, $\pts_h$. Let the empirical and the expected disagreement w.r.t. $f\in \Fcal$ be defined as follows.
\begin{align*}
    &\widehat{\Delta}(i,j;f) = \frac{1}{N}\sum_{i=1}^N \abr{f(\pt_i | x_{h,n}) - f(\pt_j | x_{h,n})}\\
    & \Delta(i,j; f) = \EE_{x_h \sim \PP(\cdot|\Ucal(\pts_h))}\sbr{\abr{f(\pt_i | x_h) -f(\pt_j | x_h)}}.
\end{align*}
Observe that $\widehat{\Delta}(i,j;f)$ is an average of $N$ i.i.d. and bounded in $[0,1]$ random variables. Furthermore, the expectation of the random variables is exactly $\Delta(i,j; f)$.  Thus, for any fixed $\pt_i,\pt_j\in \pts_h$ and $f\in \Fcal$, given $N = O\rbr{|\pts_h|^2\log\rbr{\frac{1}{\delta}}}$ samples, due to Hoeffding's inequality, it holds that
\begin{align}
    \abr{\widehat{\Delta}(i,j;f) - \Delta(i,j;f)} \leq \frac{1}{24 |\pts_h|}, \label{eq: ppe sample complexity relation hoeffding 1}
\end{align}
with probability greater than $1-\delta$. Applying the union bound on all $f\in \Fcal$ we get that~\eqref{eq: ppe sample complexity relation hoeffding 1} holds with probability greater than $1-\frac{\delta}{|\pts_h|^2H}$ for all  $f\in \Fcal$ given $$ N = O\rbr{|\pts_h|^2\log\rbr{\frac{|\Fcal||\pts_h|H}{\delta}}}.$$

Let $\widehat{\Delta}(i,j) \equiv \widehat{\Delta}(i,j;\hat{f}_h), \Delta(i,j) \equiv \Delta(i,j;\hat{f}_h)$ where $\hat{f}_h$ is the solution of the maximum liklihood objective. Since~\eqref{eq: ppe sample complexity relation hoeffding 1} holds w.r.t. $f\in \Fcal$, it implies that
\begin{align}
\abr{\widehat{\Delta}(i,j)  -\Delta(i,j)} \leq  \frac{1}{24|\pts_h|}, \label{eq: relation 1 path elimination version 2}
\end{align}
with probability greater than $1-\frac{\delta}{|\pts_h|^2H}$. Furthermore, notice that
\begin{align}
     & \abr{\Delta^\star(i,j)  - \Delta(i,j)} \nonumber  \\
     &= \abr{\EE_{x_h \sim \PP(\cdot|\Ucal(\pts_h))}\sbr{\abr{f_h^\star(\pt_i | x_h) - f_h^\star(\pt_j | x_h)} - \abr{\hat{f}_h(\pt_i | x_h) - \hat{f}_h(\pt_j | x_h)} }}   \nonumber \\
     &\leq \EE_{x_h \sim \PP(\cdot|\Ucal(\pts_h))}\sbr{\abr{\hat{f}_h(\pt_i | x_h) - f_h^\star(\pt_i | x_h)}} + \EE_{x_h \sim \PP(\cdot|\Ucal(\pts_h))}\sbr{\abr{\hat{f}_h(\pt_j | x_h)  - f_h^\star(\pt_j | x_h)}}  \nonumber \\
     &\leq 2\epsilon, \label{eq: relation 2 path elimination version 2}
\end{align}
where the second relation holds since $\abr{|a| - |b|} \leq |a-b|$ due to the triangle inequality, and the third relation holds with probability greater than $1-\frac{\delta}{|\pts_h|^2H}$ due to \pref{thm:mle} where $2\epsilon= \frac{2}{24|\pts_h|}$ for $N=O\rbr{|\pts_h|^2 \log\rbr{\frac{|\Fcal||\pts_h|^2H}{\delta}}}$. Observe that \pref{thm:mle} is applicable due to the realizability \pref{assum: realizability Fcal}.

By combining~\eqref{eq: relation 1 path elimination version 2} and~\eqref{eq: relation 2 path elimination version 2} it holds that
\begin{align}
     \abr{\widehat{\Delta}(i,j)-\Delta^\star(i,j) } \leq 2\epsilon +\frac{1}{24|\pts_h|} =  \frac{1}{8|\pts_h|}. \label{eq: exact and estimated disagreement}
\end{align}
By taking a union bound on all $\nu_i,\nu_j\in \pts_h$ we get that~\eqref{eq: exact and estimated disagreement} for all  $\nu_i,\nu_j\in \pts_h$ with probability greater than $1-\frac{\delta}{H}.$

Finally, by \pref{lem: margin for inverse path prediction near determinstic} it holds that 
\begin{align*}
     \begin{cases}
     \abr{\Delta^\star(i,j)} \leq \frac{1}{4|\pts_h|} & \textit{$\pt_i$ and $\pt_j$ reach the same endogenous state on $\Mcal_{D,\eta}$}\\
     \abr{\Delta^\star(i,j)} \geq \frac{1}{|\pts_h|}  & o.w.
     \end{cases}
\end{align*}
if $\Psi_{h-1}$ is a perfect policy cover, which holds conditioning on $G_{h-1}$. This fact, together with~\eqref{eq: exact and estimated disagreement} implies that with probability greater than $1 - \frac{\delta}{H}$
\begin{align*}
     \begin{cases}
     \abr{\widehat{\Delta}(i,j)} \leq \frac{3}{8|\pts_h|} & \textit{$\pt_i$ and $\pt_j$ reach the same endogenous state on $\Mcal_{D,\eta}$}\\
     \abr{\widehat{\Delta}(i,j)} \geq \frac{7}{8|\pts_h|}  & o.w.
     \end{cases}
\end{align*}

Thus, for any part $\nu_i,\nu_j$ the elimination process succeeds with probability greater than $1 - \frac{\delta}{H}$ since we set the elimination threshold to be $\frac{5}{8|\pts_h|}$ in \pref{line:elimination} of \pref{alg:genik_path_elim}. That is,  with probability greater than $1 - \frac{\delta}{H}$ at the end of the $h^{th}$ episode we are left with a perfect policy cover on the $\eta$ close deterministic MDP, conditioning on $G_{0: h-1}$.
\end{proof}

The following lemma highlights the existence of margin for MDPs which are $\eta$ close to a deterministic MDP for $\eta \leq 1/(4|\Sd|H).$

\begin{restatable}[Existence of Margin for Inverse Path Prediction Near Deterministic Dynamics]{lemma}{lemmaMarginforPPENearDet}\label{lem: margin for inverse path prediction near determinstic}
Assume the transition model is $\eta$ close to deterministic dynamics. Let $\Psi_{h-1}$ be a perfect policy cover of the $\eta$ close deterministic dynamics and  $\pts_h = \Psi_{h-1}\times \Acal$ its extension. Let ${\Delta^\star(i,j) = \EE_{x'\sim \PP(\cdot| \Ucal(\pts_h))}\sbr{\abr{f^\star(\pt_i \mid x') -  f^\star(\pt_j \mid x')}}}$ for a fixed $\pt_i,\pt_j\in \pts_h.$ Then,
\begin{align*}
     \begin{cases}
     \Delta^\star(i,j) \leq \frac{\eta h |\Sdh|}{|\pts_h|} & \textit{$\pt_i$ and $\pt_j$ reach the same endogenous state on $\Mcal_{D,\eta}$}\\
     \Delta^\star(i,j) \geq 2\frac{1-2\eta h}{|\pts_h|}  & o.w.
     \end{cases}
\end{align*}
In particular, if $\eta \leq 1/(4|\Sdh|h)$, then we have 
\begin{align*}
     \begin{cases}
     \Delta^\star(i,j) \leq \frac{1}{4|\pts_h|} & \textit{$\pt_i$ and $\pt_j$ reach the same endogenous state on $\Mcal_{D,\eta}$}\\
     \Delta^\star(i,j) \geq \frac{1}{|\pts_h|}  & o.w.
     \end{cases}
\end{align*}
\end{restatable}

\begin{proof}
For a path $\pt$, let $s(\pt)$ denote the state $s\in\Sdh$ reached by the path $\pt$ in the corresponding deterministic MDP. Then \pref{lem: perturbations of state action frequency}, we have that 
\begin{equation}
    \PP(s_h=s(\pt)|\pt) \geq 1-\eta h\quad \mbox{and for any $s\in\Sdh$} \quad  \PP(s_h=s\ne s(\pt)|\pt) \leq \eta h.
    \label{eq:perturb_prob_path}
\end{equation}

Using this, and the form of the Bayes optimal predictor from \pref{lem: bayes optimal classifier}, we have that for any two paths $\pt_i$ and $\pt_j$ such that $s(\pt_i) = s(\pt_j)$:
\begin{align*}
   \Delta^\star(i,j)&=\EE_{x\sim \PP(\cdot| \Ucal(\pts_h))}[|f^\star(\pt_i \mid x) -  f^\star(\pt_j \mid x)|] \nonumber \\
    &= \sum_{s\in\Sdh} \PP(s| \Ucal(\pts_h))\abr{f^\star(\pt_i \mid s) -  f^\star(\pt_j \mid s)} \nonumber \\
    &= \sum_{s\in\Sdh}\PP(s)\frac{\abr{\PP(s|\pt_i)-\PP(s|\pt_j)}}{\sum_{\pt\in\pts_h} \PP(s|\pt)}\\
    &= \sum_{s\in\Sdh} \PP(s)\frac{\abr{\PP(s|\pt_i)-\PP(s|\pt_j)}}{|\pts_h|\PP(s)}\\
    &= \frac{\left(\abr{\PP(s(\pt_i)|\pt_i)-\PP(s(\pt_i)|\pt_j)} + \sum_{s\ne s(\pt_i)} \abr{\PP(s|\pt_i)-\PP(s|\pt_j)}\right)}{|\pts_h|}\\
    &\leq \frac{\left((1-(1-\eta h)) + (|\Sdh|-1)\eta h\right)}{|\pts_h|}\\
    &= \frac{\eta h |\Sdh|}{|\pts_h|}.
\end{align*}

Here the first step follows from the form of $f^\star$ in \pref{lem: bayes optimal classifier}, which only depends on the latent endogenous state. Second step further expands the definition of $f^\star$ and the next step uses the uniform distribution over paths that we roll-in with. The inequality follows from our earlier bound~\eqref{eq:perturb_prob_path}. 

Similarly, if two paths $\pt_i$ and $\pt_j$ do not lead to the same endogenous state, then we get 
\begin{align*}
   \Delta^\star(i,j)&=\sum_{s\in\Sdh} \frac{\abr{\PP(s|\pt_i)-\PP(s|\pt_j)}}{|\pts_h|}\\
    &\geq \frac{\abr{\PP(s(\pt_i)|\pt_i)-\PP(s(\pt_i)|\pt_j)}}{|\pts_h|} + \frac{\abr{\PP(s(\pt_j)|\pt_i)-\PP(s(\pt_j)|\pt_j)}}{|\pts_h|}\\
    &\geq 2\frac{1-2\eta h}{|\pts_h|},
\end{align*}
where the first inequality follows since we can drop the non-negative terms corresponding to the states other than $s(\pt_i)$ and $s(\pt_j)$, while the second bound follows from \pref{eq:perturb_prob_path}. 
\end{proof}

\subsection{Help Lemmas}

\begin{lemma}[Policy Cover in Deterministic Environment]\label{lem: optimal policy is a subset of initial Q}
Let $\Psi_{h-1}$ be a policy cover for time step $h$ and let $\pts_{h}= \Psi_{h-1} \circ \Acal$. Then, for any $s\in \Sdh$ there exists a $\pt\in \pts_{h}$ such that $\pt$ reaches $\sd\in \Sdh$ in $h$ time-steps. That is, $\pts_{h}$ is a super set of the policy cover.
\end{lemma}
\begin{proof}
If $\sd\in \Sdh$ it means it can be reached in $h$ time-steps. This implies that exists a state $\sd_{h}\in \Sdh$ and an action $a$ such that $\sd_{h} = f(\sd_h,a)$. Since for any $\sd_h\in \Sdh$ there exists $\pt$ that reaches $\sd_h$ and since we extend $\Psi_{h-1}$ by taking all possible actions the claim follows.
\end{proof}

\newpage
\section{Planning with Approximate Policy Cover in EX-BMDP}
\label{app: plannning in EX-BMDP analysis}

Given access to the policy cover $\algppe$ outputs, we can efficiently explore in an EX-BMDP  as we show in this section. Interestingly, although the policy cover is obtained by exploring the endogenous part of the state space it still allows to effectively explore the full state space. We now address the problem of learning a near optimal policy w.r.t. a \emph{general reward function} via access to the output of $\algppe$. Later, we consider a more specific case, in which the reward is a function of the \emph{endogenous state space}. For the first case, we apply the \PSDP{} algorithm, while for the latter we can apply the more efficient Value Iteration (VI) procedure.

\paragraph{Planning with a general reward function.} In \pref{app: plannning in EX-BMDP analysis PSDP} we consider a general reward function of the observations, that is $r_h(x,a)$. Given the approximate policy cover $\algppe$ outputs we can apply the \PSDP~\cite{bagnell2004policy} algorithm, since it only requires access to a sufficiently good policy cover. Intuitively, given a sufficiently good policy cover we can explore the full state space -- both the endogenous and exogenous -- and learn an optimal policy based on the dynamics-programming procedure of \PSDP.

We make the next policy completeness assumption~\cite{dann2018oracle,misra2020kinematic}, required for the success of the \PSDP procedure.
\begin{assumption}[Policy Completeness] \label{assum: policy completness}
For any non-stationary policy represented by $\Pi$, ${\pi=\pi_1\circ \cdot \circ \pi_H}$ where $\cbr{\pi_h}_{h=1}^H \in \Pi$ for all $h\in [H]$ there exists $\pi \in \Pi$ for which
\begin{align*} 
    \forall x\in \Xcal_h: \ \pi(x) = \arg\max_{a} Q^{\pi}_h(x,a). 
\end{align*}
\end{assumption}

Given access to such a policy class $\Pi$, and given the output policy cover of $\algppe$, \PSDP has the following guarantees for an EX-BMDP (see \pref{app: plannning in EX-BMDP analysis PSDP} for a proof).
\begin{restatable}[\PSDP for EX-BMDP]{theorem}{theoremPSDPforEXBMDP}\label{thm: psdp for ex bmdp}
Let $\epsilon\in (0,1/2)$ and $\delta\in (0,1)$ and assume that $\cbr{\Psi_h}_{h=1}^H$ is an $\eta H$ near deterministic policy cover for the endogenous state space for all $h\in [H]$, such that $\eta\leq \frac{1}{4 S H}$.  Assume that \PSDP is given $N = O\rbr{\frac{S^2AH^4\log\rbr{\frac{|\Pi|H}{\delta}}}{\epsilon^2}}$ in total. Then, with probability greater than $1-\delta$, \PSDP returns a near optimal policy $\widehat{\pi}$ such that
\begin{align*}
    V(\pi^\star)  - V(\hat{\pi}) \leq  \epsilon+ H^3\eta,
\end{align*}
\end{restatable}
Although the analysis is standard, the final result demonstrates an interesting phenomena: the sample complexity of \PSDP depends on the quality and  cardinality of the policy cover, and not on the cardinality of the underlying state space. Specifically, the cardinality of the latent state of an EX-BMDP is $\Sfcard = \Sdcard\times\Sxcard$, whereas \PSDP learns a near optimal policy with $O(S^2AH^2\log\rbr{\frac{|\Pi|}{\delta}}/\epsilon^2)$ samples. At a higher level, the latter result can be thought of as an extension of \pref{prop: policy cover and endognous policies}, in which, access to an exact policy cover was assumed.

\paragraph{Planning with an endogenous reward function.} By further inspecting $\algppe$ it can be seen $\algppe$ can also return the model of the $\eta$ close deterministic endogenous dynamics $T_D$. Instead of merely eliminating paths in $\algppe$, \pref{line:elimination}, we can obtain the deterministic model of the endogenous dynamics by tracking which $(\sd_{h-1},a)$ pairs reached to $\sd_h$ at each time step. In case the reward function depends only on the endogenous dynamics, for all $x\in \Xcal$ $ r(x,a) =  r(\edo(x),a),$ we can simply find the optimal policy via VI w.r.t. the $\eta$ close deterministic endogenous dynamics.

The performance guarantee of the VI procedure as well as full description of the algorithm is supplied in \pref{app: plannning in EX-BMDP analysis VI}.
\begin{restatable}[Value Iteration for EX-BMDP]{proposition}{propVIforPlanning}\label{prop: value iteration for planning}
Let $\epsilon,\delta\in (0,1)$ and assume that $T_{D}$ is the $\eta$ close deterministic model $\algppe$ outputs. Assume that \pref{alg:VI_alg} have access to $N = O\rbr{\frac{SAH^2 \log\rbr{\frac{SAH}{\delta}}}{\epsilon^2}}$ samples in total. Then, the policy \pref{alg:VI_alg} outputs is $6\eta H^3 +\epsilon$ optimal, that is
\begin{align*}
    V(\pi^\star)  - V(\hat{\pi}) \leq  \epsilon + 6H^2\eta.
\end{align*}
\end{restatable}
Utilizing VI as oppose to \PSDP can dramatically reduce the computational burden. Furthermore, see that by  \pref{prop: endo reward and optimal policy}, there exists an optimal policy which is endogenous.

\subsection{General Observation Based Reward: Policy Search by Dynamic Programming}\label{app: plannning in EX-BMDP analysis PSDP}
\begin{algorithm}[H]
\caption{$\PSDP(\epsilon,\delta,\{\Psi_t\}_{h=1}^{H})$}
\label{alg:psdp_alg}
\setstretch{1.2}
\begin{algorithmic}[1]
\State {\bf require: } $\eta H$ near deterministic policy cover $\cbr{\Psi_h}_{h=1}^H$, $\epsilon,\delta>0$ accuracy and confidence level
\State {\bf initialize: }$\widehat{\pi} = \cbr{\emptyset}$
\For{$h=H, H-1, \cdots, 1$}
\State $\Dcal = \emptyset$
\State Set $N = O\rbr{\frac{|\Psi_h|A \log\rbr{\frac{|\Pi|A}{\delta}}}{\epsilon^2}}$
\For{$N$ times}
\State $(x, a, p,\hat{Q}_{\widehat{\pi}}) \sim \unf(\Psi_h) \circ \unf(\Acal) \circ \widehat{\pi}$
\State $\Dcal \leftarrow \Dcal \cup \{(x, a, p, \hat{Q}_{\widehat{\pi}})\}$
\EndFor
\State $\widehat{\pi}_h = \arg\max_{\pi} \sum_{(x, a, p, \hat{Q}_{\widehat{\pi}}) \in \Dcal} \rbr{ \frac{\widehat{Q}_{\widehat{\pi}}(x_{h},a)\one\{a'=a\}}{1/A}}$
\State $\widehat{\pi} \gets \widehat{\pi}_h\circ \widehat{\pi}$
\EndFor
\Return $\widehat{\pi}$
\end{algorithmic}
\end{algorithm}

Let $\widehat{\pi}$ be the policy at the beginning of the $h^{th}$ iteration. Hence, it is an $H-(h+1)$ non-stationary policy, defined on steps $h+1,..,H$. From contextual bandit guarantees we have that
\begin{equation}
    \EE_{x \sim P_h(\cdot | \unf(\pts_{h}))}\sbr{Q^{\widehat{\pi}}_{h}(x; \pi\circ \widehat{\pi}) - Q_h^{{\widehat{\pi}}}(x; \widehat{\pi}_h\circ \widehat{\pi})} \le \epsilon_{cb} \defeq 4\sqrt{\frac{A}{N}\ln\left(\frac{2|\Pi|}{\delta}\right)}, \label{eq: contextual bandit guarantee}
\end{equation}
with probability at least $1-\delta$. The policy $\hat{\pi}_h$ is an output of an offline contextual bandits oracle~\cite{langford2008epoch,agarwal2014taming}
\begin{align*}
    \hat{\pi}_h \in \arg\max_{\pi\in \Pi} \sum_{x_i,a_i,r_i} \EE_{a'\sim \pi(\cdot|x)}\sbr{ \frac{r_i \ind\cbr{a'=a}}{1/A}},
\end{align*}
and $r_i$ is a role in policy which takes an action $a_i$ and follows $\hat{\pi}$ until the last time step. In \pref{lem: contextual bandits guarantee} we formally state the result which is also establishing in \citep{misra2020kinematic}, Proposition 6.

\theoremPSDPforEXBMDP*

\begin{proof}
From the difference lemma we have:
\begin{align}
    V(\pi^\star) -     V(\hat{\pi}) &= \sum_{h=1}^H \EE_{x \sim P_h(\cdot | \pi^\star)} \left[ Q^{\hat{\pi}_{h+1:H}}_{h+1}(x, \pi^\star_h(x)) - Q^{\hat{\pi}_{h+1:H}}_{h+1}(x; \widehat{\pi}_h(x))\right].\label{eq: relation 1 pspd analysis}
\end{align}
Next, we fix an $h\in [H]$ and bound each term of the above sum. Remember that, conditioning on the good event,  $\Psi_{h}$ is a minimal policy cover of the $\eta$ near deterministic endogenous MDP $\Mcal_D$. Let the set of reachable endogenous states, after $h$ time steps, of $\Mcal_D$ be denoted by $\Sd_{h}^D$. Then, the following holds.
\begin{align}
    &\EE_{x \sim P_h(\cdot | \pi^\star)} \left[ Q^{\hat{\pi}_{h+1:H}}_{h+1}(x, \pi^\star_h(x)) - Q^{\hat{\pi}_{h+1:H}}_{h+1}(x; \widehat{\pi}_h(x))\right] \nonumber \\
    & = \sum_{\sd\in \Sd_{h}^D, \sx\in \Sxh} \PP_h(\sd, \sx| \pi^\star)\EE_{x\sim q(\cdot| \sd, \sx)} \left[ Q^{\hat{\pi}_{h+1:H}}_{h+1}(x, \pi^\star_h(x)) - Q^{\hat{\pi}_{h+1:H}}_{h+1}(x; \widehat{\pi}_h(x))\right]  \nonumber\\
    &\quad +\sum_{\sd\in \Sdh / \Sd_{h}^D, \sx\in \Sxh} \PP_h(\sd, \sx| \pi^\star)\EE_{x\sim q(\cdot| \sd, \sx)} \left[ Q^{\hat{\pi}_{h+1:H}}_{h+1}(x, \pi^\star_h(x)) - Q^{\hat{\pi}_{h+1:H}}_{h+1}(x; \widehat{\pi}_h(x))\right]   \nonumber\\
    &\leq \sum_{\sd\in \Sd_{h}^D, \sx\in \Sxh} \PP_h(\sd, \sx| \pi^\star)\EE_{x\sim q(\cdot| \sd, \sx)} \left[ Q^{\hat{\pi}_{h+1:H}}_{h+1}(x, \pi^\star_h(x)) - Q^{\hat{\pi}_{h+1:H}}_{h+1}(x; \widehat{\pi}_h(x))\right]   \nonumber\\
    &\quad +H\sum_{\sd\in \Sdh / \Sd_{h}^D, \sx\in \Sxh} \PP_h(\sd, \sx| \pi^\star) \tag{Values are in $[0,H]$} \\
    &=\sum_{\sd\in \Sd_{h}^D, \sx\in \Sxh} \PP_h(\sd, \sx| \pi^\star)\EE_{x\sim q(\cdot| \sd, \sx)} \left[ Q^{\hat{\pi}_{h+1:H}}_{h+1}(x, \pi^\star_h(x)) - Q^{\hat{\pi}_{h+1:H}}_{h+1}(x; \widehat{\pi}_h(x))\right]+H\sum_{\sd\in \Sdh / \Sd_{h}^D} \PP_h(\sd| \pi^\star) \label{eq: relation 2 pspd analysis}
\end{align}
where the last relation holds by marginalizing over $\sx$.

{\bf Bound on the first term of~\eqref{eq: relation 2 pspd analysis}.} By a standard PSDP analysis as follows.
\begin{align}
    &\sum_{\sd\in \Sd_{h}^D, \sx\in \Sxh} \PP_h(\sd, \sx| \pi^\star)\EE_{x\sim q(\cdot| \sd, \sx)} \left[ Q^{\hat{\pi}_{h+1:H}}_{h+1}(x, \pi^\star_h(x)) - Q^{\hat{\pi}_{h+1:H}}_{h+1}(x; \widehat{\pi}_h(x))\right] \nonumber \\
    &=\sum_{\sd\in \Sd_{h}^D, \sx\in \Sxh} \frac{\PP_h(\sd, \sx| \pi^\star)}{\PP_h(\sd, \sx| \unf(\Psi_h))}\PP_h(\sd, \sx| \unf(\Psi_h))\EE_{x\sim q(\cdot| \sd, \sx)} \left[Q^{\hat{\pi}_{h+1:H}}_{h+1}(x, \pi^\star_h(x)) - Q^{\hat{\pi}_{h+1:H}}_{h+1}(x; \widehat{\pi}_h(x))\right]  \nonumber
     \\
    &\leq \sum_{\sd\in \Sd_{h}^D, \sx\in \Sxh} \frac{\PP_h(\sd, \sx| \pi^\star)}{\PP_h(\sd, \sx| \unf(\Psi_h))}\PP_h(\sd, \sx| \unf(\Psi_h))\EE_{x\sim q(\cdot| \sd, \sx)} \left[ \max_{\pi(x)}Q^{\hat{\pi}_{h+1:H}}_{h+1}(x, \pi(x)) - Q^{\hat{\pi}_{h+1:H}}_{h+1}(x; \widehat{\pi}_h(x))\right]  \nonumber\\
    &\leq \norm{ \frac{\PP_h(\cdot| \pi^\star)}{\PP_h(\cdot| \unf(\Psi_h))}}_\infty \sum_{\sd\in \Sd_{h}^D, \sx\in \Sxh}\PP_h(\sd, \sx| \unf(\Psi_h))\EE_{x\sim q(\cdot| \sd, \sx)} \left[ \max_{\pi(x)}Q^{\hat{\pi}_{h+1:H}}_{h+1}(x, \pi(x)) - Q^{\hat{\pi}_{h+1:H}}_{h+1}(x; \widehat{\pi}_h(x))\right]  \nonumber\\
    &\leq \norm{ \frac{\PP_h(\cdot| \pi^\star)}{\PP_h(\cdot| \unf(\Psi_h))}}_\infty \EE_{x\sim \PP_h(\cdot| \unf(\Psi_h))} \left[ \max_{\pi(x)}Q^{\hat{\pi}_{h+1:H}}_{h+1}(x, \pi(x)) - Q^{\hat{\pi}_{h+1:H}}_{h+1}(x; \widehat{\pi}_h(x))\right]  \nonumber\\
    &\leq \norm{ \frac{\PP_h(\cdot| \pi^\star)}{\PP_h(\cdot| \unf(\Psi_h))}}_\infty \epsilon_{cb}.\label{eq: relation 4 pspd analysis}
\end{align}
The forth relation holds as we added additional positive terms, and $\norm{ \frac{\PP_h(\cdot| \pi^\star)}{\PP_h(\cdot| \unf(\Psi_h))}}_\infty\geq 0.$ The fifth relation holds by the contextual bandits guarantee~\eqref{eq: contextual bandit guarantee} and due to the policy completeness \pref{assum: policy completness} by which the maximizer is contained in $\Pi$. 

We now bound the importance sampling ration. 
\begin{align*}
    \norm{ \frac{\PP_h(\cdot| \pi^\star)}{\PP_h(\cdot| \unf(\Psi_h))}}_\infty   = \max_{\sd \in \Sd_{h}^D,\sx\in \sx}  \frac{\PP_h(\sd,\sx| \pi^\star)}{\PP_h(\sd,\sx| \unf(\Psi_h))}.
\end{align*}
First, by Bayes' theorem it holds that $$\PP_h(\sd,\sx| \pi^\star) = \PP_h(\sd| \pi^\star,\sx)\PP_h(\sx|\pi^\star) = \PP_h(\sd| \pi^\star,\sx)\PP_h(\sx),$$
since the exogenous state is not affected by the policy.

Second, observe that $$\PP_h(\sd,\sx| \unf(\Psi_h)) = \PP_h(\sx)\PP_h(\sd| \unf(\Psi_h))$$ since the $\algppe$ outputs an endogenous policy cover (thus, we can apply \pref{prop: decoupling of endognous policies}). Furthermore, for any endogenous state $\sd\in \Sd_{h}^D$ it holds that
\begin{align}
    \PP_h(\sd| \unf(\Psi_h))= \frac{1}{|\Psi_h|}\sum_{\nu\in \Psi_h} \PP_h(\sd| \nu) \geq \frac{1-\eta h}{\Psi_h}, \label{eq: state is reachable pspd}
\end{align}
since there exists $\nu\in \Psi_h$ such that $$\PP_h(\sd| \nu)\geq \PP_h(\sd| \nu,T_D)-\eta h = 1-\eta h.$$ Remember that $\Psi_h$ is a (minimal) policy cover of the near deterministic dynamics $\Mcal_D$ and $\sd\in \Sd_h^D$ which means there exists $\pt\in \Psi_h$ such that $\PP_h(\sd| \nu,T_D)=1$.  Thus, we get 
\begin{align*}
    &\norm{ \frac{\PP_h(\cdot| \pi^\star)}{\PP_h(\cdot| \unf(\Psi_h))}}_\infty = \max_{\sd \in \Sd_{h}^D,\sx\in \sx}  \frac{\PP_h(\sd| \pi^\star,\sx)}{\PP_h(\sd| \unf(\Psi_h))}\\
    &\leq \max_{\sd \in \Sd_{h}^D,\sx\in \sx}  \frac{\PP_h(\sd| \pi^\star,\sx)}{\PP_h(\sd| \unf(\Psi_h))}\\
    &\leq \frac{|\Psi_h|}{ 1-\eta h}.  \tag{By~\eqref{eq: state is reachable pspd}}
\end{align*}

{\bf Bound on the second term of~\eqref{eq: relation 2 pspd analysis}}. Observe that for any $\sd\in \Sd_h / \Sd_h^D$ it holds that
\begin{align}
\PP_h(\sd|\pi, P_{D,\eta})=0, \label{eq: translating policy covers relation 1}    
\end{align}
since $\sd$ is a non-reachable state in the $\eta$ close deterministic MDP $\Mcal_D$. Using this, and utilizing \pref{lem: perturbations of state action frequency} we get,
\begin{align*}
    &\sum_{\sd\in \Sdh / \Sd_{h}^D} \PP_h(\sd| \pi^\star)\\
    &= \sum_{\sd\in  \Sdh/\Sdh^{D,\eta}} \abr{\PP_h(\sd| \pi^\star)} = \sum_{\sd\in  \Sdh/\Sdh^{D,\eta}} \abr{\PP_h(\sd| \pi^\star) - \PP_h(\sd|\pi^\star, T_{D})} \tag{By~\eqref{eq: translating policy covers relation 1}}\\
    &\leq \sum_{\sd\in \Sdh} \abr{\PP_h(\sd| \pi^\star) - \PP_h(\sd|\pi, T_{D})}\\
    &= \norm{\PP_h(\cdot| \pi^\star) - \PP_h(\cdot|\pi^\star, T_{D})}\\
    &\leq h\eta.  \tag{By \pref{lem: perturbations of state action frequency}, holds for any $\pi$}
\end{align*}
Thus, the second term of~\eqref{eq: relation 2 pspd analysis} is bounded by 
\begin{align*}
    H\sum_{\sd\in \Sdh / \Sd_{h}^D} \PP_h(\sd| \pi^\star)\leq H^2\eta.
\end{align*}

{\bf Combining the bounds on the first and second term of~\eqref{eq: relation 2 pspd analysis}.} Plugging these bounds and using $|\Psi_h|\leq S$ (condiniong on the good event of $\algppe$), we conclude that
\begin{align*}
V(\pi^\star)  - V(\hat{\pi}) \leq  \frac{SH}{ 1-\eta h}\epsilon_{cb}+ H^3\eta\leq  2SH\epsilon_{cb}+ H^3\eta,
\end{align*}
since $\eta\leq \frac{1}{4 S H}\leq \frac{1}{4H}$ by assumption and since $S\geq 1$. Choosing
\begin{align*}
    N = O\rbr{\frac{S^2H^4 \log\rbr{\frac{|\Pi|H}{\delta}}}{\epsilon^2}},
\end{align*}
and applying a union bound such that \pref{lem: contextual bandits guarantee} holds for any $h\in [H]$ we conclude the proof.
\end{proof}

\subsection{Endogenous State Reward: Value Iteration}\label{app: plannning in EX-BMDP analysis VI}

In this section, we suggest a computationally easier procedure to get a near optimal policy given the output of $\algppe$. If the reward function depends only on the endogenous state then there exists an endogenous optimal policy (see \pref{prop: endo reward and optimal policy}). We show we can utilize this fact and get a near optimal policy by applying the value iteration procedure (VI). 

First, we claim the following:
    \emph{$\algppe$ can also return the $\eta$ near deterministic MDP of the endogenous dynamics $\Mcal_D$.}
Consider the elimination step of \pref{line:elimination}. When such elimination occurs, it implies that two paths $\pt_i,\pt_j$ lead to the same endogenous state. Conditioning on the good event, there is a one-to-one correspondence between $\nu\in \pts_h = \Psi_{h-1} \circ \Acal$ and $(\sd,a)$ where $\sd\in \Sd_{h-1}$ and endogenous state of the previous time step. This allows us to decode the states from the paths after the elimination step.

Let $\sd(\pt_i)$ and  $\sd(\pt_j)$ be the unique states of $\Mcal_D$ that are reached by $\pt_{i,h}$ and $\pt_{j,h}$ at time step $h$. Denote also $\pt_{i,h}= \pt_{i,h-1}\circ a_{i,h}$ and $\pt_{j,h}= \pt_{j,h-1}\circ a_{j,h}$ where $\pt_{l,h-1}$ denotes the first $h-1$ actions of the path $\pt_{l,h}$, and $ a_{l,h}$ denotes the action at time step $h$. If $\sd_{h,ij} = \sd_h(\pt_{i,h})=\sd_h(\pt_{j,h})$, i.e., $\pt_{i,h}$ and $\pt_{j,h}$ leads to the same endogenous state, $\algppe$  will eliminate one of the paths. In this case, we represent the state reached by both $\pt_i$ and $\pt_j$ using a single index. Furthermore, by an induction argument, $\pt_{i,h-1}$ and $\pt_{j,h-1}$ leads to different endogenous states $\sd(\pt_{i,h-1})$ and $\sd(\pt_{j,h-1})$. Thus, we record there is a transition from $(\sd(\pt_{i,h-1}) ,a_{i,h})$  and $(\sd(\pt_{j,h-1}) ,a_{j,h})$ to  $\sd_{h,ij}$.

Thus, instead of merely eliminating paths to create a minimal policy cover at each time step, we can also recover the $\eta$-close deterministic dynamics $\Mcal_D$.  Given this approximate MDP and an endogenous reward function, we can simply apply VI and recover a near optimal policy. Observe that we recover only a near optimal policy since the true dynamics is not the deterministic dynamics of $\Mcal_D$. However, the suboptimality gap can be easily bounded via a value difference lemma, since the models are $\eta$ close.

\begin{algorithm}[t]
\caption{$\VI(\epsilon,\delta, T_D, \Psi_h)$}
\label{alg:VI_alg}
\setstretch{1.2}
\begin{algorithmic}[1]
\State {\bf require: } $\epsilon,\delta>0$ $\cbr{\Psi_h}_{h=1}^H$, $T_D$ model of the near deterministic endogenous MDP, 
\State {\bf initialize: } Allocate $\cbr{Q_h(s,a)}_{h\in [H],s\in  [|\Psi_h|], a\in [A]}$
\For{$h = H,H-1,...,1$}
    \State Set $N = O\rbr{\frac{\log\rbr{\frac{|\Psi_h|A}{\delta}}}{\epsilon^2}}$
    \For{ $s\in [\Psi_h], a\in [a]$}
        \State Collect a dataset $\Dcal_{s,a}$ of $N$ \emph{i.i.d.} reward instances by following $\nu_s\circ a$ where $\nu=idx(s)$
        \State Estimate immediate reward $\hat{r}_h(s,a) = \frac{1}{N} \sum_{R_n\in \Dcal_{s,a}} R_n$
    \EndFor
\EndFor
\Return Optimal policy w.r.t. $\cbr{\hat{r}_h(s,a)}_{h\in [H],s_h\in |\Psi_h|,a\in \Acal}$ and $T_D$ via Value Iteration
\end{algorithmic}
\end{algorithm}

\propVIforPlanning*
\begin{proof}
{\bf First step.} The reward is approximated well. We saw that $\algppe$ returns the $\eta$ near deterministic model $\Mcal_D$ (with probability greater than $1-\delta$) denoted by $T_D$. Fix $h\in[H]$. To estimate the endogenous reward  $(\hat{s},a)$ up to accuracy $\epsilon>0$ it is sufficient to follow every $\nu\in \Psi_h$ and apply action $a\in \Acal$, where $\nu$ is the policy that reaches a unique state  $\sd \in \Sd^D_h$. This holds, since there is a one-to-one correspondence between states recovered by $\algppe$ and the reachable state space, in $h$ time steps, of the $\eta$ close deterministic MDP $\Mcal_D$. 

The estimated reward function is given by $r_D(\hat{s}_h,a) = \EE[ R(\hat{s}_h,a_h)\mid \nu\circ a]$, where $\hat{s}_h$ is equivalent to some open-loop policy $\nu\in \Psi_h$. It holds that,
\begin{align*}
    &r_D(\hat{s}_h,a) = \PP_h(s_h=\hat{s}_h,a_h=a|\nu\circ a)\EE[ R(s_h=\hat{s}_h,a=a)\mid \nu\circ a,\hat{s}_h,a_h] +\PP_h(s_h\neq \hat{s}_h|\nu\circ a)\\
    &=\PP_h(s_h=\hat{s}_h,a_h=a|\nu\circ a)\EE[ R(s_h=\hat{s}_h,a=a)\mid \hat{s}_h,a_h] +\PP_h(s_h\neq \hat{s}_h|\nu\circ a)\\
    &\leq r_h(\hat{s}_h,a) + \eta H. \tag{\pref{lem: perturbations of state action frequency}}
\end{align*}
On the other hand, via similar reasoning and since $R\in [0,1]$,
\begin{align*}
    \EE[ R(\hat{s}_h,a_h)\mid \nu\circ a] \geq (1-\eta H) r_h(\hat{s}_h,a) - \eta H \geq  r_h(\hat{s}_h,a) -2 H\eta.
\end{align*}
Thus,
\begin{align*}
    |\EE[ R(\hat{s}_h,a_h)\mid \nu\circ a]  - r_h(\hat{s}_h,a) |\leq 2\eta H.
\end{align*}

Since the estimated reward is $\widehat{r}_h(\hat{s}_h,a) = \frac{1}{N}\sum_{n} R_n$ where $ R_n\in [0,1]$ and $\EE[R_N]=\EE[ R(\hat{s}_h,a_h)\mid \nu\circ a]$, we get that given $N=O\rbr{\frac{H^2\log(\frac{|\Psi_h|AH}{\delta})}{\epsilon^2}}$ samples  for each $s\in |\Psi_h|,a\in \Acal,h\in [H]$, 
\begin{align*}
    |\widehat{r}_h(\hat{s}_h,a) - r_D(\hat{s}_h,a)| \leq \frac{\epsilon}{2H}
\end{align*}
 by Hoeffding's inequality and applying a union bound. Since $\Sdcard\geq |\Psi_h|$ for any $h$ the total needed number of samples is
 \begin{align*}
     N=O\rbr{\frac{SAH^2\log(\frac{SAH}{\delta})}{\epsilon^2}}.
 \end{align*}

{\bf Bounding gap from optimality.} Let $\Sd_h^D$ be the set of reachable states on the $\eta$ close endogenous deterministic MDP $\Mcal_D$. Denote by $\Mcal$ the true endogenous MDP. By the first step, and the fact both models are $\eta$ close it holds that for any  $h\in[H],s\in \Sd_h^D,a\in \Acal$
\begin{align*}
    &|\hat{r}(\hat{s},a) - r(\hat{s},a)|\leq \frac{\epsilon}{2H} + 2\eta H\\
    & \norm{T(\cdot| \hat{s},a) - T_D(\cdot| \hat{s},a))} \leq \eta.
\end{align*}

By the value difference lemma, it holds that for any $s\in \Sd$ it holds that
\begin{align*}
    &\abr{V^\pi_1(s;\Mcal) - V^\pi_1(s; \Mcal_D)} \\
    &\leq   \EE\left[\sum_{h'=h}^H \abr{r(s_{h'},a_{h'}) - \hat{r}(s_{h'},a_{h'})} + \abr{(T - T_D )(\cdot \mid s_{h'},a_{h'})^\top V^{\pi}_{h'+1}(\cdot;\Mcal)} \mid s_h=s,\pi,T_D \right]\\
    &\leq   \EE\left[\sum_{h'=h}^H \abr{r(s_{h'},a_{h'}) - \hat{r}(s_{h'},a_{h'})} + \norm{(T - T_D )(\cdot \mid s_{h'},a_{h'})}_1 \norm{V^{\pi}_{h'+1}(\cdot;\Mcal)}_{\infty} \mid s_h=s,\pi,T_D \right] \tag{Holder's inequality}\\
    &\leq 3\eta H^2 + \epsilon/2, 
\end{align*}
since for all $\sd\in \Sd$ $V^{\pi}_{h'+1}(\sd)\in [0,H]$, and the transition model is $\eta$ close. Thus, an optimal policy on $\Mcal_D$, denoted by $\widehat{\pi}$ is near optimal on the true MDP.
\begin{align*}
    V^{\hat{\pi}}_1(\sd; \Mcal) -  V^{\pi^\star}_1(\sd;\Mcal) \geq V^{\hat{\pi}}_1(\sd;\Mcal_D) -  V^{\pi^\star}_1(\sd; \Mcal_D)  -6 \eta H^3  - \epsilon,
\end{align*}
since $\hat{\pi}$ is optimal on $\Mcal_D$. This implies that
\begin{align}
    &V^{\hat{\pi}}_1(\sd)\geq V^{\pi^\star}_1(\sd; \Mcal) - 6 \eta H^3 - \epsilon  \nonumber \\
    \rightarrow& \sum_{\sd}\mu(\sd) V^{\hat{\pi}}_1(\sd; \Mcal)\geq \sum_{\sd}\mu(\sd) V^{\hat{\pi}}_1(\sd; \Mcal) - 6 \eta H^3 - \epsilon \label{eq: pre final bound VI},
\end{align}
since $\sum_{\sd}\mu(\sd)=1$ and $\mu(\sd)\geq 0$. Observe that, by the law of total probability, for both $\pi = \pi^*,\hat{\pi}$ it holds that
\begin{align*}
    &\sum_{\sd_1}\mu(\sd_1) V^{\pi}_1(\sd_1; \Mcal) = 
    \sum_{h=2}^H\sum_{\sd_h} \sum_{\sd_1}\mu(\sd_1)\PP_h(\sd_h|\sd_1,\pi) r_\pi(\sd_h) = \sum_{h=2}^H\sum_{\sd_h} \PP_h(\sd_h|\pi) r_\pi(\sd_h)
\end{align*}
Furthermore, observe that $\hat{\pi}$ is an open-loop policy and, thus, it is an endogenous policy. Furthermore, by \pref{prop: endo reward and optimal policy} $\pi^\star$ can be chosen to be an endogenous policy. This implies that for any $h\in [H],\sff\in \Sf$ such that $\sff= (\sd,\sx)$ 
\begin{align*}
    \PP_h(\sff|\pi^\star) = \PP_h(\sd|\pi) \PP_h(\sx)
\end{align*}
for $\pi = \pi^*,\hat{\pi}.$ By this fact, for both $\pi = \pi^*,\hat{\pi}$, it holds that
\begin{align*}
    &\sum_{h=2}^H\sum_{\sd_h} \PP_h(\sd_h|\pi) r_\pi(\sd_h) = \sum_{h=2}^H\sum_{\sd_h}\rbr{\sum_{\sx_h} \PP(\sx_h)} \PP_h(\sd_h|\pi) r_\pi(\sd_h)\\
    &=\sum_{h=2}^H \sum_{\sd_h}\sum_{\sx_h} \PP_h(\sx)\PP_h(\sd_h|\pi) r_\pi(\sd_h) \\
    &= \sum_{h=2}^H \sum_{\sff_h} \PP_h(\sff_h|\pi) r_\pi(\sd_h)  = V_1(\pi)
\end{align*}
Thus, we get that 
\begin{align}
    \sum_{\sd_1}\mu(\sd_1) V^{\pi}_1(\sd_1;\Mcal) = V_1(\pi) \label{eq: from endo to exo value},
\end{align}
where $V_1(\pi)$ is the value of $\pi$ on the joint endogenous-exogenous state space.

Observe that by \pref{prop: endo reward and optimal policy}, when the reward is a function of the endogenous state space, an optimal policy of $\Mcal$ is also an optimal policy on the joint endogenous-exogenous state space. Thus, combining~\eqref{eq: from endo to exo value} and~\eqref{eq: pre final bound VI} implies the result,
\begin{align*}
    V(\hat{\pi})\geq V^\star - 6 \eta H^2 - \epsilon,
\end{align*}
where $V^\star$ is the optimal value on the join endogenous-exogenous state space.
\end{proof}

\newpage

\newpage
\section{Existing Results}\label{app: existing results}
\begin{lemma}[E.g.~\cite{dann2017unifying}, Lemma E.15] \label{lem: value difference}
Consider two MDPs $\mathcal{M} = (\Scal, \Acal, T, r,H)$ and $\mathcal{M}' = (\Scal, \Acal, T',r',H)$. For any policy $\pi$ and any  $s,h$ the following relation holds:
\begin{align*}
    &V^\pi_h(s; \mathcal{M}) - V^\pi_h(s;\mathcal{M}') \\
    &= \EE\sbr{\sum_{h'=h}^H (r_h(s_{h'},a_{h'}) - r'_{h'}(s_{h'},a_{h'})) + (T - T')(\cdot \mid s_{h'},a_{h'})^\top V^{\pi}_{h'+1}(\cdot; \Mcal)\mid s_h=s,\pi,T'}.
\end{align*}
\end{lemma}

The following result is a consequence of existing bounds (e.g.~\cite{rosenberg2019online}, Lemma 28 or~\cite{efroni2021reinforcement}, Lemma 21). We provide the proof of this result for completeness and show it is a direct consequence of the value difference lemma.
\begin{lemma}[Perturbations of State Action Frequency Measure]\label{lem: perturbations of state action frequency}
Let $\pi$ be a fixed policy. Let $\Mcal_1$ and $\Mcal_{2}$ be two MDPs with $\eta$ close transition model, i.e., for any $s,a$ 
\begin{align*}
    \norm{T_1(\cdot|s,a) - T_2(\cdot| s,a)}\leq \eta.
\end{align*}
Then, for any $h\in [H]$ it holds that
\begin{align*}
    \norm{\PP_{h}(\cdot|\Mcal_1,\pi) - \PP_{h}(\cdot|\Mcal_2,\pi)}_1 \leq \eta h.
\end{align*}
\end{lemma}
\begin{proof}
Fix an $s\in \Scal_h$, and denote $$\PP_{h}(s|\Mcal_1,\pi) = \EE[\ind\cbr{s_h=s} | \Mcal_1],\PP_{h}(s|\Mcal_2,\pi) = \EE[\ind\cbr{s_h=s} | \Mcal_2].$$ This implies that
\begin{align}
    &\PP_{h}(s|\Mcal_1,\pi) - \PP_{h}(s|\Mcal_2,\pi) = \EE\sbr{\ind\cbr{s_h=s} \mid \Mcal_1} - \EE\sbr{\ind\cbr{s_h=s} \mid \Mcal_2} \nonumber \\
    & = \EE\sbr{\sum_{h'=1}^h\sum_{s'\in \Scal_{h'+1}}\rbr{T_{1,h'}(s'|s_{h'},a_{h'}) - T_{2,h'}(s' |s_{h'},a_{h'})}V_{h'+1}(s'; \Mcal_{2},s) \mid T_1,\pi}\label{eq: perturbations of state action frequencey}
\end{align}
where the last relation holds by the value difference lemma (\pref{lem: value difference}). See that 
\begin{align}
V_{h'+1}(s'; \Mcal_{2},s) =\EE[\ind\cbr{s_h=s} | \Mcal_{2}, s', \pi]\geq 0, \label{eq: perturbations of state-action freq, help relation 1}
\end{align}
which also implies that for any $s'\in \Scal_{h'+1}$
\begin{align}
    \sum_{s\in \Scal_h} V_{h'+1}(s'; \Mcal_{2},s) =\EE[\sum_{s\in \Scal_h}\ind\cbr{s_h=s} | \Mcal_{2}, s', \pi]=1\label{eq: perturbations of state-action freq, help relation 2}
\end{align}
since the indicator is non-zero only on a single state at the $h^{th}$ time step. By utilizing the above we get
\begin{align*}
    &\norm{\PP_{\pi}(\cdot|\Mcal_1) - \PP_{\pi}(\cdot|\Mcal_2)}_1 = \sum_{s\in \Scal_h} \abr{\PP_{h}(s|\Mcal_1,\pi) - \PP_{h}(s|\Mcal_2,\pi)}\\
    &=\sum_{s\in \Scal_h} \abr{\EE\sbr{\sum_{h'=1}^h\sum_{s'\in \Scal_{h'+1}}\rbr{T_{1,h'}(s'|s_{h'},a_{h'}) - T_{2,h'}(s' |s_{h'},a_{h'})}V_{h'+1}(s'; \Mcal_{2},s) |T_1,\pi}} \tag{By~\eqref{eq: perturbations of state action frequencey}}\\
    &\leq \EE\sbr{\sum_{h'=1}^h\sum_{s'\in \Scal_{h'+1}}\abr{T_{1,h'}(s'|s_{h'},a_{h'}) - T_{2,h'}(s' |s_{h'},a_{h'})}\sum_{s\in \Scal_h}\abr{V_{h'+1}(s'; \Mcal_{2},s)} |T_1,\pi} \tag{Triangle's inequality}\\
    &=\EE\sbr{\sum_{h'=1}^h\sum_{s'\in \Scal_{h'+1}}\abr{T_{1,h'}(s'|s_{h'},a_{h'}) - T_{2,h'}(s' |s_{h'},a_{h'})}\sum_{s\in \Scal_h}V_{h'+1}(s'; \Mcal_{2},s) |T_1,\pi} \tag{By~\eqref{eq: perturbations of state-action freq, help relation 1}}\\
    &=\EE\sbr{\sum_{h'=1}^h\sum_{s'\in \Scal_{h'+1}}\abr{T_{1,h'}(s'|s_{h'},a_{h'}) - T_{2,h'}(s' |s_{h'},a_{h'})} |T_1,\pi}\tag{By~\eqref{eq: perturbations of state-action freq, help relation 2}}\\
    &\leq \eta h.
\end{align*}
where the last relation holds by assumption since by $\Mcal_1$ and $\Mcal_2$ are $\eta$ close in the $L_1$ norm,
\begin{align*}
    \sum_{s'\in \Scal_{h'+1}}\abr{T_{1,h'}(s'|s_{h'},a_{h'}) - T_{2,h'}(s' |s_{h'},a_{h'})} = \norm{T_{1,h'}(\cdot|s_{h'},a_{h'}) - T_{2,h'}(\cdot |s_{h'},a_{h'})}\leq \eta.
\end{align*}
\end{proof}

\begin{lemma}[see \cite{langford2008epoch,misra2020kinematic}] \label{lem: contextual bandits guarantee}
Let $D = (x_i,a_i,r_i)$ be sampled i.i.d and $r_i\in [0,H]$. Let $\hat{\pi}$ be a solution of the offline contextual bandit optimization routine 
\begin{align*}
    \hat{\pi} \in \arg\max_{\pi\in \Pi} \sum_{x_i,a_i,r_i} \EE_{a'\sim \pi(\cdot|x)}\sbr{ \frac{r_i \one\cbr{a'=a}}{1/A}}.
\end{align*}
Then,
\begin{align*}
    \EE_{x,r}[r(\hat{\pi}(x))] \geq \max_{\pi\in \Pi} \EE_{x,r}[r(\pi(x))] -\epsilon_{cb},
\end{align*}
where $\epsilon_{cb} = 4H \sqrt{\frac{A \log\rbr{\frac{2|\Pi|}{\delta}}}{N}}$ for $\epsilon_{cb}\leq 1/2$.
\end{lemma}

\begin{theorem}[MLE Guarantees,\cite{agarwal2020flambe}]\label{thm:mle} Fix $\delta \in (0, 1)$. If $f^\star \in \Fcal$ (realizability assumption), then the maximum likelihood estimator $\hat{f}$ satisfies:
\begin{equation*}
    \EE_{x' \sim D}\left[\TV{\hat{f}(\cdot \mid x) - f^\star(\cdot \mid x)} \right]  \le \epsilon = \sqrt{\frac{2}{N} \ln \frac{|\Fcal|}{\delta}},
\end{equation*}
with probability at least $1-\delta$.
\end{theorem}

\newpage
\section{Experiment Details}
\label{app:exp}

\subsection{Algorithm Details.}

We describe the implementation details of various algorithms for the combination lock problem. We describe the implementation details of $\algppe$, which provably solves this task, as well as three alternatives, {\tt Homer}, {\tt PPO} and {\tt PPO + RND}. For each algorithm, we do grid search over their most crucial hyperparameters.  We measure performance based on number of episodes needed to achieve a mean regret of at most $\nicefrac{V(\pi^\star)}{2}$. We run each experiment 5 times with different seeds and report median performance. For each value of $h$, we select the hyperparameter with the best median value. 

\paragraph{$\algppe$ Details.} Our implementation of $\algppe$ is almost identical to the pseudocode in~\pref{alg:genik_path_elim}. For each time step $h$, we collect a dataset of size $N$ and solve a multi-class classification problem (\pref{line:explore-app}). We train the model with mini-batches and use Adam optimization. We perform gradient clipping to limit the norm of gradient to a given value $\kappa_{n}$. We separate a certain percent ($q_{pct}$) of the dataset and use it as a validation set. We train the model for a maximum number of epochs ($n_{max}$) and compute the performance on the validation set after each epoch. We stop when the performance stops improving for $\kappa_p$ number of epochs where $\kappa_p$ term is called \emph{patience}. We use the model with the best performance on the validation set. All parameters are initialized randomly and trained by the algorithm. 

We use a two-layer feed-forward network to model the function class $\Fcal$. We use Leaky-ReLu non-linearity and use hidden dimension ($h_{dim}$) of 56. The last layer applies a linear transformation to map the hidden vector to a vector of size equal to the number of paths to predict over. Finally, we apply a softmax layer to convert this vector to a probability distribution.

Hyperparameter values for $\algppe$ for the combination lock problem are described in~\pref{tab:dpcid_hyperparam}. We do grid search over $N \in \{500, 2000, 5000\}$.  

\begin{table*}[h]
    \centering
    \begin{tabular}{l|c}
    \hline
         \textbf{Hyperparameter} & \textbf{Values} \\
         \hline
         Learning rate & 0.001\\
         $N$ & $\{500, 2000, 5000\}$ \\
         Batch size & 256\\
         Grad clip norm ($\kappa_{n}$) & 10\\
         Patience ($\kappa_p$) & 20 \\
         Optimization & Adam \\
         Max epochs ($n_{max}$) & 50\\
         Hidden dimension ($h_{dim}$) & 56\\
         Validation percent ($q_{pct}$) & 20\%\\
         Parameter Initialization & PyTorch 1.4 Default\\
         \hline
    \end{tabular}
    \caption{Hyperparameter values for $\algppe$}
    \label{tab:dpcid_hyperparam}
\end{table*}

\paragraph{${\tt Homer}$ Details.} Similar to $\algppe$, ${\tt Homer}$~\citep{misra2020kinematic} is an iterative algorithm that learns a policy cover incrementally by learning state-abstraction per time step. ${\tt Homer}$ learns the state-abstraction using noise-contrastive estimation by training a model to predict if a given transition $(x, a, x')$ is causal or acausal. The algorithm  creates a dataset of $N$ quads $(x, a, x', y)$ where $(x, a, x')$ is a transition and $y=1$ implies that it is causal and $y=0$ implies it is acausal. To describe the data collection process, we define the sampling procedure $(x, a, x') \sim \unf(\Psi_{h-1}) \circ \unf(\Acal)$ which first samples $\pi \sim \unf(\Psi_{h-1})$ and follows $\pi$ till time step $h-1$ to observe $x$, and then take an action $a \sim \unf(\Acal)$, and observe $x' \sim T(\cdot \mid x, a)$. ${\tt Homer}$ generates a single datapoint by sampling two transitions $(x^{(1)}, a^{(1)}, x'^{(1)})$ and $(x^{(1)}, a^{(1)}, x'^{(1)}$ from $\unf(\Psi_{h-1}) \circ \unf(\Acal)$. It then samples $y \sim \unf(\{0, 1\})$ and if $y=1$ then we add $(x^{(1)}, a^{(1)}, x'^{(1)}, y)$ to the dataset, otherwise, we add $(x^{(1)}, a^{(1)}, x'^{(2)}, y)$. The learning approach uses a budget $N_{budget}$ for the number of abstract states to learn at each time step. We use the code provided to us by authors with their permission. We use the model architecture proposed  by~\cite{misra2020kinematic} who solve a problem similar to combination lock but without exogenous distractors. 

The original ${\tt Homer}$ algorithm proposes using the general PSDP algorithm to learn the policy cover. However, PSDP is an extremely computationally inefficient algorithm. The authors discussed a more greedy approach, however, that approach still relies on solving a classification problem. Since our domain is deterministic, therefore, we use an even simpler open-loop policy search to learn the policy cover. Formally, given a reward function $R$ and horizon $h$, we create a dataset of tuples $(\pi \circ a, r)$ by sampling $\pi \in \unf(\Psi_{h-1})$, $a \sim \unf(\Acal)$, and then following the open-loop policy $\pi$ till time step $h-1$ and then taking action $a$, and $r$ is the total reward achieved. For each extended open-loop policy $\pi \circ a$, we compute the average total reward $\bar{r}_{\pi \circ a}$ using the collected dataset. Finally, we compute the optimal open-loop policy as $\arg\max_{\pi\circ a} \bar{r}_{\pi \circ a}$. We reuse the dataset from the state abstraction learning for the planning phase.

Hyperparameter values for ${\tt Homer}$ are provided in~\pref{tab:homer_hyperparam}. Most computational oracle specific hyperparameters are chosen exactly as for $\algppe$. We do grid search over the value of $n \in \{2000, 5000, 10000\}$. We found that for no values of $n$, ${\tt Homer}$ was able to solve the problem for $H=5$. However, removing exogenous distractors from the problem resulted in success.

\begin{table*}[h]
    \centering
    \begin{tabular}{l|c}
    \hline
         \textbf{Hyperparameter} & \textbf{Values} \\
         \hline
         $N_{budget}$ & 2\\
         $N$ & $\{2000, 5000, 10000\}$\\
         Learning rate & 0.001\\
         Batch size & 256\\
         Grad clip norm ($\kappa_{n}$) & 10\\
         Patience ($\kappa_p$) & 20 \\
         Optimization & Adam \\
         Max epochs ($n_{max}$) & 50\\
         Hidden dimension ($h_{dim}$) & 56\\
         Validation percent ($q_{pct}$) & 20\%\\
         Parameter Initialization & PyTorch 1.4 Default\\
         \hline
    \end{tabular}
    \caption{Hyperparameter values for ${\tt Homer}$ and ${\tt ID}$}
    \label{tab:homer_hyperparam}
\end{table*}

\paragraph{${\tt ID}$ Details.} Our implementation of ${\tt ID}$ baseline is exactly similar to ${\tt Homer}$. The only difference is that instead of predicting whether a given transition $(x, a, x')$ is real or imposter, we predict the action $a$ given $(x, x')$ for real transitions. As such, the hyperparameter values for ${\tt ID}$ is same as in ~\pref{tab:homer_hyperparam}.

\paragraph{${\tt PPO}$ and ${\tt PPO + RND}$ Details.} PPO is an actor-critic algorithm which uses entropy-regularization for performing exploration. We also consider augmentation of PPO with exploration bonus using random network distillation ({\tt PPO + RND}). Random network distillation uses a fixed randomly initialized network $g$. Given an observation $x$, the model trains another network to predict the output $g(x)$. An exploration bonus is derived from the prediction error. As the model explores part of the state space, its prediction error for observations emitted by these states goes down, and the model is no longer incentivized to visit these states.

Hyperparameter values for ${\tt PPO}$ and ${\tt PPO+RND}$ are shown in~\pref{tab:ppo_hyperparam}. We do grid search over entropy coefficient in $\{0.1, 0.01\}$. For ${\tt PPO + RND}$, we also do grid search over RND bonus coefficient in $\{100, 500\}$. 

\begin{table*}[h]
    \centering
    \begin{tabular}{l|c}
    \hline
         \textbf{Hyperparameter} & \textbf{Values} \\
         \hline
         Entropy coefficient & $\{0.1, 0.01\}$\\
         Number of PPO updates & 4\\
         Clipping parameter & 0.1\\
         Learning rate & 0.001\\
         Batch size & 256\\
         Grad clip norm ($\kappa_{n}$) & 10\\
         Patience ($\kappa_p$) & 20 \\
         Optimization & Adam \\
         Max epochs ($n_{max}$) & 50\\
         Hidden dimension ($h_{dim}$) & 56\\
         Validation percent ($q_{pct}$) & 20\%\\
         Parameter Initialization & PyTorch 1.4 Default\\[.1cm]
         \hline 
         \multicolumn{2}{l}{only for ${\tt PPO + RND}$} \\
         \hline
        RND bonus coefficient & $\{100, 500\}$ \\
         \hline
    \end{tabular}
    \caption{Hyperparameter values for ${\tt PPO}$ and ${\tt PPO+RND}$}
    \label{tab:ppo_hyperparam}
\end{table*}

\paragraph{${\tt Bisimulation}$ Details.} We use the bisimulation code provided by~\cite{zhang2021learning}.\footnote{https://github.com/facebookresearch/deep\_bisim4control} We make minimal changes to the repository in order to refactor it for our experiments. Two changes were required, however, to adapt the codebase to our experiments. Firstly, we use a two-layer feed-forward network as encoder and decoder instead of a convolutional neural network (CNN). Since observations in the combination lock experiment consist of a 1D feature vector, therefore, a CNN architecture didn't make much sense. Secondly, the codebase used soft-actor critic for a continuous action space problems. We adapt it to our purpose by discretizing the agent's action. Formally, given a continuous action $u \in [-1, 1]$ generated by the policy, we take the action  $a = \floor{1/2 \times (u + 1) \times 10}$. Since each discrete action has the same probability of being generated under uniform distribution over $u$, therefore, this relaxation does not introduce any extra hardness.

\paragraph{Compute Infrastructure and Different Runs.} We run experiments on a clusters with a mixture of P40, P100, and V100 GPUs. Each experiment runs on a single GPU in a docker container. We ran each experiment 5 times with different seeds.

\subsection{Additional Detail for State Decoding Experiment.}

We use $m=5000$ samples to evaluate the state decoding methods. We run each experiment 10 times with different seed and show mean and standard deviation in~\pref{fig:combolock-results}c. Lastly, we define the distribution $D(x) = \EE_{\pi \sim \unf(\Psi)}\left[ \PP_3(x \mid \pi)\right]$ where $\Psi$ contains an open-loop policy for each of the three states reachable at time step $3$, namely, $\{s_{3, a}, s_{3, b}, s_{3, c}\}$.

\subsection{Additional Detail for Visual Grid World Experiment.}

The agent can take five actions: forward by one step $(F)$, turn-left by 90-degree $(L)$, turn right by 90-degree $(R)$, and two compositional actions $LF$ and $RF$ that first take action $R$ or $L$, and then take action $F$. The agent receives a reward of +1 for reaching the goal and a reward of -1 for reaching lava and a small negative reward otherwise. 

Our implementation of $\algppe$ remains the same as before. However, we use a different architecture for $\Fcal$. We use a two-layer convolutional network with ReLu non-linearity. The first layer applies 16 $8\times 8$ kernel with stride $4$ and the second layer applies 32 $2\times 2$ kernel with stride $2$. Finally, we flatten the representation and pass it through a linear layer to a vector of appropriate size. We do not do any image pre-processing and do not use any pre-trained models.

\subsection{$\algppe$ pseudocode used for experiments}

We present the pseudocode that we used in experiments in~\pref{alg:genik_path_elim_full}. This pseudocode is optimized for problems with  deterministic endogenous transition dynamics with reward function that depends on endogenous state. However, as stated earlier, when the reward function depends on exogenous state or observation, then we can use PSDP (\pref{app: plannning in EX-BMDP analysis PSDP}. And when the reward function depends only on endogenous state but the transition dynamics are near-deterministic, then we can use the decoder to learn the transition dynamics and reward and simply use value iteration.

The algorithm is exactly the same as $\algppe$ with extra details to show how value iteration is done in practice. In particular, we show how transition function and reward function are estimated. Formally, for extracting the transition function we define a map $\Ucal$ that given an index of a path $i$ contains a set of paths that are \emph{merged} with the $i^{th}$ path, that is $\pt_i$ (\pref{line:u-map-app}). Intuitively, two paths merge if they reach the same endogenous state. Initially, each map is only merged with itself and hence $\Ucal(i)$ is defined as $\{\pt_i\}$. When we compare path $i$ and $j$ and eliminate $\pt_j$, we add all the paths that were merged with $\pt_j$ as also merged with $\pt_i$ (\pref{line:u-map-app-merge}).

We define the recovered latent tabular MDP $\widehat{\Mcal}$ as follows. We firstly, recover the state space, transition dynamics and reward function for each time step separately. We define a state space $\hat{\Scal}_h$ for the $h^{th}$ time step as containing a state for each path in $\Psi_h$. We then define a deterministic transition model $\hat{T}_{h-1}: \hat{\Scal}_{h-1} \times \Acal \rightarrow \hat{\Scal}_h$ as $\hat{T}(i, a) = j$, if the path $\pt^{h-1}_i \circ a \in \Psi_{h-1} \times \Acal$ with $\pt^{h-1}_i = \Psi_{h-1}$, gets merged with path $\pt_j \in \Psi_h$ (\pref{line:transition-estimation}). Note that we use the superscript $h-1$ to denote that $\pt^{h-1}_i$ is not the $i^{th}$ path in $\Psi_{h-1} \times \Acal$ but in $\Psi{h-2}\times \Acal$. Further, note that indices of a path are defined with respect to the set $\Psi_t \times \Acal$ containing them, and remain fixed for the whole training time. Lastly, we compute the reward by using average of reward in the dataset collected corresponding to the appropriate path (\pref{line:reward-estimation}).

Finally, we do value iteration on the recovered MDP $\widehat{\Mcal}$ which has a set of states $\hat{\Scal}_h$ reachable at time step $h$, a action space $\Acal$, a transition dynamics $\hat{T}_h$ and reward function $\hat{R}_h$ for the $h^{th}$ time step, a horizon $H$, and a deterministic start state of $1 \in \hat{\Scal}_1$ (\pref{line:value-iteration}). As the latent endogenous transition dynamics are deterministic, therefore, an optimal open loop policy exists. We return this optimal policy along with policy cover, decoder, and recovered MDP. These provide pretty much all important objects that we can recover from a given Exogenous Block MDP.

\begin{algorithm}[t]
\caption{$\algppe(N)$: \algppelong as run in practice}
\label{alg:genik_path_elim_full}
\setstretch{1.25}
\begin{algorithmic}[1] 
\State Set $\Psi_1 = \{\bot\}$ and $\hat{\Scal}_1 = \{1\} $\mycomment{$\bot$ denotes an empty path}
\For{$h=2,\ldots,H + 1$}
\State Collect a dataset $\Dcal$ of $N$ \emph{i.i.d.} tuples $(x, r, \pt)$ where $\pt \sim \unf(\Psi_{h-1}\circ \Acal)$, $r = R(x_{h-1}, a_{h-1})$, and $x \sim \PP(x_h \mid \pt)$. \label{line:explore-app}
\State Define $\Ucal(i) = \{\pt_i\}$ for all i in $\{1, 2, \cdots, |\Psi_{h-1} \circ \Acal|\}$. \label{line:u-map-app}
\State Solve multi-class classification problem:
            $
            \hat{f}_{h} = \arg\max_{f \in \Fcal} \sum_{(x, r, \pt) \in \Dcal} \ln f(\textrm{idx}(\pt) \mid x)
            $.
\For{$1 \le i < j \le |\Psi_{h-1}\circ \Acal|$}
        \State Calculate the path prediction gap:
            $
          \widehat{\Delta}(i, j) = \frac{1}{N}\sum_{(x, \pt) \in \Dcal} \abr{\hat{f}_h(i | x) - \hat{f}_h(j | x)}.
            $
        \If{$\widehat{\Delta}(i, j) \leq \frac{5/8}{|\pts_{h}|}$}, 
        \State eliminate path $\pt$ with $\mathrm{idx}(\pt)=j$. \mycomment{$\pt_i$ and $\pt_j$ visit the same state}
        \State $\Ucal(i) = \Ucal(i) \cup \Ucal(j)$ \mycomment{paths merged with $\pt_j$ are now merged with $\pt_i$} \label{line:u-map-app-merge}
        \EndIf\label{line:elimination-app}
    \EndFor
\State $\Psi_h$ is defined as the set of all paths in $\Psi_{h-1} \circ \Acal$ that have not been eliminated in~\pref{line:elimination-app}.
\State Define decoder $\hat{\phi}_h: x \mapsto \min_{i} \left\{ i \mid \hat{f}_h(i \mid x) \ge \max_j \hat{f}_h(j \mid x) - \Ocal(\nicefrac{1}{|\pts_h|}), i \in |\pts_h| \right\}.$
\State Define $\hat{\Scal}_h = \{\textrm{idx}(\nu) \mid \nu \in \Psi_h\}$ \mycomment{create a state for each path which is not eliminated}
\State Define $\hat{T}_{h-1}: \hat{\Scal}_{h-1} \times \Acal \rightarrow \hat{\Scal}_h$ as $T(i, a) = j$ if $\pt^{h-1}_i \circ a \in \Ucal(j)$ where $\pt^{h-1}_i \in \Psi_{h-1}$.\label{line:transition-estimation}
\State Define $\hat{R}_{h-1}: \hat{\Scal}_{h-1} \times \Acal \rightarrow [0, 1]$ as $\hat{R}_h(i, a) = {\tt average}\left\{r \mid (x, r, \pt) \in \Dcal, \pt = \pt_i \circ a \right\}$ \label{line:reward-estimation}
\EndFor
\State Perform value iteration on the tabular MDP $\widehat{\Mcal} = \left(\{\hat{\Scal}_h\}_{h=1}^{H+1}, \Acal, \{\hat{T}_h\}_{h=1}^H, \{R_h\}_{h=1}^H, H, 1\right)$ and return the optimal open-loop policy $\hat{\pi}$. \label{line:value-iteration}
\State \Return $\hat{\pi}$, $\widehat{\Mcal}$, $\left\{\hat{\phi}_h\right\}_{h=2}^H$, and $\left\{\Psi_h\right\}_{h=2}^H$ \mycomment{return optimal policy, learned latent MDP, decoder, and policy cover}
\end{algorithmic}
\end{algorithm}

\end{document}